\documentclass[10pt,journal,compsoc]{IEEEtran}

\usepackage{times}
\usepackage{euscript}
\usepackage{epsfig}
\usepackage{graphicx}
\usepackage{amsmath,bm,amsfonts,mathrsfs}
\usepackage{mcode}
\usepackage{amssymb}
\usepackage{algorithm}
\usepackage{algpseudocode}

\usepackage{enumerate}
\usepackage{threeparttable}
\usepackage{multirow}
\usepackage{booktabs}
\usepackage{url}
\usepackage{caption,color}
\usepackage{threeparttable}
\usepackage{balance}
\usepackage{amsmath}
\usepackage{amssymb}
\usepackage{url}
\usepackage{stmaryrd}
\usepackage{pgf}
\usepackage{tikz}
\usepackage{pgfplots}
\pgfplotsset{compat=1.9}
\usepackage{bigstrut}
\usepackage{multirow}
\usepackage{dblfloatfix}
\usepackage{pdfpages}
\usepackage{fancyhdr}
\usepackage{enumitem}

\usepackage{graphicx}
\usepackage[lofdepth,lotdepth]{subfig}

\usepackage{colonequals}

\usepackage{xr}

\newcommand{\st}{{\text{s.t.}}}

\newcommand{\B}{\bm{\mathcal{B}}}

\newcommand{\E}{\bm{\mathcal{E}}}

\newcommand{\I}{\bm{\mathcal{I}}}

\newcommand{\Sss}{\bm{\EuScript{S}}}

\newcommand{\Sm}{\bm{S}}

\newcommand{\norm}[1]{\lVert#1\rVert}

{%
\begin{list}{#1}{
\vspace{-\topsep}
\vspace{-\partopsep}
\setlength{\itemindent}{0cm}
\setlength{\rightmargin}{0cm}
\setlength{\listparindent}{0cm}
\settowidth{\labelwidth}{#1}
\setlength{\leftmargin}{\labelwidth}
\addtolength{\leftmargin}{\labelsep}
\setlength{\itemsep}{0cm}
}%
}%
{%
\end{list}
\vspace{-\topsep}
\vspace{-\partopsep}
}

{\begin{enumerate}%
}%
{\end{enumerate}}

\usepackage{docmute} 

\makeatletter
\def\input@path{{../}}
\makeatother

\usepackage{array}
\newcolumntype{L}{>{\arraybackslash}m{6cm}}
\usepackage{relsize}
\tikzset{fontscale/.style = {font=\relsize{#1}}}

\newcounter{alphasect}
\def\alphainsection{0}

\let\oldsection=\section
\def\section{%
	\ifnum\alphainsection=1%
	\addtocounter{alphasect}{1}
	\fi%
	\oldsection}%

\renewcommand\thesection{%
	\ifnum\alphainsection=1%
	\Alph{alphasect}
	\else%
	\arabic{section}
	\fi%
}%

\newenvironment{alphasection}{%
	\ifnum\alphainsection=1%
	\errhelp={Let other blocks end at the beginning of the next block.}
	\errmessage{Nested Alpha section not allowed}
	\fi%
	\setcounter{alphasect}{0}
	\def\alphainsection{1}
}{%
	\setcounter{alphasect}{0}
	\def\alphainsection{0}
}%

\begin{document}

\title{Morton Filters for Superior Template Protection for Iris Recognition\thanks{This is an extended work of our earlier submission to BTAS-2019 \cite{kiran2019morton}.}}

\author{Kiran B. Raja$^{\dagger\ddag}$  \quad R. Raghavendra$^{\ddag}$  \quad Sushma Venkatesh$^{\ddag}$  \quad Christoph Busch$^{\ddag}$  \\
	$^{\dagger}$ The Norwegian Colour and Visual Computing Laboratory, NTNU - Gj{\o}vik, Norway\\
	$^{\ddag}$ Norwegian Biometrics Laboratory, NTNU - Gj{\o}vik, Norway\\
	\{\tt\small kiran.raja; raghavendra.ramachandra; sushma.venkatesh; christoph.busch\} @ntnu.no\\
}

\maketitle
\thispagestyle{empty}

\begin{abstract}
In this work, we address the fundamental performance issues of template protection for iris verification. We base our work on the popular Bloom-Filter templates protection  and address the key challenges like sub-optimal performance and low unlinkability. Specifically, we focus on cases where Bloom-filter templates results in non-ideal performance due to presence of large degradations within iris images. Iris recognition is challenged with number of occluding factors such as presence of eye-lashes within captured image, occlusion due to eyelids, low quality iris images due to motion blur amongst many other. All of such degrading factors result in obtaining non-reliable iris codes and thereby provide non-ideal biometric performance. These factors further directly impact the protected templates derived from the iris images when classical Bloom-filters are employed. To this end, we propose and extend our earlier ideas of Morton-filters for obtaining better and reliable templates for iris. Morton filter based template protection for iris codes is based on leveraging the intra-class and inter-class distribution by exploiting the low-rank iris codes to derive the stable bits across the iris images for a particular subject and also analyzing the discriminable bits across various subjects. Such low-rank non-noisy iris codes enables realizing the template protection in a superior way which not only can be used in constrained setting, but also can be used in relaxed iris imaging. We further extend the work to analyze the applicability to visible spectrum iris images by employing a large scale public iris image database - UBIRIS (v1 and v2), captured in a unconstrained setting. Through a set of thorough experiments, we demonstrate the applicability of proposed approach and vet the strengths and weakness. Yet another contribution of this work stems in assessing the security of the proposed approach where factors of Unlinkability is studied to indicate the antagonistic nature to relaxed iris imaging scenarios.
\end{abstract}

\section{Introduction}
\begin{figure}[htp]
	\centering
	\resizebox{0.8\linewidth}{!}{
		\begin{tabular}{ccc}
			\hline\\
			\includegraphics[width=0.33\textwidth]{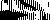} & 
			\includegraphics[width=0.33\textwidth]{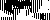} & 
			\includegraphics[width=0.33\textwidth]{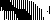} \\
			\color{black} \Huge\textbf{Sample 1} & \color{black} \Huge\textbf{Sample 2}& \color{black} \Huge\textbf{Sample 3}\\ 
			\includegraphics[width=0.33\textwidth]{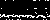} & \includegraphics[width=0.33\textwidth]{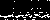}& \includegraphics[width=0.33\textwidth]{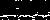} \\
			\color{black} \Huge\textbf{Bloom Template - 1} & \Huge\color{black} \Huge\textbf{Bloom Template - 2}& \color{black} \Huge\textbf{Bloom Template - 3}\\
			\hline\\
			\multicolumn{3}{c}{\Huge \textbf{Subject 1}}\\
			\hline\\
			\includegraphics[width=0.33\textwidth]{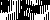} & 
			\includegraphics[width=0.33\textwidth]{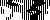} & 
			\includegraphics[width=0.33\textwidth]{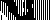} \\
			\color{black} \Huge\textbf{Sample 1} & \color{black} \Huge\textbf{Sample 2}& \color{black} \Huge\textbf{Sample 3}\\ 
			\includegraphics[width=0.33\textwidth]{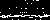} & \includegraphics[width=0.33\textwidth]{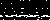}& \includegraphics[width=0.33\textwidth]{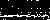} \\
			\color{black} \Huge\textbf{Bloom Template - 1} & \Huge\color{black} \Huge\textbf{Bloom Template - 2}& \color{black} \Huge\textbf{Bloom Template - 3}\\
			\hline\\
			\multicolumn{3}{c}{\Huge \textbf{Subject 2}}\\
			\hline\\
		\end{tabular}
	}
	\caption{Bloom Protected Templates for two subjects for multiple samples from CASIA v4 Distance Dataset \cite{casia-v4-db}. Note the iris code is not consistent across multiple samples for same subject and therefore the Bloom templates obtained are not consistent. Further, a high number of same bits are set for bloom templates for both subjects leading to high number of false accepts.}
	\label{fig:failure-analysis-bloom}
	\vspace{-6mm}
\end{figure}

With the growing need for secure access control in many domains, biometrics has been employed as an ubiquitous way to identify and verify the identity of subjects. Among the well used biometric characteristics such as face, fingerprint, iris, palmprint etc., iris recognition has been preferred way for highly secure applications. The iris patterns begins to form during the third month of gestation and the structures creating it's striking patterns are developed by the eighth month \cite{kronfeld1962gross, wolff1948anatomy}. Despite the pigment accretion continuing in the postnatal years, the layers of the iris have both ectodermal and mesodermal origin, consisting of dilator and sphincter muscles, a vascularized stroma, and an anterior layer with a genetically determined density of melanin pigment granules \cite{daugman2001epigenetic,imesch1997color}. A combination of all of these factors results in a complex visible iris pattern displaying distinctive features such as arching ligaments, crypts,ridges, and a zigzag collarette\cite{daugman2001epigenetic} making it a unique biometric feature with low probability of impostor collision \cite{daugman2001epigenetic, daugman2009iris}. 

\begin{table*}[b]
	\caption{State-of-art approaches for template protection in iris recognition}
	\label{tab:sota-iris-template}
	\centering
		\Large
		\resizebox{0.98\linewidth}{!}{
		\begin{tabular}{cLLccc}\hline
			Previous Work & Approach & Contribution &  Dataset Type & Database & Accuracy\\\hline
			Yang and Verbauwhede \cite{yang2007secure} & Error Correcting Code (ECC) based BTP &  Bose-Chaudhuri-Hochquenghem (BCH) code of a random bit-stream & Constrained NIR Iris & -- &  -- \\\hline
			Nandakumar and Jain \cite{nandakumar2008multibiometric} & Fuzzy-vault scheme to derive private keys from iris patterns. & Fixed-length binary vector representation of iriscode into an unordered set representation & Constrained NIR & CASIA v1 Iris & -- \\\hline
			Maiorana et al., \cite{maiorana2014iris} & Turbo codes with soft-decoding for iris & High performance in terms of both verification rates and security & Constrained NIR  & CASIA-Iris V4 database & -- \\\hline
			Zhang et al., \cite{zhang2009robust} & Concatenated coding scheme and bit masking scheme & A bit masking scheme was proposed to minimize and randomize the errors & Constrained Internal & CASIA Interval & 0.52\% EER\\\hline
			Rathgeb et al. \cite{rathgeb2013alignment} & Bloom-filter based biometric template protection & Alignment free template creation & Constrained NIR & CASIA-v3 Interval Iris & 1.19 \% EER \\\hline
			Rathgeb and Busch \cite{rathgeb2014cancelable, rathgeb2014application} & Adaptive Bloom filter-based transforms & The irreversible mixing transform generating alignment-free templates & Constrained NIR & IITD Iris Dataset & 0.5\% EER\\\hline
			Gomez-Barrero  et al. \cite{gomez2018multi} & Generic framework for generating an irreversible representation & Feature level fusion of different biometrics (face and iris) to a single protected template & Constrained NIR & IITD Iris Database version 1.0 & 0.5\% EER \\\hline
			Lai et al \cite{lai2017cancellable} & Cancellable iris template with Jaccard similarity matcher & Low error rate and attack resistant & Constrained NIR  & CASIA v3 iris database & 0.16\% EER \\\hline\hline
			&  &  &  & IITD Iris Dataset,   & 0\% EER for NIR, \bigstrut[t]\\
			This work \cite{kiran2019morton} & Morton-Filter & Very low error & Constrained and  &(NIR and VIS) & 15 \% EER on VIS\\
			& Template Protection& rate and high attack& Unconstrained &CASIA v4 Distance Dataset & (Equivalent to\\
			& & resistances& &UBIRIS v1, UBIRIS v2 & Unprotected domain)\\\hline
		\end{tabular}
	}
\end{table*}

With the proven statistical analysis and it's impeccable accuracy, iris recognition has seen large scale deployment in various secure access control applications \cite{daugman2009iris, daugman2016information}. The large scale deployment of iris recognition has in turn resulted in systems that capture iris patterns from same set of people across various services and sectors ranging from private entity operations to government controlled border crossing (e.g., passenger management in Schipol, Netherlands and immigration control in United Arab Emirates), civilian ID management (e.g., Aadhar card management in India). Unlike the passwords, when the biometric data in plain form (e.g., iris images) is stolen, it cannot be replaced due to inherent limitation of any person having limited amount of biometric characteristics. This specific problem has led the biometrics research towards a new direction where approaches were proposed to store the features from biometric modalities rather than in plain image domain. Further, to avoid the misuse of database compromises, it was also proposed to protect the features extracted from the biometric modalities. An implicit requirement in biometric systems is therefore to compare the biometric features in protected domain in the modern day systems. 

Considering such a demand within biometric systems, ISO/IEC JTC1 SC27 committee \cite{ISO24745} has standardized the need to protect the biometric features under \textsl{Biometric Template Protection} \cite{BTPReview, rathgeb2013alignment, ratha2001enhancing,rathgeb2015towards,rathgeb2015towardscancelable, gomez2018multi,hermans2014bloom}. This standard is further aligned to the newer guidelines from the European General Data Protection Regulations (EU-GDPR) \cite{regulation2018gdpr} which demands the strict need for privacy preservation and data protection. The three fundamental requirements of template protection respecting the ISO standards and GDPR are \textsl{irreversibility}, \textsl{unlinkability} and \textsl{revokability} which are briefly discussed below. The concept of \textsl{irreversibility} is to enforce that the biometric features in the protected domain will not lead to reconstruction of biometric sample that can lead to either direct or indirect association with a subject. Number of works have underlined this need by demonstrating ability to reconstruct biometric samples when the features are not stored in protected manner for iris \cite{galbally2012iriscode}, face \cite{juefei2016learning} and fingerprint \cite{feng2010fingerprint}. Secondly, the \textsl{unlinkability} ensures that any subject using two different services with same biometric biometric modality should not be identified by linking the protected features. The specific challenge of linking of biometric templates across two services compromises the integrity of biometric systems as shown in recent work \cite{hermans2014bloom}. Thirdly, the concept of \textsl{revokability} ensures the mitigation measure when the biometric systems are compromised. It is therefore required that the template protection scheme can revoke and replace the protected templates if a such a need should arise. Apart from the three regulations, it is also needed to ensure that the performance of biometric system is not not degraded due to template protection mechanism itself. 

Motivated by such factors, a number of works have been reported in the recent past for achieving biometric template protection for various modalities\cite{BTPReview, rathgeb2013alignment, ratha2001enhancing,rathgeb2015towards,rathgeb2015towardscancelable, gomez2018multi}. Given the focus of this work, we limit ourselves to template protection schemes for iris recognition. We first note a number of template protection schemes proposed for iris recognition considering the wide scale deployment \cite{pillai2011secure,rathgeb2013alignment, rathgeb2015towards,rathgeb2015towardscancelable, gomez2018multi} and then briefly review the existing template protection schemes for the iris recognition. Subsequently, we identify the set of unsolved challenges for template protection within iris recognition in the section below.

\subsection{Related Works}
A brief overview of the state-of-art template protection schemes proposed in the recent works is first reviewed in this section. As it can be noted from the Table~\ref{tab:sota-iris-template}, most of the works on the iris template protection are focused on the Near-Infra-Red (NIR) domain and further the data employed for validating the previously proposed template protection corresponds to constrained capture setting  (a summary of state-of-art works are presented in the Appendix of this article). While noting these two factors, we also note that the accuracy of most of the proposed approaches are very high as a direct consequence of data stemming from constrained setting. In another direction, we make another observation that a number of recent works have been inspired by the recently proposed Bloom-Filter based template protection scheme\cite{rathgeb2013alignment,rathgeb2015towards,rathgeb2015towardscancelable, gomez2018multi}. Given the wide popularity of the Bloom-Filter based template protection schemes, we identify the key limitations of the Bloom-Filter based template protection, especially in scaling up to unconstrained iris template protection where a higher false accepts and false rejects are noted. Secondly, the previously proposed approaches have limited the validation to iris images captured in NIR spectrum and no work has been reported in Visible Spectrum (VIS) iris recognition. In an effort to address such limitations, we present a new framework for template protection which not only is able to scale up to unconstrained iris images, but also across capture spectrum. 

\subsection{Challenges and Our Contributions}
\begin{figure}[htp]
	\centering
	\includegraphics[width=0.95\linewidth]{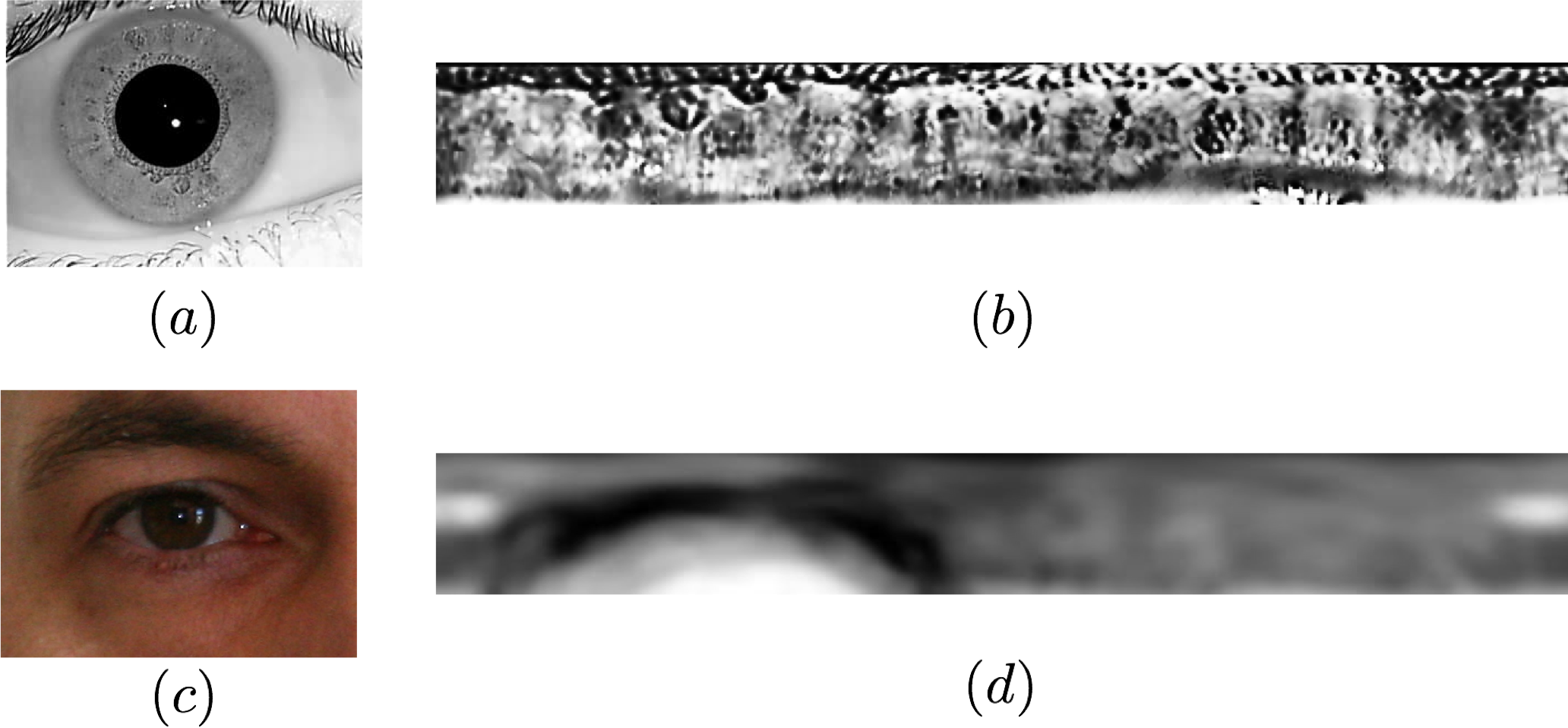}
	\caption{The degradation of iris quality from constrained capture to iris-on-move capture. (a) presents the iris captured in constrained setting as provided in IITD v1 Iris Dataset \cite{kumar2010comparison} and (b) presents the iris captured from on-the-move scenario as provided by UBIRIS Dataset \cite{proenca2009ubiris}. (b) and (d) represent the corresponding segmented and normalized iris image for (a) and (c).}
	\label{fig:iris-quality-degradtion}
\end{figure}

From the number of works listed above it can be noted that most of the works focus on constrained iris data captured in close cooperation. While the practical iris recognition systems need to operate at within a stipulated time, they often relax the constraints for the capture. Under such relaxed capture conditions, the iris recognition suffers from number of quality degrading factors such as motion blur, reflection from ambient light, reflection of eye-lashes on iris and partial iris capture due to partial closure of lids \cite{hollingsworth2009best, proencca2015iris,daugman2009iris, daugman2016information}. These factors are further aggravated in the iris-on-the move systems where not just the quality is impacted, but also the details of the captured iris by itself is substantially low. As observed from the Figure~\ref{fig:iris-quality-degradtion}, the capture in unconstrained settings results in quality par-below the one captured in constrained and fully-cooperative scenario. Secondly, the capture of iris images in Near-Infra-Red (NIR) results in superior iris features while the capture in visible spectrum results in iris features that are often with even lower iris pattern details. 

Such inherent challenges arising out of capture problems pose challenge in obtaining a reliable representation of iris codes subsequent to feature extraction (for instance, 2D/1D Gabor features). The direct impact of such inferior quality iriscodes can be witnessed through low performance reported in many earlier works \cite{hollingsworth2009best, proencca2015iris,daugman2009iris, daugman2016information}. A number of strategies have been proposed in earlier works to handle the problems of inferior quality iris codes to improve the recognition performance \cite{hollingsworth2009best, proencca2015iris,daugman2009iris, daugman2016information}. The sub-optimal quality iris data can impact not only the recognition accuracy but also the subsequent operations based on iris code, specifically iris template protection. This specific aspect of degraded performance of template protection schemes with Bloom-Filter due to unconstrained iris capture and inferior quality iriscodes was noted and illustrated in our recent work\cite{kiran2019morton}.

With a detailed analysis of Bloom-filter based template protection for iris recognition in unconstrained setting, we established the limitations of classical Bloom-filter based template protection in scaling for unconstrained setting where the data is significantly noisy. As it can be noted from the Figure~\ref{fig:failure-analysis-bloom}, the iriscodes in unconstrained iris capture from CASIA.v4 distance dataset \cite{casia-v4-db} results in unreliable iriscodes that differ for the same subject across captures. This implicitly impacts the Bloom Filter based template protection where similar locations are set in the protected templates leading to high number of false accepts. High number of false accepts therefore defeats the purpose of high security in iris biometrics systems. 

Driven by such problem, specifically for creation of protected templates even under noisy representation, we present a new approach employing the recently proposed \textit{Morton Filters} \cite{breslow2018morton}. Morton Filters introduces several key improvements to currently well employed \textit{Bloom-Filters} simply by creating multiple buckets with a predetermined logic. With such an architecture, Morton Filter approach supports compressed format that permits a sparse template that can be stored compactly in memory. Further, the multi-bucket architecture of Morton Filters reduces the False Accepts and False Rejects considerably over the traditional Bloom Filters with minimal computational overhead. Motivated by the architecture facilitating such improvements over Bloom Filters \cite{bender2018bloom, breslow2018morton}, we propose a new protected template creation mechanism using the Morton Filter approach on iriscodes.

In this version of our work, we extend the Morton Filter based iris template protection by specifically modelling inter-class and intra-class distribution of iris codes which is known to provide well separated comparison scores following statistical distribution motivated by earlier works \cite{daugman2016information,daugman2009iris}. The key motivation is to explore class distribution to make the template protection roust for unconstrained iris capture which typically suffers performance degradation in general iris recognition \cite{dong2011iris, tan2013adaptive, hollingsworth2009best, proencca2015iris, dong2011iris, hu2017optimal}. We specifically exploit the inter-class and intra-class distribution to extract robust iriscodes to the benefit of template protection such that multiple buckets can be easily composed. Such buckets facilitate optimal template creation through Morton Filter principles. To this extent, we employ low rank iriscodes that correspond to relatively non-noisy iriscodes, discriminable codes that differ from iriscodes of other subjects and a combination code using both representation of iriscodes. 

Our initial assertion of such an idea was validated in our earlier work\cite{kiran2019morton} where the biometric performance was significantly improved by optimizing both the false accepts and false rejects simultaneously.  While noting the previous works limiting to constrained iris data \cite{rathgeb2015towards,rathgeb2015towardscancelable, gomez2018multi,hermans2014bloom}, we validated the approach slightly unconstrained data \cite{kiran2019morton}. In this work, we take a step further to extend the approach to truly unconstrained iris recognition. Further, the approach is validated on the visible spectrum iris recognition through an evaluation on large scale public visible spectrum iris database. The key contributions of this extended work therefore are listed as below: 
\begin{itemize}[noitemsep]
	\item Proposes a new approach for template protection of iris codes using \textit{Morton Filters} in a multi-bucket approach exploiting various stable bits and discriminable bits within the iriscodes\cite{kiran2019morton}.
	\item Presents the key idea behind modelling the intra-class and inter-class distribution to the advantage of biometric template protection along with the theoretical background.
	\item An extensive analysis of the proposed approach is presented to validate the scalability of proposed approach by employing both constrained and unconstrained iris databases. Further, the approach is analyzed on both NIR spectrum and VIS spectrum iris recognition. To the best of our knowledge, this is the first work attempting to study the template protection scheme on large scale unconstrained iris database in both VIS and NIR spectrum.
	\item Additionally, the security analysis using unlinkability framework is provided to validate the applicability of proposed template protection scheme while benchmarking it with the Bloom-Filter based template protection scheme.
\end{itemize}

In the remainder of this paper, we present the principles and theory of Morton Filters in Section~\ref{sec:morton-filter}.  Section~\ref{sec:proposed-approach} presents the approach of template protection using Morton Filters mechanism. Section~\ref{sec:experiments-results} provides the experimental results along with the details of databases employed for evaluation. Section \ref{sec:unlinkability-analysis} discusses the security analysis for linkability issues followed by the conclusions and potential future works in Section \ref{sec:conclusions}.

\begin{figure*}[b]
	\centering
	\includegraphics[width=0.85\linewidth]{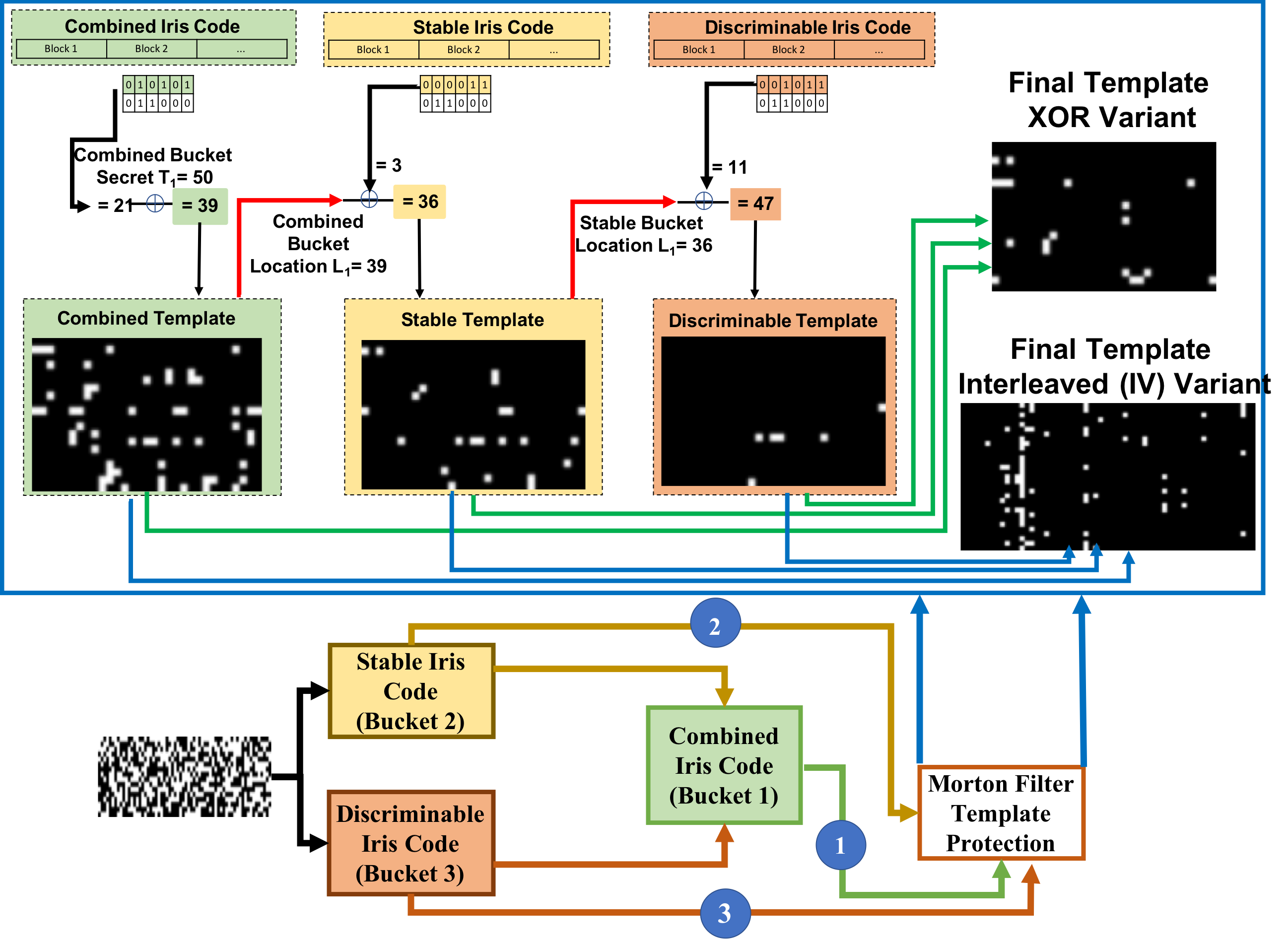} 
	\caption{Morton Filtering based protected template creation.}
	\label{fig:morton-filter-framework}
\end{figure*}


\section{Morton Filters}
\label{sec:morton-filter}
Morton Filters were originally proposed for the Approximate Set Membership Data Structures (ASMDSs)  in the field of computing to make the storage efficient\footnote{An ASMDS like a set data structure answers set membership queries (i.e., is an item $e$ an element of the set $S$?)} \cite{breslow2018morton}. Specifically, Morton Filters were designed to facilitate the lookups, insertions, and deletions unlike the Bloom-Filters which do not allow the dynamic changes. The key improvements come from the introduction of compressed format permitting logically sparse filter and leveraging metadata to prune unnecessary memory accesses. As a third major improvement over the Bloom-Filters, Morton-Filters heavily bias insertions through the use of a single hash function for primary bucket while allowing multiple buckets. As it can be deduced, Bloom-Filters set the same bit over and again for multiple various entries due to inherent design limitation of using single bucket operation. A significant drawback of this is that it does not allow efficient querying of false negatives due to absence of locality of reference\cite{breslow2018morton}. Although, this can be handled by adding extra number of hash functions, at a particular point the hash functions by themselves will overshadow actual length of original data or have high collision rate when few hashes are employed. Another alternative is to move the set bits to a different location based on the empty slots by constant look-up. While the former strategy can reduce false rejects, the later strategy can result in high number of false accepts both of which are not ideal in any operational scenario. This being the primary reasons, Morton-Filters formulated the multi-bucket approach to handle the problem efficiently. Through realization of multiple buckets set membership can be queried effectively leading to lower false negatives and false positives\cite{breslow2018morton}. 


Thus, with the paradigm of multiple buckets (say for instance $H_1, H_2 \ldots H_n$) within the Morton-Filters, the primary bucket $H_1$ is favoured heavily for insertions before proceeding with the rest of buckets. For negative lookups, the Morton-Filter employs an Over-flow Tracking Array (OTA), a simple bit vector that tracks when data cannot be placed within $H_1$ and moves to other available buckets. Negative lookups only require accessing a single bucket (i.e., OTA), in most cases even when the filter is heavily loaded. This unique architecture makes the lookup (positive, false positive, or negative) to access one bucket and at most $2$ leading to query efficiency of upto $50\%$\cite{breslow2018morton}.

In terms of implementation, Morton filters build upon the concepts of the Bloom Filters which operate by employing $k$ number of hash representations corresponding to number of blocks. The final representation $T$ in a Bloom-Filter of a predetermined size is first initialized to $0$.  For every chosen block within $n$ number of blocks, a particular location $x,y$ is set which corresponds to the final hashed representation template $T$.
\begin{align}
T(x,y) &= 1 \qquad if \ h^n_k= (y) \\
&\textrm{for a given column $x$ in chosen block $n$} 
\label{eqn:bloom-template}
\end{align}

While in the \textit{Morton Filters}, filling of each bucket relies on the fingerprint of previous hash value as denoted by Eqn.~\ref{eqn:bloom-template} and the output position within a new template given $T(x,y)$ is set to $1$. If the $T_1(x,y)$ is already set, another bit at a different location is set within a new template corresponding to $T_2(x, y)$. The process progresses for the number of designed buckets if all the bits in the previous buckets are already set. A detailed theory of the Morton-Filters is further provided in the original article \cite{breslow2018morton} and we limit at this point to diverge into the details of how Morton-Filters are employed for template protection in the upcoming sections.

\section{Morton Filter Iris Template Protection}
\label{sec:proposed-approach}
Intrigued by the design considerations of multi-bucket approach proposed in Morton-Filters to handle the false accepts and false rejects (or false positives and false negatives), we adapt the framework for template protection of iriscodes. The details of the template protection scheme based on Morton-Filters are presented in this section. 

\subsection{Morton Filters for Iris Template Protection}
\label{ssec:morton-filter-iris}
While the implementation of the Bloom-Filter based template protection for iris recognition is relatively straight forward, it has to be noted that Morton-Filter template protection differs in certain aspects. The core of Morton-Filters relies on having multiple buckets and in the very least case, more than one bucket is needed as discussed in previous sections. It can therefore be deduced that to make the template protection compatible to Morton-Filters, number of buckets need to be designed for iriscodes. A trivial idea for this can be to divide the iriscodes into blocks and thereafter consider them as separate buckets. While this may seem feasible at the first instance, it has to be noted that the iris imaging is impacted number of factors and thereby resulting in unreliable blocks for certain segments of the iris/iriscodes. Secondly, the iriscodes do not provide any correlational factors across different blocks due to the random structure of iris owing to biological factors\cite{daugman2001epigenetic}. Thus a more apprehensible manner of bucket formulations remains the first task. Therefore, we first focus on principles for suitable bucket design in an effort to obtain optimal templates.  

\subsubsection{Bucket Creation for Iriscodes}
\label{sssec:bucket-creation}
Within various deployed iris recognition systems, it is commonly observed that iris images are captured in multiple attempts or as a stream of video. The general idea behind using  multiple captures or video of any particular iris is to obtain the non-noisy part of iriscodes to minimize the error in comparison. In an analogy within the signal (or image) representation, the non-noisy iriscodes lie within the subspace of the complete iriscodes consisting of both noisy and non-noisy parts. Thus, obtaining the subspace corresponding to non-noisy iriscode can provide us with at-least one bucket for the realization of Morton-Filters based template protection.

For $k$ capture attempts of an iris image, the iriscodes ($I$) consisting of noisy and non-noisy parts can be represented as:
\begin{equation}
\I = [\I_{1}, \I_{2}, \ldots \I_{k}]
\label{eqn:multiple-iris}
\end{equation}
where each iriscode is of dimension $x,y$ pixels. Given the task at hand is to obtain a non-noisy iriscodes from the complete iriscode, it can be represented as composite of non-noisy part and noisy part as below:
\begin{equation}
\I_m=\Sm_m+\E_m
\label{eqn:sparse-representation}
\end{equation}
where $\Sm_m$ is low-rank non-noisy part and $\E_m$ is sparse error part within the iriscode corresponding to noisy data. As the capture conditions can vary across multiple captures, it can be easily deduced that both non-noisy part and the noisy part of iris code can vary in spatial locations. Thus, to obtain a stable non-noisy subspace of iriscode, one can explore multiple approaches such as obtaining Principal Component Analysis (PCA) or a median weighted approach \cite{meng2013cyclic, hu2015exploiting}. Given the recent formulation for obtaining the superior non-noisy subspace using tensor formulation from our recent work\cite{kiran2019stablebits}, we employ the same in this work.

Thus, by stacking the iriscodes obtained from multiple ($k$) capture attempts, a tensor of iriscodes can be represented as $\I\in\mathbb{R}^{x\times y \times k}$. The tensor formulation thus leads to easy recovery within the tensor space which corresponds to common bits across iriscodes across capture attempts. In an alternative interpretation, the solution to low-rank recovery of tensor space provides the iriscodes which are relatively non-noisy. Recovering non-noisy iriscodes from the tensor space thus translates to obtaining $\Sss_m$, low-rank non-noisy component and $\E_m$, sparse error component (i.e., low tubal rank component \cite{kilmer2011factorization}) from set of noisy iriscodes represented as $\I=\Sss_m+\E_m\in\mathbb{R}^{x \times y \times k}$:
\begin{align}
\min_{\Sss,\ \E} \ \norm{\Sss}_*+\lambda\norm{\E}_1, \ \st \ \I=\Sss+\E,
\label{eqn:trpca}
\end{align}
where $\norm{\Sss}_*$ is the tensor nuclear norm. However, the challenge of obtaining a sensible solution still remains in Eqn.~\ref{eqn:trpca} and in order to address this challenge, we express the $\lambda=1/\sqrt{\max(x,y) \times k}$ \cite{kilmer2011factorization} such that non-noisy subspace of iriscodes can be obtained. Given the formulation in Eqn.~\ref{eqn:trpca} and a reasonable expression of $\lambda$, we employ the software package provided by \cite{lu2018unified} to solve the Eqn.\ref{eqn:trpca} with no additional parameter tuning. Thereby, with the obtained non-noisy subspace, we derive one bucket ($\B^1_k$) of $k^{th}$ iriscode for a particular subject by simply using it as a weight map as given below:
 \begin{equation}
 \B^1_k=I_k*\Sss
 \label{eqn:bucket-stable}
 \end{equation}

While the first bucket is created from the above steps, the Morton-Filter architecture needs at-least another additional bucket. Thus, we explore the intra-class discriminablity of iriscodes exploring the findings from statistical trials provided in earlier works\cite{daugman2009iris}. Under the assumption that binomial distribution of iriscodes, all the bits within the iriscodes corresponding to $0$ and $1$ are equi-probable and randomly distributed\cite{daugman2009iris}. Thus, if the probability of $i^{th}$ bit equalling to $1$ is given by $p^*_i$ and the probability of $i^{th}$ bit equalling to $0$ is given by $q^*_i$, it can be safely written that $p^*_i + q^*_i = 1$ for the $i^{th}$ bit within an iriscode. Thus, discriminablity of a particular bit $d_i$ for a particular iriscode from the rest of the iriscodes can be formulated as:
\begin{equation}
d_{i}=p_{i}q_{i}^{*}+q_{i}p_{i}^{*} 
 \label{eqn:discriminability-bit}
\end{equation}
where $p_i$ is the probability of $i^{th}$ bit equalling to $1$ and $q_i$ is the probability of $i^{th}$ bit equalling to $0$ for any other iriscode \cite{hu2015exploiting} in an ideal case. Expressing, $q^*_i = 1 - p^*_i$, the Eqn~\ref{eqn:discriminability-bit} becomes
\begin{equation}
d_{i} = (1-2p_{i})p_{i}^{*} + p_{i}^{*} 
\label{eqn:discriminability-bit-user}
\end{equation}
It can therefore be noted that if the $p_{i} = 0.5$, the $d_{i} = 0.5$ and the implication is that intra-class value of the $i^{th}$ bit is highly uncertain being equi-probable. An alternative formulation leads to the fact discriminablity $d_{i}$ of one subject will be lower when the $i^{th}$ bit of rest of the subjects are more likely to be $1$. Given the number of iriscodes for a user available under multiple capture, the Eqn~\ref{eqn:discriminability-bit} can be used to derive discriminablity for each subject. As the discriminable maps can be estimated for different $k$ iriscodes, a stable discriminable map $D$ for a particular subject can be obtained by a normalized average of $n$ iriscodes.

\begin{equation*}
D_k = \frac{d_k-\min\{d_{k=1 \ldots n}\}}{\max\{d_{k=1 \ldots n}\}-\min\{d_{k=1 \ldots n}\}}
\end{equation*}

The second bucket $\B^2_k$ for an iriscode $k$ for a particular subject based on the discriminable map can therefore be created by simply weighting it with the iriscode:
\begin{equation}
\B^2_k=I_k*D_k
\label{eqn:bucket-discriminable}
\end{equation}

\subsection{Morton-Filter Template Protection Framework for Iris Recognition}
\label{ssec:proposed-approach-iris}
As discussed in the previous sections, Morton-Filter architecture can be used with at-least two buckets. However, given the limited number of hash functions employed to activate the bits in the protected template, noise factors that may occasionally impact the unreliability of the stable bits (despite bit selection) and discriminablity of the bits across iriscodes, we propose a third and auxiliary bucket that can be derived from combining both the buckets (Refer Eqn.~\ref{eqn:bucket-stable} and Eqn.~\ref{eqn:bucket-discriminable}). Thus,  bucket $B^3_k$ from an iriscode $k$ can be expressed as:
\begin{equation}
\B^3_k = I_k*D
\label{eqn:bucket-combined}
\end{equation}
While the designed buckets should be heavily activated in accordance to the principles of Morton-Filters, the second bucket or subsequent buckets are sparsely activated. Thus, we consider a different order by making the combined map iriscode as the biasing bucket (or first bucket i.e., ($\B^3_k$)) under the fact that a portion of bits commonly present in stability and discriminablity bucket can be activated simultaneously. Our assertion is that such a change of order helps in minimizing the false accepts and false rejects. Further, we activate bits in other buckets (stable and discriminable bits) in a regulated manner by employing a bijective XOR operation (indicated by $\oplus$). Thus, the sequence of the operations in template creation for an iriscode $k$ can be listed as below:
\begin{align} 
T^1_k &= 0 \qquad //Template \ Stable \ IC \ \nonumber \\  
T^2_k &= 0 \qquad //Template \ Discriminable \ IC\nonumber\\  
T^3_k &= 0 \qquad //Template \ Combined \ IC\nonumber\\ 
i^3_k &= Hash(B^3_k(i)) \nonumber\\ 
i^2_k &= \left( i^3_k \oplus Hash(B^2_k(i)) \right) \nonumber\\
i^1_k &= \left( i^2_k \oplus Hash(B^1_k(i)) \right) \nonumber\\
T^3_k &= 1 \ \text{if $i^3_k==1$}\nonumber\\
T^2_k &= 1 \ \text{iff $i^2_k==1 \ \&\& \ T^3_k==1$}\nonumber\\ 
T^1_k &= 1 \ \text{iff $i^1_k==1 \ \&\& \ T^3_k==1 \ \&\& \ T^2_k==1$}
\label{eqn:final-morton-template}
\end{align}
where $i_1$ and $i_2$ are indices to be set in two separate buckets.

The complete framework for Morton-Filter template based protected template creation is depicted in the Figure~\ref{fig:morton-filter-framework}. As with a regular iris recognition system, the framework begins with the normalized iris image which is further employed to extract the iriscode. For the sake of simplicity and backward compatibility, we employ the 1D Log Gabor representation owing to the fact that many early works on iris template protection have employed the same. While the architecture can improve the performance of the template protection, we note that this may suffer from the same challenge of linkability issues as indicated earlier \cite{hermans2014bloom}. In order to mitigate any such potential linkability issues, we adapt the private keys for creation of protected templates as described in recent work \cite{gomez2018multi,bringer2017security} along with a bijective function on all the three buckets such that the unique bits within the iris template is retained. 

As noted from Eqn~\ref{eqn:final-morton-template}, the bits in the final template are activated based on bits of other buckets. Such an architecture not only results in robust templates, but also makes the guessability attacks harder if not fully eliminate given only sparse number of bits are activated across the protected iris template. As it can be seen from the Figure~\ref{fig:morton-filter-framework}, the protected template is created by iteratively checking the bit location indicated by hash function and set if it is empty. If not, the hash value is carried forward to next bucket by determining the location based on the values in block under consideration and the previously obtained location for earlier bucket. It can be observed that fewer number of bits are set in the last bucket while more number of bits are set in the first bucket from Figure \ref{fig:morton-filter-framework}. 

\begin{figure}[htp]
	\centering
	\resizebox{0.85\linewidth}{!}{
		\begin{tabular}{ccc}
			\includegraphics[width=0.33\textwidth]{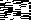} & 
			\includegraphics[width=0.33\textwidth]{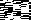} & 
			\includegraphics[width=0.33\textwidth]{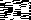} \\\\
			\color{black} \Huge\textbf{Combined Map} & \color{black} \Huge\textbf{Stable Map}& \color{black} \Huge\textbf{Discriminable Map}\\ 
			\\
			\includegraphics[width=0.33\textwidth]{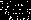} & \includegraphics[width=0.33\textwidth]{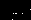}& \includegraphics[width=0.33\textwidth]{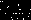} \\\\
			\color{black} \Huge\textbf{Combined Template} & \Huge\color{black} \Huge\textbf{Stable Template}& \color{black} \Huge\textbf{Discriminable Template}\\
			\\
			&
			\includegraphics[width=0.33\textwidth]{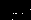} & \includegraphics[width=0.41\textwidth]{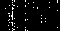}\\\\
			\color{black} \textbf{} & \color{black} \Huge\textbf{Proposed - XOR}& \color{black} \Huge\textbf{Proposed - Interleaved (IV)}\\
		\end{tabular}
	}
	\caption{Multiple templates of iris code using low rank non-noisy representation, discriminative representation,  and combined representation of the same binarized iris code (partial iris code) for a sample iris from IITD Iris Dataset \cite{zhao2015accurate}}
	\label{fig:proposed-template}
\end{figure}

\subsubsection{Two Variants of Template Protection Scheme}
\label{sssec:proposed-approach-iris-variants}
As observed from the Eqn~\ref{eqn:final-morton-template}, it can be deduced that the final protected template can be obtained from $T^1_k, T^2_k, T^3_k$, one can think of using all the three templates. Thus, the first variant we propose is the Interleaved Variant (IV) of all the three independent templates for an iriscode $k$ as indicated below:
\begin{equation}
T^{IV}_k = T^1_k || T^2_k || T^3_k
\label{eqn:interleaved-template}
\end{equation}
In order to diffuse the arrangement of bits, we further introduce a random ordered interleaving within the above Eqn~\ref{eqn:interleaved-template} to avoid any correlation based guessability attacks.

In the second variant, we simply employ a bijective XOR function to eliminate the bits within the protected template which are not common across all the three individual protected templates. The operation can therefore be given by:
\begin{equation}
T^{XOR}_k = T^1_k \oplus T^2_k \oplus T^3_k
\label{eqn:xor-template}
\end{equation}

The two variants of the proposed template protection can be further seen in the illustrated Figure~\ref{fig:proposed-template}. As observed, the XOR variant of the template protection provides the template of the size that is equal to the three independent templates resulting in a compact protected template size. As a second observation, it can be evidently seen that the size of the template triples for the IV version of the protected template. As a direct implication of this, one can deduce the high performance of the IV version as compared to it's XOR variant.

\section{Experiments and Results}
\label{sec:experiments-results}
We present the experimental evaluation on four different datasets and the corresponding results along with an analysis of the results. The first set of experiments relate to constrained iris acquisition by employing IITD Iris Database version 1.0 \cite{kumar2010comparison} and the second set of experiments relate to unconstrained iris acquisition using CASIA.v4 distance dataset \cite{casia-v4-db}. While the former is captured in highly cooperative setting, the latter is captured at various stand-off distance resulting in non-ideal iris images with significant degradation. Both of these datasets are captured in NIR spectrum mimicking the deployment iris systems. In another set of experiments, we employ the iris images captured in the visible spectrum by employing the UBIRIS v1\cite{proencca2005ubiris} and UBIRIS v2\cite{proenca2009ubiris} dataset. The key motivation behind such experiments on the unconstrained visible spectrum iris data is to evaluate scalability of the proposed approach for capture domain independence. Further, this set of experiments also establishes the robustness on degraded data stemming from unconstrained capture setting with significant degradations.

We further provide the comparison of performance from proposed approach against unprotected version and protected template using Bloom-Filters. For each of iriscode in unprotected domain and protected domain, we employ a simple Hamming Distance measure \cite{daugman2016information} to obtain the compare score.  

\subsection*{Performance Metrics}
\label{ssec:performance}
All the results in Detection Error Trade-off (DET) curves along with indication of Genuine Match Rate (GMR) where GMR is (1 - False Non Match Rate (FNMR)) at a False Match Rate of $0.01\%$. In addition to the DET graphs, we also present the results in Error Rate (EER\%) to indicate the symmetric error rates. 

\subsection{IITD Iris Database version 1.0}
\label{sec:database}

IITD Iris Database version 1.0 \cite{kumar2010comparison} is a constrained iris database consisting of data captured from $224$ subjects and $5$ images per iris. We employ the set of left iris images in the lines of earlier work and to provide fair comparison to earlier works \cite{rathgeb2015towardscancelable, bringer2017security}. Thus, in our work, we employ  data from $1120$ iriscodes from $224$ subjects with $5$ iris codes per subject. We employ 1D Log-Gabor encoding \cite{masek2003recognition} for the unprotected iris templates and the subsequently, use the same encoding to obtain obtain protected templates in the lines of earlier works on Bloom-filter based template protection \cite{rathgeb2013alignment,bringer2017security,gomez2018multi}. Further, it can be noted that the our approach does not heavily depend on feature space and thus allowing the freedom to employ any other binary encoding scheme for iris feature encoding. In order to consistent with the earlier works, we also present the comparison to earlier works\cite{bringer2017security, gomez2018multi}, we present the results in various size of block widths $\ell\in\{4, 8,16,32\}$.

\begin{table}[htpb]
	\centering
	\caption{Results of proposed template protection schemes compared to other schemes on IITD Iris Dataset.}
	\resizebox{0.48\textwidth}{!}{%
		\begin{tabular}{clcccc}
			\hline
			&       & \multicolumn{2}{c}{5 Bits} & \multicolumn{2}{c}{10 Bits} \bigstrut\\
			\hline
			& Iris code & EER   & GMR @ & EER   & GMR @ \bigstrut\\
			& &    & 0.01\% FMR &    & 0.01\% FMR \bigstrut\\
			\hline
			\hline
			Unprotected & Log-Gabor & 0,36  & 99,11 & 0,36  & 99,11 \bigstrut[t]\\
			\hline
			\hline
			\multicolumn{1}{c}{\multirow{10}[3]{*}{Protected}} & Bloom - 4 & 0,38  & 99,33 & 0,62  & 99,38 \\
			& Bloom - 8  & 0,39  & 99,38 & 0,44  & 99,55 \\
			& Bloom - 16 & 0,40  & 99,15 & 0,26  & 98,84 \\
			& Bloom - 32 & 0,83  & 98,57 & 0,34  & 98,08 \bigstrut[b]\\
			\hline
			& \multicolumn{5}{c}{Interleaved Variant (IV)} \bigstrut\\
			\cline{2-6}          & \textbf{Proposed -  4}     & \textbf{0.00}  & \textbf{100.00} & \textbf{0.00}  & \textbf{100.00} \bigstrut[t]\\
			& Proposed - 8     & 0.00  & 100.00 & 0.00  & 100.00 \\
			& Proposed - 16    & 0.00  & 100.00 & 0.00  & 100.00 \\
			& Proposed - 32    & 0.00  & 100.00 & 0.01  & 99.78 \\
			\cline{2-6}& \multicolumn{5}{c}{XOR Variant} \bigstrut\\
			\cline{2-6}          & \textbf{Proposed - 4} & \textbf{0,00}  & \textbf{100,00} & {5.89}  & {64.62} \bigstrut[t]\\
			& Proposed - 8 & 0.00  & 100.00 & 2.34  & 90.63 \\
			& Proposed - 16 & 0.00  & 100.00 & 1.42  & 94.98 \\
			& Proposed - 32 & 0.01  & 99.67 & \textbf{1.10}  & \textbf{96.32} \bigstrut[b]\\
			\hline
		\end{tabular}%
	}
	\label{tab:iitd-iris-performance}%
\end{table}%

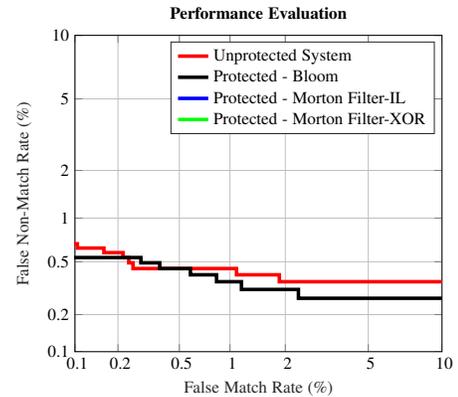
\begin{figure}[htp]
	\centering
	\resizebox{0.33\textwidth}{!}{%
%
%
\begin{tikzpicture}
\tikzset{lw/.style = {line width=4pt}}
\begin{axis}[%
width=0.4\textwidth,
at={(1.011in,0.642in)},
scale only axis,
xmin=-3.09023224677291,
xmax=-1.28155156321185,
xtick={-3.09023224677291,-2.8781616977673,-2.57582930094926,-2.32634787735637,-2.05374890849825,-1.64485362793551,-1.28155156321185},
xticklabels={{  0.1},{  0.2},{ 0.5},{  1},{  2},{  5},{  10}},
xlabel style={font=\color{white!15!black}},
xlabel={False Match Rate (\%)},
ymin=-3.09023224677291,
ymax=-1.28155156321185,
ytick={-3.09023224677291,-2.8781616977673,-2.57582930094926,-2.32634787735637,-2.05374890849825,-1.64485362793551,-1.28155156321185},
yticklabels={{  0.1},{  0.2},{ 0.5},{  1},{  2},{  5},{  10}},
ylabel style={font=\color{white!15!black}},
ylabel={False Non-Match Rate (\%)},
axis background/.style={fill=white},
title style={font=\bfseries},
title={Performance Evaluation},
xmajorgrids,
ymajorgrids,
legend style={legend cell align=left, align=left, draw=white!15!black}
]
\addplot [color=red, line width=2.0pt]
  table[row sep=crcr]{%
-1.28146031198599	-2.69010951320479\\
-2.08316982574463	-2.69010951320479\\
-2.08316982574463	-2.65056549662084\\
-2.29416697448759	-2.65056549662084\\
-2.29416697448759	-2.61477704977736\\
-2.80415634683369	-2.61477704977736\\
-2.80415634683369	-2.58205375205721\\
-2.8252390040505	-2.58205375205721\\
-2.8252390040505	-2.5518818104109\\
-2.85349273235747	-2.5518818104109\\
-2.85349273235747	-2.52386822672112\\
-2.94754556101977	-2.52386822672112\\
-2.94754556101977	-2.49770547621218\\
-3.07827896878026	-2.49770547621218\\
-3.07827896878026	-2.4731482564224\\
-3.10205147765763	-2.4731482564224\\
};
\addlegendentry{Unprotected System}

\addplot [color=black, line width=2.0pt]
  table[row sep=crcr]{%
-1.28146031198599	-2.78473531868344\\
-1.98914846999675	-2.78473531868344\\
-1.98914846999675	-2.73436876677459\\
-2.26975761948951	-2.73436876677459\\
-2.26975761948951	-2.69010951320479\\
-2.39245530285812	-2.69010951320479\\
-2.39245530285812	-2.65056549662084\\
-2.52132974877309	-2.65056549662084\\
-2.52132974877309	-2.61477704977736\\
-2.67257180106216	-2.61477704977736\\
-2.67257180106216	-2.58205375205721\\
-2.76539110187993	-2.58205375205721\\
-2.76539110187993	-2.5518818104109\\
-3.10205147765763	-2.5518818104109\\
};
\addlegendentry{Protected - Bloom}

\addplot [color=blue, line width=2.0pt]
table[row sep=crcr]{%
	-3.09023224677291 -3.09023224677291\\
};
\addlegendentry{Protected - Morton Filter-IL}

\addplot [color=green, line width=2.0pt]
table[row sep=crcr]{%
	-3.09023224677291 -3.09023224677291\\
};
\addlegendentry{Protected - Morton Filter-XOR}

\end{axis}
\end{tikzpicture}%
	}
	\caption{Comparison of biometric performance using DET for IITD Iris Dataset. Proposed approach is depicted with $4$ blocks and $5$ bits alongside similar configuration with Bloom-filter template protection. It has to noted that the EER being close to $0$ for proposed approach, the curves do not appear on the DET curves.}
	\label{fig:iitd-iris-results-det}
	\vspace{-3mm}
\end{figure}

\begin{figure*}[htp]
	\centering
	\subfloat[Bloom-Filter][Proposed -Interleaved (IV)]{
		\resizebox{0.26\textwidth}{!}{%
%
\begin{tikzpicture}

\begin{axis}[%
width=4.589in,
height=4.589in,
at={(1.731in,0.8in)},
scale only axis,
xmin=-4.06711094425711,
xmax=-0.776584219466475,
xtick={-3.09023224677291,-2.8781616977673,-2.57582930094926,-2.32634787735637,-2.05374890849825,-1.64485362793551,-1.28155156321185,-0.841621234874822,-0.253347103317183},
xticklabels={{  0.1},{  0.2},{ 0.5},{  1},{  2},{  5},{  10},{  20},{  40}},
xlabel style={font=\color{white!15!black}},
xlabel={False Match Rate (\%)},
ymin=-3.77823333604081,
ymax=-0.487706611250172,
ytick={-3.09023224677291,-2.8781616977673,-2.57582930094926,-2.32634787735637,-2.05374890849825,-1.64485362793551,-1.28155156321185,-0.841621234874822,-0.253347103317183},
yticklabels={{  0.1},{  0.2},{ 0.5},{  1},{  2},{  5},{  10},{  20},{  40}},
ylabel style={font=\color{white!15!black}},
ylabel={False Non-Match Rate (\%)},
axis background/.style={fill=white},
title style={font=\bfseries},
title={Performance Evaluation},
xmajorgrids,
ymajorgrids,
legend style={legend cell align=left, align=left, draw=white!15!black}
]
\addplot [color=red, line width=2.0pt]
  table[row sep=crcr]{%
-2.49630371636701	-4.10728600851987\\
-2.49630371636701	-3.34832774472366\\
-2.91068090731931	-3.34832774472366\\
-2.91068090731931	-3.15119901292087\\
-2.93695030533083	-3.15119901292087\\
-2.93695030533083	-3.03080527854315\\
-2.99419276551973	-3.03080527854315\\
-2.99419276551973	-2.9428525156018\\
-3.00356848399547	-2.9428525156018\\
-3.00356848399547	-2.87306993638837\\
-3.03339624482218	-2.87306993638837\\
-3.03339624482218	-2.81497853074029\\
-3.04666600451247	-2.81497853074029\\
-3.04666600451247	-2.76507004886801\\
-3.10916439007026	-2.76507004886801\\
-3.10916439007026	-2.72122409350551\\
-3.1294909319071	-2.72122409350551\\
-3.1294909319071	-2.68205817368984\\
-3.21764547776804	-2.68205817368984\\
-3.21764547776804	-2.6466193302419\\
-3.46963879091774	-2.6466193302419\\
-3.46963879091774	-2.6142220406517\\
-3.68432513594859	-2.6142220406517\\
-3.68432513594859	-2.58435620614578\\
-3.76466839049332	-2.58435620614578\\
-3.76466839049332	-2.55663168921695\\
-3.83595460122114	-2.55663168921695\\
-3.83595460122114	-2.53074322899776\\
-4.09789654033338	-2.53074322899776\\
};
\addlegendentry{Morton Filter - 4}

\addplot [color=green, line width=2.0pt]
  table[row sep=crcr]{%
-2.07146229957703	-4.10728600851987\\
-2.07146229957703	-3.34832774472366\\
-2.32144825044051	-3.34832774472366\\
-2.32144825044051	-3.15119901292087\\
-2.34448112371621	-3.15119901292087\\
-2.34448112371621	-3.03080527854315\\
-2.40115196065845	-3.03080527854315\\
-2.40115196065845	-2.9428525156018\\
-2.43925615521278	-2.9428525156018\\
-2.43925615521278	-2.87306993638837\\
-2.53591663490351	-2.87306993638837\\
-2.53591663490351	-2.81497853074029\\
-2.57923213133475	-2.81497853074029\\
-2.57923213133475	-2.76507004886801\\
-2.59029661291597	-2.76507004886801\\
-2.59029661291597	-2.72122409350551\\
-2.67264710645601	-2.72122409350551\\
-2.67264710645601	-2.68205817368984\\
-2.75331440628726	-2.68205817368984\\
-2.75331440628726	-2.6466193302419\\
-2.78471278608514	-2.6466193302419\\
-2.78471278608514	-2.6142220406517\\
-2.85568932285753	-2.6142220406517\\
-2.85568932285753	-2.58435620614578\\
-2.86190992014122	-2.58435620614578\\
-2.86190992014122	-2.55663168921695\\
-2.93114866691941	-2.55663168921695\\
-2.93114866691941	-2.53074322899776\\
-2.9428525156018	-2.53074322899776\\
-2.9428525156018	-2.50644734792178\\
-2.9696872900142	-2.50644734792178\\
-2.9696872900142	-2.48354663673005\\
-3.04938549247553	-2.48354663673005\\
-3.04938549247553	-2.46187875325004\\
-3.09315528179961	-2.46187875325004\\
-3.09315528179961	-2.44130853083116\\
-3.09946389093134	-2.44130853083116\\
-3.09946389093134	-2.42172219589956\\
-3.10916439007026	-2.42172219589956\\
-3.10916439007026	-2.40302305109544\\
-3.11916659112157	-2.40302305109544\\
-3.11916659112157	-2.36796601725059\\
-3.1587775266915	-2.36796601725059\\
-3.1587775266915	-2.35147419069502\\
-3.1785587934434	-2.35147419069502\\
-3.1785587934434	-2.33559815306794\\
-3.20406607503732	-2.33559815306794\\
-3.20406607503732	-2.32028984532648\\
-3.22703950642694	-2.32028984532648\\
-3.22703950642694	-2.29121089122826\\
-3.24168740984726	-2.29121089122826\\
-3.24168740984726	-2.26394926114673\\
-3.2845450155626	-2.26394926114673\\
-3.2845450155626	-2.22596896796151\\
-3.3023063772912	-2.22596896796151\\
-3.3023063772912	-2.21399288721743\\
-3.42035675117058	-2.21399288721743\\
-3.42035675117058	-2.20232617724878\\
-3.45908933184439	-2.20232617724878\\
-3.45908933184439	-2.17985390515379\\
-3.46963879091774	-2.17985390515379\\
-3.46963879091774	-2.16901821791134\\
-3.62242993401636	-2.16901821791134\\
-3.62242993401636	-2.15843136960871\\
-3.68432513594859	-2.15843136960871\\
-3.68432513594859	-2.1379558674284\\
-3.73513216551748	-2.1379558674284\\
-3.73513216551748	-2.12804523313229\\
-3.88052518228662	-2.12804523313229\\
-3.88052518228662	-2.11833931177746\\
-4.09789654033338	-2.11833931177746\\
};
\addlegendentry{Morton Filter - 8}

\addplot [color=blue, line width=2.0pt]
  table[row sep=crcr]{%
-3.29626971091532	-4.10728600851987\\
-3.29626971091532	-3.34832774472366\\
-3.88052518228662	-3.34832774472366\\
-3.88052518228662	-3.03080527854315\\
-4.09789654033338	-3.03080527854315\\
};
\addlegendentry{Morton Filter - 16}

\addplot [color=black, line width=2.0pt]
  table[row sep=crcr]{%
-3.25706572150305	-4.10728600851987\\
-3.25706572150305	-3.34832774472366\\
-3.52913091836441	-3.34832774472366\\
-3.52913091836441	-3.15119901292087\\
-3.70853359601769	-3.15119901292087\\
-3.70853359601769	-3.03080527854315\\
-3.83595460122114	-3.03080527854315\\
-3.83595460122114	-2.9428525156018\\
-3.88052518228662	-2.9428525156018\\
-3.88052518228662	-2.87306993638837\\
-4.09789654033338	-2.87306993638837\\
};
\addlegendentry{Morton Filter - 32}

\end{axis}

\begin{axis}[%
width=7.778in,
height=5.833in,
at={(0in,0in)},
scale only axis,
xmin=0,
xmax=1,
ymin=0,
ymax=1,
axis line style={draw=none},
ticks=none,
axis x line*=bottom,
axis y line*=left,
legend style={legend cell align=left, align=left, draw=white!15!black}
]
\end{axis}
\end{tikzpicture}%
		}
		\label{fig:casia-interleaved}}
	\qquad
	\subfloat[Proposed][Proposed - XOR]{
		\resizebox{0.26\textwidth}{!}{%
			\input{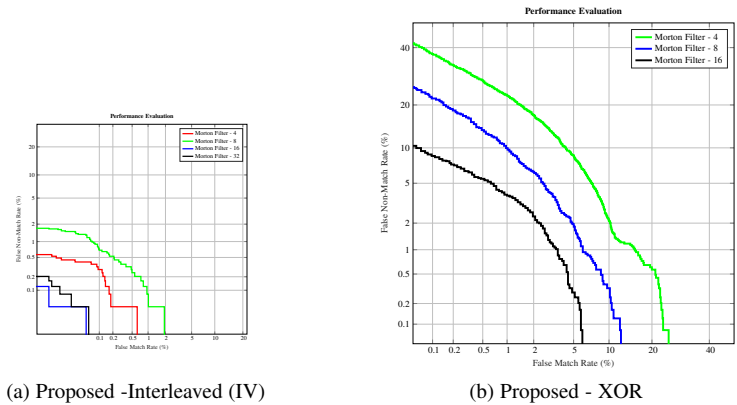}
		}
		\label{fig:casia-xored}}
	\caption{Performance of multiple configurations for proposed template protection on CASIA v4 Distance dataset.}
	\label{fig:casia-iris-multiple-config}
	\vspace{-3mm}
\end{figure*}

\subsection{Results on IITD Iris Dataset}
The empirical results of the proposed template protection scheme along with the comparison to unprotected templates and protected templates using Bloom Filter approach are presented in Table \ref{tab:iitd-iris-performance}. From the Table \ref{tab:iitd-iris-performance}, the following observations can be made on the proposed approach:
\begin{itemize}
	\item The proposed approach is antagonistic to block size unlike the approach based on Bloom Filter which is fairly robust in smaller block sizes and degrades in performance with increasing block size. A similar observation for Bloom-Filter based template protection was reported in earlier work\cite{bringer2017security} which noted the degradation of biometric with increasing block width and higher bits within each blocks. 
	\item A near ideal performance from proposed approach with IV configuration is obtained due to the fact that the diversity of the templates are high as the size of the template is thrice the size of simple Bloom-Filter. One can therefore employ other strategies such as bit compaction to obtain lower template size to retain the performance and decrease the size of final protected template.
	\item As anticipated, the compact size of XOR variant of the proposed approach results in moderate degradation in performance.  An introspection into this behaviour revealed the fact that the XOR operation cancels out most of the bits if they are not set across multiple individual templates created from Morton Filters. When such a operation a carried out, the collision of the bits across multiple protected templates starts to happen. The collision rate along with the low number of activated bits in final template jointly  degrade the performance in XOR variant.
\end{itemize}
The DET curves in Figure \ref{fig:iitd-iris-results-det} depicts the performance of proposed approach. Further, to illustrate the antagonistic nature of proposed approach towards various block sizes, the DET curves are presented in Appendix in Figure \ref{fig:iitd-iris-multiple-config}. As it can be noted, various configurations of the proposed approach for a bit length of $5$ provides near ideal performance for protected templates.

\subsection{CASIA V4 Distance Iris Dataset}
\label{ssec:casia-database}
Considering the earlier works focusing on the constrained iris recognition, we evaluate the proposed approach on unconstrained capture setting by employing  CASIA.v4 distance dataset \cite{casia-v4-db}. CASIA.v4 capture includes parts of the face image due to stand-off distance of $3 \ meters$ in the acquisition setting from a total of $142$ subjects. Further, it has to be noted that the captured iris images suffers from degradation unlike ideal iris images due to blinking, occlusion due to eye-lids, specular reflection, presence of eye-glasses and motion blur all leading to real-life iris acquisition. We therefore employ classical Viola-Jones eye detector to detect the eye region alone and then correct the errors manually for any undetected/wrongly detected eye regions \cite{viola2001rapid}.  Further to eye region localization, we segment and normalize the iris region using \cite{masek2003recognition} to derive the iris images of size $100 \times 360$ pixels. The iris codes are further extracted using $1D \ Log-Gabor$ features for the baseline evaluation \cite{masek2003recognition}.

\begin{table}[htbp]
	\centering
	\caption{Results from proposed approach on CASIA v4 Dataset}
	\resizebox{0.46\textwidth}{!}{%
		\begin{tabular}{lcccc}
			\hline
			& \multicolumn{2}{c}{5 Bits} & \multicolumn{2}{c}{10 Bits} \bigstrut\\
			\hline
			Configurations & EER   & GMR @   & EER   & GMR @\bigstrut\\
			& & 0.01\% FMR &    & 0.01\% FMR \bigstrut\\
			\hline
			& \multicolumn{4}{c}{Bloom Filter (Naive)} \bigstrut\\
			\hline
			Bloom - 4     & 36.24 & 0.65  & 38.52 & 0.20 \bigstrut[t]\\
			Bloom - 8     & 40.00 & 0.24  & 41.65 & 0.08 \\
			Bloom - 16    & 41.70 & 0.24  & 43.77 & 0.08 \\
			Bloom - 32    & 40.07 & 0.08  & 44.87 & 0.12 \\
			\hline
			& \multicolumn{4}{c}{Bloom Filter (Low-rank non-noisy/stable Iris Codes)} \bigstrut\\
			\hline
		    Bloom - 4  	  &27.12 & 29.27 & 31.87 & 14.63 \bigstrut[t]\\
		    Bloom - 8     &31.83 & 11.38 & 36.09 & 4.07 \\
		    Bloom - 16    &34.31 & 8.94  & 39.92 & 2.44 \\
		    Bloom - 32    &32.92 & 18.29 & 42.12 & 3.66 \\
			\hline
			& \multicolumn{4}{c}{Interleaved (IV)} \bigstrut\\
			\hline
			Proposed - 4     & {0.13}  & {99.51} & 0.53  & 96.63 \bigstrut[t]\\
			Proposed - 8     & 0.37  & 98.37 & 2.40  & 82.07 \\
			Proposed - 16    & \textbf{0.04}  & \textbf{99.96} & 1.96  & 86.87 \\
			Proposed - 32    & 0.04  & 99.88 & 1.92  & 88.66 \\
			\hline
			& \multicolumn{4}{c}{XOR} \bigstrut\\
			\hline
			Proposed - 4     &  \textbf{0.00}  & \textbf{100.00} & 0.00  & 100.00 \bigstrut[t]\\
			Proposed - 8     & 6.27  & 52.93 &  6.61  & 39.88 \\
			Proposed - 16    & 3.40  & 69.88 & 11.21 & 18.70 \\
			Proposed - 32    & 2.11  & 86.18 & 8.37  & 27.93 \\
			\hline
		\end{tabular}%
	}
	\label{tab:casia-results}%
\end{table}%

\begin{figure*}[htp]
	\centering
	\subfloat[Bloom-Filter][Proposed -Interleaved (IL)]{
		\resizebox{0.3\textwidth}{!}{%
			\input{pgftikz/det-interleaved-different-key1-btas2019-morton-4-32-Bits-5-ubiris-part1-template.mat.tex}
		}
		\label{fig:ubiris-interleaved}}
	\qquad
	\subfloat[Proposed][Proposed - XORed]{
		\resizebox{0.3\textwidth}{!}{%
			\input{pgftikz/det-xor-different-key1-btas2019-morton-4-32-Bits-5-ubiris-part1-template.mat.tex}
		}
		\label{fig:ubiris-xored}}
	\caption{Performance of multiple configurations for proposed template protection on UBIRIS v1 dataset.}
	\label{fig:ubiris-iris-multiple-config}
	\vspace{-4mm}
\end{figure*}

\begin{figure*}[b]
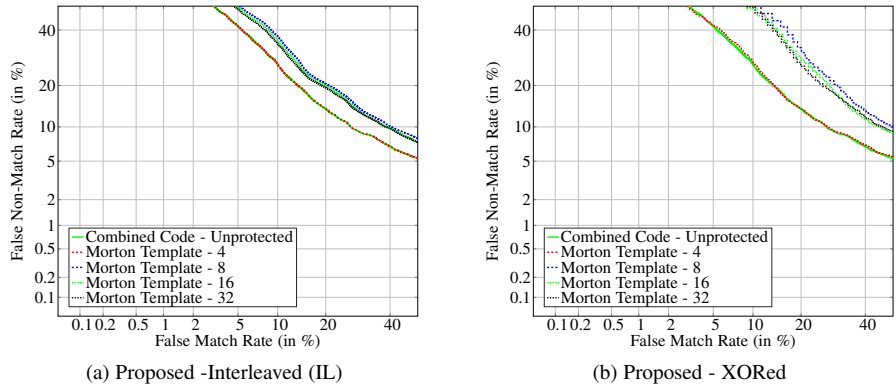

	\centering
	\subfloat[Bloom-Filter][Proposed -Interleaved (IL)]{
		\resizebox{0.3\textwidth}{!}{%
			\input{pgftikz/det-interleaved-different-key1-btas2019-morton-4-32-Bits-5-ubiris-part2-template.mat.tex}
		}
		\label{fig:ubiris-v2-interleaved}}
	\qquad
	\subfloat[Proposed][Proposed - XORed]{
		\resizebox{0.3\textwidth}{!}{%
			\input{pgftikz/det-xor-different-key1-btas2019-morton-4-32-Bits-5-ubiris-part2-template.mat.tex}
		}
		\label{fig:ubiris-v2-xored}}
	\caption{Performance of multiple configurations for proposed template protection on UBIRIS v2 dataset.}
	\label{fig:ubiris-v2-iris-multiple-config}
	\vspace{-4mm}
\end{figure*}
\subsection{Results on CASIA v4 Distance Iris Dataset}
Table \ref{tab:casia-results} presents the results of  proposed template protection along Bloom Filter approach. Noting the low performance of naive Bloom-Filter on CASIA v4 dataset, we also employ the stable bits to extract the Bloom-Filter template. Specifically, we employ the Eqn~\ref{eqn:bucket-stable} provided in the Section~\ref{sssec:bucket-creation} to extract the stable bits from iriscode. As it can be noted from Table \ref{tab:casia-results}, there is an marginal performance improvement over naive Bloom-Filter templates but much below the required operational performance in a practical biometric system. Further, in the lines of our experiments on IITD Iris dataset, we also present the results on both IV and XOR variant of the proposed approach on CASIA v4 dataset. The clear improvement of the proposed approach can be seen from the Table \ref{tab:casia-results} and impelled by such a improvement, we note the following points:

\begin{itemize}[noitemsep]
	\item While in the case of Bloom-Filter templates, same set of bits are activated due to noisy nature of the iriscodes. This in turn results in sub-optimal protected template results validating the motivation and initial assertion for this work. It can be noted from Fig. \ref{fig:failure-analysis-bloom}, that for the iris template across sessions for the same subject, bloom filter approach results in dissimilar protected templates for the same subject across different captures leading to a significant false rejects. 
	\item While in the proposed approach, the creation of multiple buckets and further activation of bits based on the predecessor buckets leads to a unique template even for different captures for a same subject. Adding the uniqueness of the template, both Interleaved variant (IV) makes the templates unique due to large size and the XOR variant eliminates the inconsistent bits. Both of these variants further lead to lower false rejects needing further investigation into design considerations of Morton-Filters for template protection. 
	\item Despite the degradations of iriscodes owing the unconstrained setting, it can be noted that the proposed approach on CASIA.v4 distance dataset \cite{casia-v4-db} is significantly stable in IV variant of proposed approach while the XOR variant produces the templates that have slightly higher collisions.
\end{itemize}
We further present the DET curves of proposed approach as shown in Figure~\ref{fig:casia-iris-multiple-config} for both variants and do not present the DET curves for the Bloom-Filter templates owing to the low performance. Supporting our initial assertion, the proposed approach through employing multiple buckets results in lower false accepts and false rejects as shown in Figure~\ref{fig:casia-iris-multiple-config}. 

\begin{figure*}[htp]
	\centering
	\subfloat[Bloom-Filter][XOR - 4 Blocks - 10 Bits]{
		\includegraphics[width=0.22\textwidth]{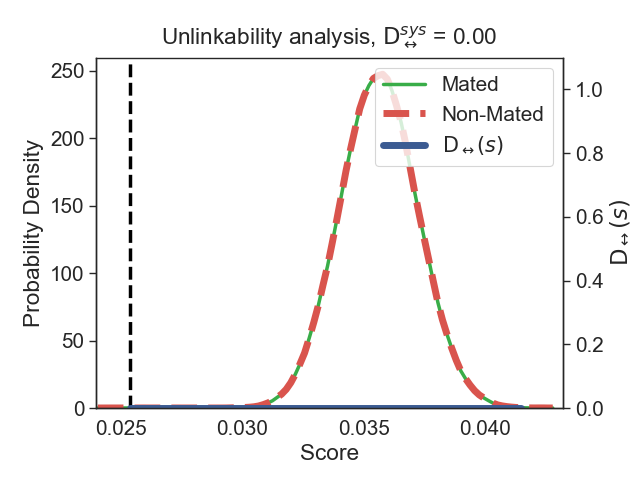}
		\label{fig:unlink-xored-1}}
	\qquad
	\subfloat[Proposed][XOR - 8 Blocks - 10 Bits]{
		\includegraphics[width=0.22\textwidth]{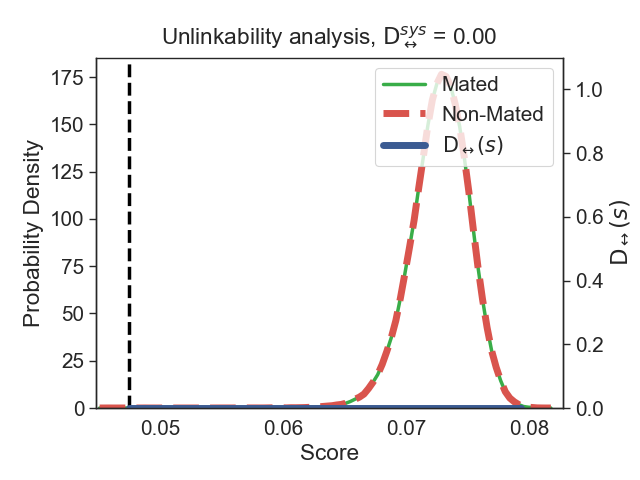}
		\label{fig:unlink-xored-2}}
	\subfloat[Proposed][XOR - 16 Blocks - 10 Bits]{
		\includegraphics[width=0.22\textwidth]{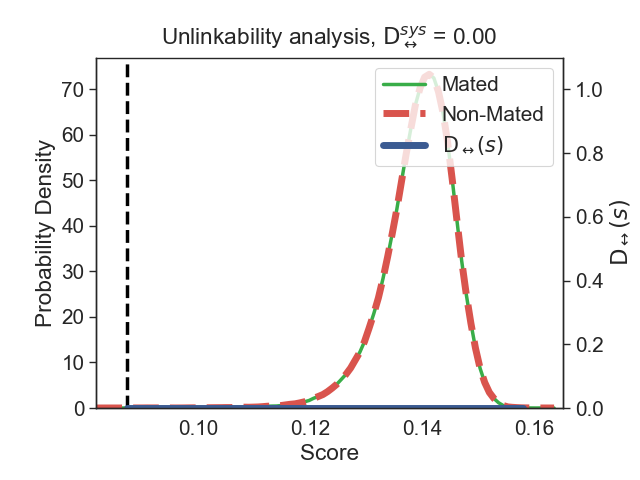}
		\label{fig:unlink-xored-3}}
	\subfloat[Proposed][XOR - 32 Blocks - 10 Bits]{
		\includegraphics[width=0.22\textwidth]{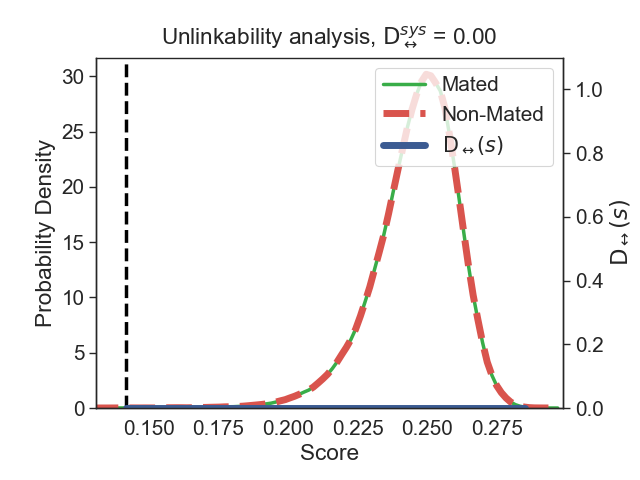}
		\label{fig:unlink-xored-4}}
	\caption{Unlinkability metrics obtained for proposed template protection scheme for various configurations}
	\label{fig:unlink-scores-iitd}
	\vspace{-3mm}
\end{figure*}

\subsection{UBIRIS v1 Dataset}
We further evaluate the proposed approach on the visible spectrum iris recognition, especially captured in unconstrained setting. To this end, we evaluate the proposed approach on UBIRIS.v1 database composed of 1877 images collected from 241 persons in two distinct sessions in visible spectrum at a stand-off distance. Unlike the other existing public and free databases for iris recognition, UBIRIS v1 incorporates images with several noise factors, thus permitting the evaluation of robustness proposed template protection scheme. The enrollment set consists of minimized noise factors, specially those relative to reflections, luminosity and contrast, having installed image capture framework inside a dark room as compared to the second session. While in the second session, the iris images are captured with the introduced natural luminosity factor. Thus, we employ the images from the first session for the enrolment template creation and second session for template verification.  Given the database has number of noise factors, we have eliminated the images that have severe segmentation errors and completely off-angled iris images. The segmentation and the normalization of the iris region is performed using the recent approach of Total-variation based segmentation \cite{zhao2015accurate} to derive the iris images of size $100 \times 360$ pixels which is then used to extract the features by employing $1D \ Log-Gabor$ representation.

\subsection{Results on UBIRIS v1 Iris Dataset}
For the sake of brevity, we present the result of proposed approach on the UBIRIS v1 iris dataset in the Figure~\ref{fig:ubiris-iris-multiple-config} and we do not report the performance of Bloom-Filter based template protection due to non-ideal performance. As it can be observed from the DET curves, the proposed approach has certain loss in terms of the biometric performance as compared to the unprotected templates as anticipated. Nevertheless, the performance is comparable in the IV version of the proposed approach as seen in Fig.\ref{fig:ubiris-interleaved}. While a severe degree of performance loss is seen across the XOR variant, it is also interesting to note that the XOR version with a block length of $4$ with $5$ bits is performing close to unprotected template. This observation is consistent to our earlier observations where we have noted that the lower block widths perform superior as compared to the larger block widths. Further, the Receiver Operating Characteristics (ROC)  to indicate the biometric performance is provided in supplementary material in Fig.~\ref{fig:ubiris-iris-multiple-config-roc}.

\subsection{UBIRIS v2 Dataset}
We further evaluate the proposed approach on the visible spectrum iris recognition captured in unconstrained setting using the second set - UBIRIS v2. Similar to UBIRIS v1, this dataset has data with various non-ideal iris images, imaging distances, subject perspectives and lighting conditions. Further, the iris data is captured under both natural and artificial lighting sources along with a session interval of a weeks between the enrolment and probe. It has to be further noted that in the second session, the location and orientation of the acquisition device and artificial light sources was changed increasing the diversity in capture conditions with a total of 261 subjects totalling to 522 irises. The segmentation and the normalization of the iris region is performed using Total-variation based segmentation \cite{zhao2015accurate} to derive the iris images of size $100 \times 360$ pixels and then extract the features using $1D \ Log-Gabor$ representation.

\subsection{Results on UBIRIS v2 Iris Dataset}
Similar to the results obtained on UBIRIS v1 dataset, the results obtained on the UBIRIS v2 iris dataset is presented in Figure~\ref{fig:ubiris-v2-iris-multiple-config}. The observations correlate to earlier observations and a slight degradation in performance can be noted as compared to the unprotected domain indicating the applicability of the proposed approach. It has to be noted that results of Bloom-Filter based template protection is not reported on this dataset due to non-ideal biometric performance. 

\subsection{Fine Tuned Experiments on UBIRIS v2 Iris Dataset}
\begin{figure}[htp]
	\centering
		\resizebox{0.33\textwidth}{!}{%
			\input{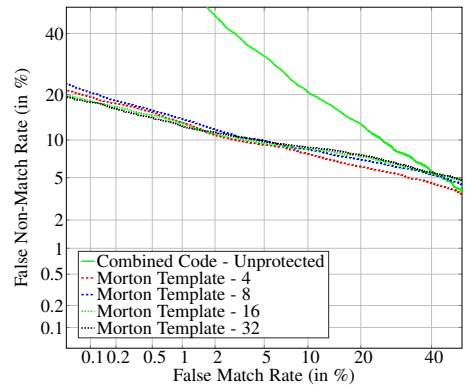}
		}
	\caption{Performance of multiple configurations with fine-tuned approach of proposed template protection on UBIRIS v2 dataset.}
	\label{fig:ubiris-v2-iris-multiple-config-finetuned}
	\vspace{-3mm}
\end{figure}

Seeking an answer to the low performance of the protected templates, we conducted an additional experiment where the segmentation masks were employed prior to deriving the stable and discriminable code  corresponding to protected templates from iriscodes. As the masks eliminate the segmentation errors and thereby provide first level of reliable bits, our assertion was that such a refined iriscodes would improve the performance of proposed template protection. In order to validate the assertion, we report the following DET as show in the Figure \ref{fig:ubiris-v2-iris-multiple-config-finetuned}. As seen from the Figure \ref{fig:ubiris-v2-iris-multiple-config-finetuned}, the performance of protected template increases through the use of the masks prior to creation of multiple buckets.

\section{Security Analysis : Unlinkability}
\label{sec:unlinkability-analysis} 
We present the security analysis to demonstrate the unlinkability of the proposed template protection using the recent Unlinkability Analysis Framework \cite{gomez2018general}. Under this framework, we measure the similarity of the mated imposter distribution versus non-mated imposter distribution. To maintain the terseness, we present the results on IITD Iris dataset  in Figure \ref{fig:unlink-scores-iitd} where a high degree of unlinkability can be observed along with the unlinkability metric by $\mathrm{D}_{\leftrightarrow}^{\mathit{sys}}$. The lower $\mathrm{D}_{\leftrightarrow}^{\mathit{sys}}$ indicates superior unlinkability. Further, one can observe the consistent unlinkability across different configurations for varying blocks and varying bits.

\section{Future Works}
\label{sec:future-works} 
Although the preliminary analysis of the security implications is carried out for unlinkability, this work has not investigated the potential threats with advanced attacks. While it can be noted that the analysis carried out for Bloom-Filter reliability can be generalized to the proposed approach and theoretical proof can be borrowed from the recent works \cite{gomez2016unlinkable, gomez2018multi}, it will be interesting to investigate the newer attacks. In the future works, we intend to analyse the robustness of proposed attacks towards such attacks. In the second direction, we shall explore the proposed approach for the other modalities to preserve the privacy of the biometric data.

\section{Conclusions}
\label{sec:conclusions}
We have addressed the problem of template protection for iris recognition in this work, specifically, we presented a novel approach for template protection employing Morton Filters on constrained and unconstrained iriscodes. Exploiting the intra-class and inter-class distribution of iriscodes, we have designed multiple buckets to realize Morton-Filter based template protection. The proposed approach has been evaluated for both performance and linkability challenges using four publicly available iris databases captured in NIR spectrum and VIS spectrum. Although these empirical validation on NIR domain demonstrate the feasibility and applicability of proposed approach, we have evaluated the approach on the unconstrained iriscodes captured in visible spectrum to measure the scalability and robustness. Along with providing security, the loss in the performance is observed to be marginal in visible spectrum iriscodes owing to noisy nature of images. The approach being robust in comparison to Bloom Filter, can be applied to other modalities with suitable adaptations in the future works.

\section*{Reproducible Research}
\label{sec:reproducible-research}
In order to facilitate the reproducible research, we intend to make the code of the proposed approach publicly available along with this article. Details on availing the code shall be provided in the final version of the manuscript. 

\section*{Acknowledgement}
This work was carried out under the partial funding of the Research Council of Norway under Grant No. IKTPLUSS 248030/O70.
\balance
{\scriptsize
\bibliographystyle{ieee}
\bibliography{stable-map-iris-template-btas2019-190418}
}
\onecolumn
\newpage
\renewcommand\thefigure{\thesection\arabic{figure}}    
\setcounter{figure}{0}    
\section*{Appendix}
\begin{alphasection}
\section{State-of-art Review for Iris Template Protection}
In this section we provide a brief overview of the state-of-art template protection schemes proposed and employed for the iris recognition. This section is supplementary to Table~\ref{tab:sota-iris-template} with specifics of previously proposed approaches.
\begin{itemize}
	\item Yang and Verbauwhede \cite{yang2007secure} proposed a iris template protection approach by employing Error Correcting Code (ECC) cryptographic technique with the reliable bits selection to improve the verification accuracy.  In their work, Bose-Chaudhuri-Hochquenghem (BCH) code of a random bit-stream was introduced to eliminate the considerable differences between the features extracted from different scans of irises such that template protection was scheme reliable. The experiments were conducted on a a subset of CASIA iris data \footnote{Details on the dataset and number of images are not provided in the article.}.
	\item Nandakumar and Jain \cite{nandakumar2008multibiometric} proposed another framework based on fuzzy-vault scheme to derive private keys from iris patterns. In the same work, they also proposed to fuse multiple biometric modalities, specifically fingerprint and iris. A salting transformation based on a transformation key was employed to indirectly convert the fixed-length binary vector representation of iriscode into an unordered set representation and further secured using the fuzzy vault. The performance for the iris template protection was reported on CASIA iris image database v1.0 consisting of 108 different eyes with 7 images per eye collected over two sessions and one image from each session was used for evaluation.
	\item Maiorana et al., \cite{maiorana2014iris} proposed a template protection framework inspired by the digital modulation paradigm where the properties of modulation constellations and turbo codes with soft-decoding were exploited to design a system. The approach was tested on the  Interval subset of the CASIA-IrisV4 database where high performance in terms of both verification rates and security was reported.
	\item Zhang et al., \cite{zhang2009robust} proposed a concatenated coding scheme and bit masking scheme to construct an iris cryptosystem. The concatenated coding scheme was proposed to embed long keys into the iris data and concatenated code combined with Reed-Solomon code and convolutional code. Further, a bit masking scheme was proposed to minimize and randomize the errors to make the error pattern more suitable for the coding scheme. The iris cryptosystem reported a FRR of 0.52\% with the key length of 938 bits on a internal database of 128 iris images captured across 3 different sessions.
	\item Rathgeb et al. \cite{rathgeb2013alignment} proposed the popular Bloom-filter based biometric template protection. The iris codes were processed through K-hashes resulting in the protected templates through transformation. While the framework was later reported to be weak against linkability attacks, \cite{hermans2014bloom}, the same was fixed by adding the private keys as proposed in \cite{gomez2016unlinkable}. The improved approach was evaluated on BioSecure Multimodal Database to demonstrate the robustness towards linkability and reversibility. 
	\item Rathgeb and Busch \cite{rathgeb2014cancelable, rathgeb2014application} proposed a framework based on adaptive Bloom filter-based transforms that were applied in order to mix binary iris biometric templates at feature level, where iris-codes are obtained from both eyes of a single subject. The irreversible mixing transform generating alignment-free templates obscured information present in different iris-codes. Further, the proposed transform was parameterized to achieve unlinkability resulting in implementing cancelable multi-biometrics. The experiments on IITD Iris Database version 1.0 resulted in EER below 0.5\% for different feature extraction methods and fusion scenarios.
	\item Rathgeb et al. \cite{rathgeb2015towards} further proposed a generic framework for generating an irreversible representation of multiple biometric templates extending the framework of adaptive Bloom filters from earlier work\cite{rathgeb2014cancelable}. The technique enabled a feature level fusion of different biometrics (face and iris) to a single protected template, improving privacy protection compared to the corresponding systems based on a single biometric trait. 
	\item Lai et al \cite{lai2017cancellable} recently proposed a new scheme to generate cancellable iris template with Jaccard similarity matcher by modifying the of Min-hashing to strengthen the privacy security in Indexing-First-One (IFO) hashing. The proposed approach was evaluated on CASIA v3 iris database and reported to result in 0.16\% Equal Error Rate (EER). Further, it is worth noting that the approach is reported to be resistant against Single Hash Attack, Multi-Hash Attack, Attack via record multiplicity and Pre-image attack. Unlike many previous works, this work also reported the unlinkability and revocability analysis on the proposed approach.
\end{itemize}

\newpage
\section{DET curves corresponding to proposed approach and invariance to block size in NIR spectrum}
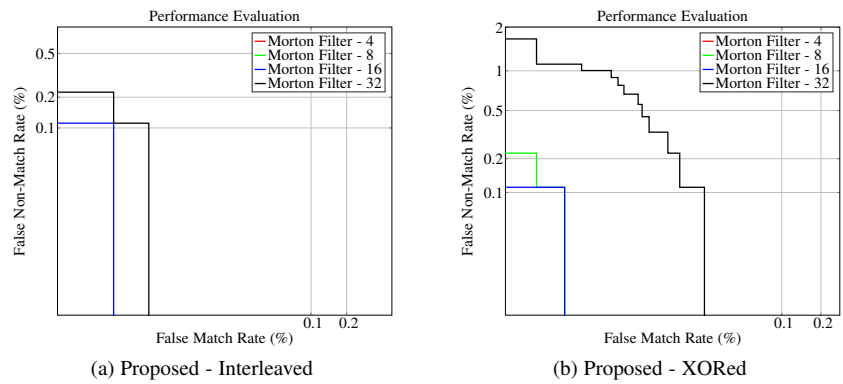
\begin{figure*}[htp]
	\centering
	\subfloat[Bloom-Filter][Proposed - Interleaved]{
		\resizebox{0.28\textwidth}{!}{%
%
\begin{tikzpicture}
\tikzset{lw/.style = {line width=8pt}}
\tikzstyle{every node}=[fontscale=24]
\begin{axis}[%
width=01\textwidth,
at={(1.011in,0.642in)},
scale only axis,
xmin=-4.59851825209886,
xmax=-2.6089695001817,
xtick={-3.09023224677291,-2.8781616977673,-2.57582930094926,-2.32634787735637,-2.05374890849825,-1.64485362793551,-1.28155156321185},
xticklabels={{  0.1},{  0.2},{ 0.5},{  1},{  2},{  5},{  10}},
xlabel={False Match Rate (\%)},
ymin=-4.38290454425593,
ymax=-2.39335579233876,
ytick={-3.09023224677291,-2.8781616977673,-2.57582930094926,-2.32634787735637,-2.05374890849825,-1.64485362793551,-1.28155156321185},
yticklabels={{  0.1},{  0.2},{ 0.5},{  1},{  2},{  5},{  10}},
ylabel={False Non-Match Rate (\%)},
axis background/.style={fill=white},
title={Performance Evaluation},
xmajorgrids,
ymajorgrids,
legend style={legend cell align=left, align=left, draw=white!15!black}
]
\addplot [color=red, line width=2.0pt]
table[row sep=crcr]{%
	-4.38290454425593 -4.59851825209886\\
};
\addlegendentry{Morton Filter - 4}

\addplot [color=green, line width=2.0pt]
table[row sep=crcr]{%
		-4.38290454425593 -4.59851825209886\\
};
\addlegendentry{Morton Filter - 8}

\addplot [color=blue, line width=2.0pt]
  table[row sep=crcr]{%
-4.26468137689947	-4.58185941944765\\
-4.26468137689947	-3.05746953454897\\
-4.79747312729058	-3.05746953454897\\
};
\addlegendentry{Morton Filter - 16}

\addplot [color=black, line width=2.0pt]
  table[row sep=crcr]{%
-4.05540528332275	-4.58185941944765\\
-4.05540528332275	-3.05746953454897\\
-4.26468137689947	-3.05746953454897\\
-4.26468137689947	-2.84334396830158\\
-4.79747312729058	-2.84334396830158\\
};
\addlegendentry{Morton Filter - 32}

\end{axis}

\begin{axis}[%
width=7.778in,
height=5.833in,
at={(0in,0in)},
scale only axis,
xmin=0,
xmax=1,
ymin=0,
ymax=1,
axis line style={draw=none},
ticks=none,
axis x line*=bottom,
axis y line*=left,
legend style={legend cell align=left, align=left, draw=white!15!black}
]
\end{axis}
\end{tikzpicture}%
		}
		\label{fig:subfig1}}
	\qquad
	\subfloat[Proposed][Proposed - XORed]{
		\resizebox{0.28\textwidth}{!}{%
%
\begin{tikzpicture}
\tikzset{lw/.style = {line width=8pt}}
\tikzstyle{every node}=[font=\Huge]
\begin{axis}[%
width=01\textwidth,
at={(1.011in,0.642in)},
scale only axis,
xmin=-4.58512088917153,
xmax=-2.77644020561046,
xtick={-3.09023224677291,-2.8781616977673,-2.57582930094926,-2.32634787735637,-2.05374890849825,-1.64485362793551,-1.28155156321185},
xticklabels={{  0.1},{  0.2},{ 0.5},{  1},{  2},{  5},{  10}},
xlabel={False Match Rate (\%)},
ymin=-3.86017072901679,
ymax=-2.05149004545572,
ytick={-3.09023224677291,-2.8781616977673,-2.57582930094926,-2.32634787735637,-2.05374890849825,-1.64485362793551,-1.28155156321185},
yticklabels={{  0.1},{  0.2},{ 0.5},{  1},{  2},{  5},{  10}},
ylabel={False Non-Match Rate (\%)},
axis background/.style={fill=white},
title={Performance Evaluation},
xmajorgrids,
ymajorgrids,
legend style={legend cell align=left, align=left, draw=white!15!black}
]
\addplot [color=red, line width=2.0pt]
table[row sep=crcr]{%
-3.86017072901679 -4.58512088917153\\
};
\addlegendentry{Morton Filter - 4}

%
%

\addplot [color=green, line width=2.0pt]
  table[row sep=crcr]{%
-4.26468137689947	-4.04103879737289\\
-4.26468137689947	-3.05746953454897\\
-4.41697305240491	-3.05746953454897\\
-4.41697305240491	-2.84334396830158\\
-4.76598895752764	-2.84334396830158\\
};
\addlegendentry{Morton Filter - 8}
\addplot [color=blue, line width=2.0pt]
  table[row sep=crcr]{%
-4.26468137689947	-4.04103879737289\\
-4.26468137689947	-3.05746953454897\\
-4.76598895752764	-3.05746953454897\\
};
\addlegendentry{Morton Filter - 16}
\addplot [color=black, line width=2.0pt]
  table[row sep=crcr]{%
-3.50862351962253	-4.04103879737289\\
-3.50862351962253	-3.05746953454897\\
-3.64225621105236	-3.05746953454897\\
-3.64225621105236	-2.84334396830158\\
-3.70643014655905	-2.84334396830158\\
-3.70643014655905	-2.71157550009372\\
-3.80793186081145	-2.71157550009372\\
-3.80793186081145	-2.61477704977736\\
-3.84589203042635	-2.61477704977736\\
-3.84589203042635	-2.53762613008386\\
-3.86717140593413	-2.53762613008386\\
-3.86717140593413	-2.4731482564224\\
-3.94417183144219	-2.4731482564224\\
-3.94417183144219	-2.41755901991457\\
-3.97605940545801	-2.41755901991457\\
-3.97605940545801	-2.36856706301687\\
-4.01258659691191	-2.36856706301687\\
-4.01258659691191	-2.32467611213493\\
-4.17325155102223	-2.32467611213493\\
-4.17325155102223	-2.28485334611196\\
-4.41697305240491	-2.28485334611196\\
-4.41697305240491	-2.12625367599754\\
-4.76598895752764	-2.12625367599754\\
};
\addlegendentry{Morton Filter - 32}
\end{axis}

\begin{axis}[%
width=7.778in,
height=5.833in,
at={(0in,0in)},
scale only axis,
xmin=0,
xmax=1,
ymin=0,
ymax=1,
axis line style={draw=none},
ticks=none,
axis x line*=bottom,
axis y line*=left,
legend style={legend cell align=left, align=left, draw=white!15!black}
]
\end{axis}
\end{tikzpicture}%
		}
		\label{fig:subfig2}}
	\caption{Performance of multiple configurations for proposed Morton Filter on IITD Iris dataset for a bit length of $5$ bits.}
	\label{fig:iitd-iris-multiple-config}
\end{figure*}

\section{Receiver Operating Characteristic (ROC) curves corresponding to proposed approach}
\begin{figure*}[htp]
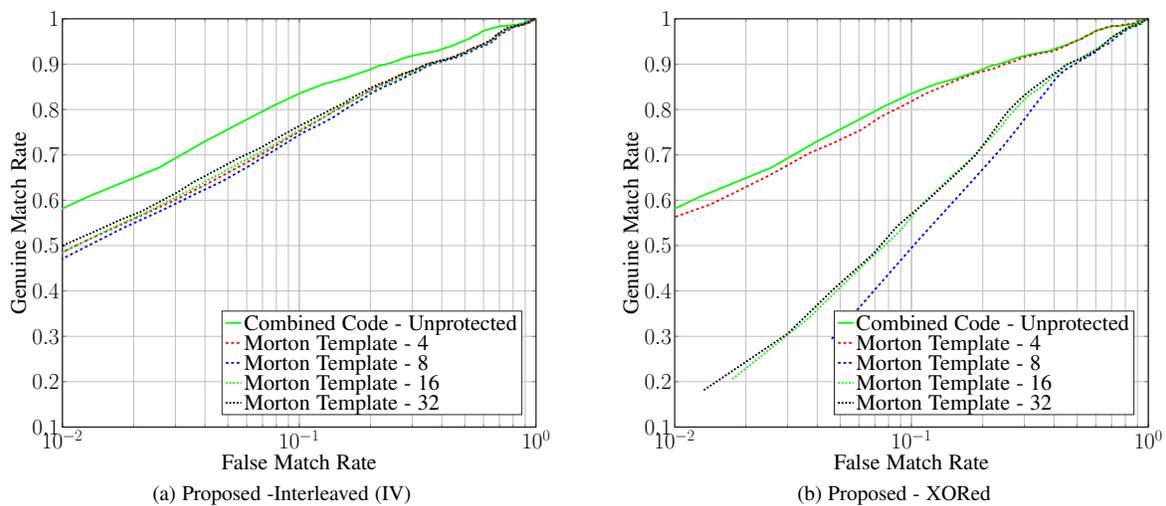

	\centering
	\subfloat[Bloom-Filter][Proposed -Interleaved (IV)]{
		\resizebox{0.4\textwidth}{!}{%
			\input{pgftikz/roc-interleaved-different-key1-btas2019-morton-4-32-Bits-5-ubiris-part1-template.mat.tex}
		}
		\label{fig:ubiris-interleaved-roc}}
	\qquad
	\subfloat[Proposed][Proposed - XORed]{
		\resizebox{0.4\textwidth}{!}{%
			\input{pgftikz/roc-xor-different-key1-btas2019-morton-4-32-Bits-5-ubiris-part1-template.mat.tex}
		}
		\label{fig:ubiris-xored-roc}}
	\caption{Performance of multiple configurations for proposed template protection on UBIRIS v1 dataset.}
	\label{fig:ubiris-iris-multiple-config-roc}
\end{figure*}

\begin{figure*}[htp]
	\centering
	\subfloat[Bloom-Filter][Proposed -Interleaved (IV)]{
		\resizebox{0.4\textwidth}{!}{%
			\input{pgftikz/roc-interleaved-different-key1-btas2019-morton-4-32-Bits-5-ubiris-part2-template.mat.tex}
		}
		\label{fig:ubiris-v2-interleaved-roc}}
	\qquad
	\subfloat[Proposed][Proposed - XORed]{
		\resizebox{0.4\textwidth}{!}{%
			\input{pgftikz/roc-xor-different-key1-btas2019-morton-4-32-Bits-5-ubiris-part2-template.mat.tex}
		}
		\label{fig:ubiris-v2-xored-roc}}
	\caption{Performance of multiple configurations for proposed template protection on UBIRIS v2 dataset.}
	\label{fig:ubiris-v2-iris-multiple-config-roc}
\end{figure*}
\clearpage
\newpage
\section{Unlinkability Analysis}
\begin{figure*}[htp]
	\centering
	\subfloat[Bloom-Filter][IL - 4 Blocks - 5 Bits]{
		\includegraphics[width=0.33\textwidth]{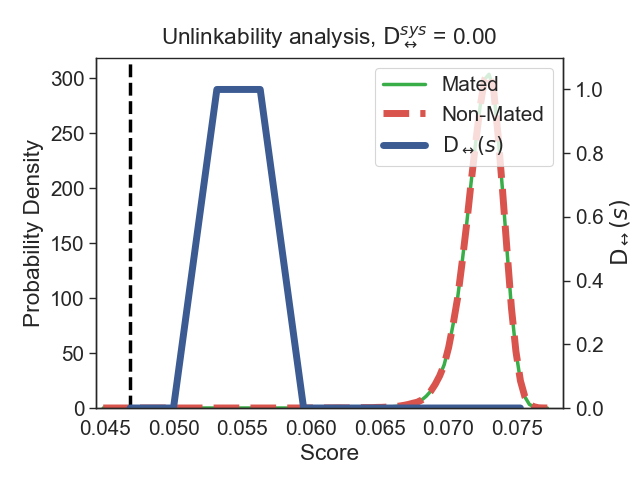}
		\label{fig:unlink-interleave-1}}
	\subfloat[Proposed][IL - 8 Blocks - 5 Bits]{
		\includegraphics[width=0.33\textwidth]{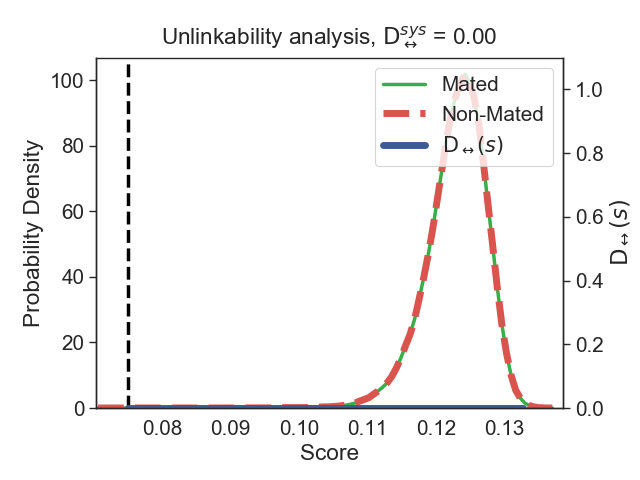}
		\label{fig:unlink-interleave-2}}\\
	\subfloat[Proposed][IL - 16 Blocks - 5 Bits]{
		\includegraphics[width=0.33\textwidth]{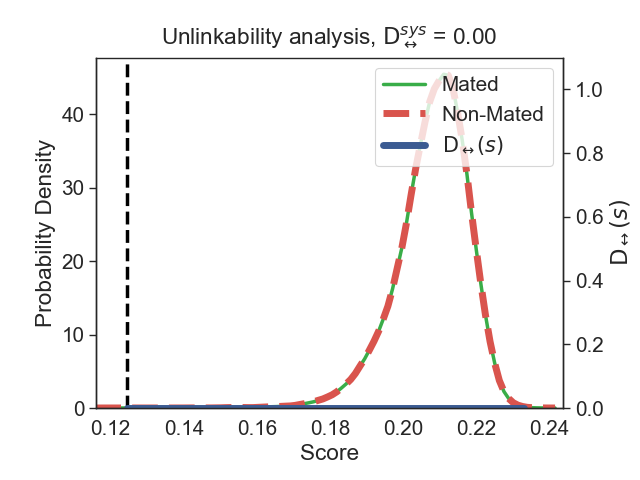}
		\label{fig:unlink-interleave-3}}
	\subfloat[Proposed][IL - 32 Blocks - 5 Bits]{
		\includegraphics[width=0.33\textwidth]{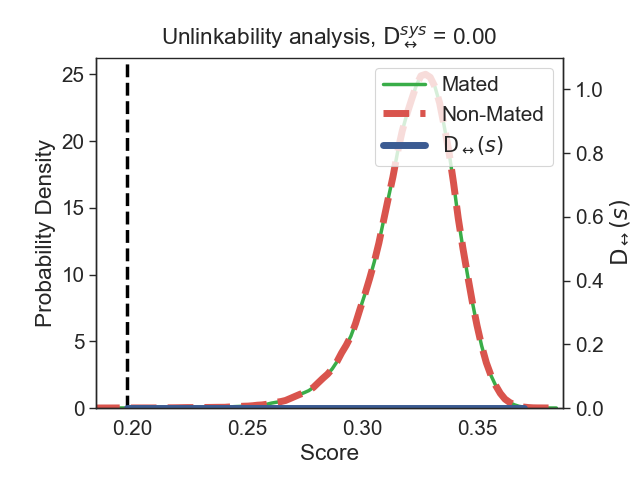}
		\label{fig:unlink-interleave-4}}\\
	\subfloat[Bloom-Filter][XOR - 4 Blocks - 10 Bits]{
		\includegraphics[width=0.33\textwidth]{figures/unlink-xored-morton-4-Bits-5_key12}
		\label{fig:unlink-xored-1-appendix}}
	\subfloat[Proposed][XOR - 8 Blocks - 10 Bits]{
		\includegraphics[width=0.33\textwidth]{figures/unlink-xored-morton-8-Bits-5_key12}
		\label{fig:unlink-xored-2-appendix}}\\
	\subfloat[Proposed][XOR - 16 Blocks - 10 Bits]{
		\includegraphics[width=0.33\textwidth]{figures/unlink-xored-morton-16-Bits-5_key12}
		\label{fig:unlink-xored-3-appendix}}
	\subfloat[Proposed][XOR - 32 Blocks - 10 Bits]{
		\includegraphics[width=0.33\textwidth]{figures/unlink-xored-morton-32-Bits-5_key12}
		\label{fig:unlink-xored-4-appendix}}
	\caption{Unlinkability metrics obtained for proposed template protection scheme for various configurations}
	\label{fig:unlink-scores-iitd-appendix}
	\vspace{-3mm}
\end{figure*}
\end{alphasection}

\end{document}


\title{Tensor Robust Principal Component Analysis with A New Tensor Nuclear Norm}

\author{Canyi~Lu,~Jiashi~Feng,~Yudong~Chen,~Wei Liu,~Member,~IEEE,~Zhouchen Lin,~\IEEEmembership{Fellow,~IEEE},
	
	 ~and~Shuicheng Yan,~\IEEEmembership{Fellow,~IEEE}
	
	\IEEEcompsocitemizethanks{\IEEEcompsocthanksitem C. Lu is with the Department of Electrical and Computer Engineering, Carnegie Mellon University (e-mail: canyilu@gmail.com).\protect
	\IEEEcompsocthanksitem J. Feng and S. Yan are with the Department of Electrical and Computer Engineering, National University of Singapore, Singapore (e-mail: elefjia@nus.edu.sg; eleyans@nus.edu.sg).
	\IEEEcompsocthanksitem Y. Chen is with the School of Operations Research and Information Engineering, Cornell University (e-mail: yudong.chen@cornell.edu).
	\IEEEcompsocthanksitem  W. Liu is with the Computer Vision Group, Tencent AI Lab, Shenzhen, China (e-mail: wliu@ee.columbia.edu).
	\IEEEcompsocthanksitem Z. Lin is with the Key Laboratory of Machine Perception (MOE), School of EECS, Peking University, Beijing 100871, China, and also with the Cooperative Medianet Innovation Center, Shanghai Jiao Tong University, Shanghai 200240, China (e-mail: zlin@pku.edu.cn).}
}

\markboth{}%
{Shell \MakeLowercase{\textit{et al.}}: Bare Advanced Demo of IEEEtran.cls for Journals}

\IEEEtitleabstractindextext{%
\begin{abstract}
	In this paper, we consider the Tensor Robust Principal Component Analysis (TRPCA) problem, which aims to exactly recover the low-rank and sparse components from their sum. Our model is based on the recently proposed tensor-tensor product (or t-product) \cite{kilmer2011factorization}. Induced by the t-product, we first rigorously deduce the tensor spectral norm, tensor nuclear norm, and tensor average rank, and show that the tensor nuclear norm is the convex envelope of the tensor average rank within the unit ball of the tensor spectral norm. These definitions, their relationships and properties are consistent with matrix cases. Equipped with the new tensor nuclear norm, we then solve the TRPCA problem by solving a convex program and provide the theoretical guarantee for the exact recovery. Our TRPCA model and recovery guarantee include matrix RPCA as a special case. Numerical experiments verify our results, and the applications to image recovery and background modeling problems demonstrate the effectiveness of our method. 
\end{abstract}
	
\begin{IEEEkeywords}
	Tensor robust PCA, convex optimization, tensor nuclear norm, tensor singular value decomposition
\end{IEEEkeywords}}
\maketitle		
\IEEEdisplaynontitleabstractindextext
\IEEEpeerreviewmaketitle
	
	\ifCLASSOPTIONcompsoc
	\IEEEraisesectionheading{\section{Introduction}\label{sec:introduction}}
	\else
	
\section{Introduction}\label{sec:introduction}
\fi
\IEEEPARstart{P}{rincipal} Component Analysis (PCA) is a fundamental approach for data analysis. It exploits low-dimensional structure in high-dimensional data, which commonly exists in different types of data, \textit{e.g.}, image, text, video and bioinformatics. It is computationally efficient and powerful for data instances which are mildly corrupted by small noises. However, a major issue of PCA is that it is brittle to be grossly corrupted or outlying observations, which are ubiquitous in real-world data. To date, a number of robust versions of PCA have been proposed, but many of them suffer from a high computational cost. 

\begin{figure}[!t]
	\centering
	\begin{subfigure}[b]{0.4\textwidth}
		\centering
		\includegraphics[width=\textwidth]{fig/fig_rpca.pdf}\vspace{2mm}
	\end{subfigure} 
	
	\begin{subfigure}[b]{0.4\textwidth}
		\centering
		\includegraphics[width=\textwidth]{fig/fig_trpcaill.pdf}
	\end{subfigure}
	\caption{\small{Illustrations of RPCA \cite{RPCA} (up row) and our Tensor RPCA (bottom row). RPCA: low-rank and sparse matrix decomposition from noisy matrix observations. Tensor RPCA: low-rank and sparse tensor decomposition from noisy tensor observations.}}
	\label{fig_trpca1}
	\vspace{-0.45cm}
\end{figure}	
	
The Robust PCA \cite{RPCA} is the first polynomial-time algorithm with strong recovery guarantees. Suppose that we are given an observed matrix $\Xm\in\mathbb{R}^{n_1\times n_2}$, which can be decomposed as $\Xm=\Lm_0+\Em_0$, where $\Lm_0$ is low-rank and $\Em_0$ is sparse. It is shown in \cite{RPCA} that if the singular vectors of $\Lm_0$ satisfy some incoherent conditions, \textit{e.g.}, $\Lm_0$ is low-rank and $\Sm_0$ is sufficiently sparse, then $\Lm_0$ and $\Sm_0$ can be exactly recovered with high probability by solving the following convex problem
\begin{equation}\label{rpca}
	\min_{\Lm,\Em} \ \norm{\Lm}_*+\lambda\norm{\Em}_1, \ \st \ \Xm=\Lm+\Em,
\end{equation}
where $\norm{\Lm}_*$ denotes the nuclear norm (sum of the singular values of $\Lm$), and $\norm{\Em}_1$ denotes the $\ell_1$-norm (sum of the absolute values of all the entries in $\Em$). Theoretically, RPCA is guaranteed to work even if the rank of $\Lm_0$ grows almost linearly in the dimension of the matrix, and the errors in $\Sm_0$ are up to a constant fraction of all entries. The parameter $\lambda$ is suggested to be set as $1/\sqrt{\max(n_1,n_2)}$ which works well in practice. Algorithmically, program (\ref{rpca}) can be solved by efficient algorithms, at a cost not too much higher than PCA. RPCA and its extensions have been successfully applied to background modeling \cite{RPCA}, subspace clustering \cite{robustlrr,liu2012fixed}, video compressive sensing \cite{waters2011sparcs}, \textit{etc}. 

One major shortcoming of RPCA is that it can only handle 2-way (matrix) data. However, real data is usually multi-dimensional in nature-the information is stored in multi-way arrays known as tensors \cite{kolda2009tensor,qi2018multi}. For example, a color image is a 3-way object with column, row and color modes; a greyscale video is indexed by two spatial variables and one temporal variable. To use RPCA, one has to first restructure the multi-way data into a matrix. Such a preprocessing usually leads to an information loss and would cause a performance degradation. To alleviate this issue, it is natural to consider extending RPCA to manipulate the tensor data by taking advantage of its multi-dimensional structure.
	
In this work, we are interested in the Tensor Robust Principal Component (TRPCA) model which aims to exactly recover a low-rank tensor corrupted by sparse errors. See Figure \ref{fig_trpca1} for an intuitive illustration. More specifically, suppose that we are given a data tensor $\X$, and know that it can be decomposed as
\begin{equation}\label{xls}
	\X=\LL_0+\E_0,
\end{equation}
where $\LL_0$ is low-rank and $\E_0$ is sparse, and both components are of arbitrary magnitudes. Note that we do not know the locations of the nonzero elements of $\E_0$, not even how many there are. Now we consider a similar problem to RPCA. Can we recover the low-rank and sparse components exactly and efficiently from $\X$? This is the problem of tensor RPCA studied in this work.
	
The tensor extension of RPCA is not easy since the numerical algebra of tensors is fraught with hardness results \cite{hillar2013most}. A main issue is that the tensor rank is not well defined with a tight convex relaxation. Several tensor rank definitions and their convex relaxations have been proposed but each has its limitation. For example, the CP rank \cite{kolda2009tensor}, defined as the smallest number of rank one tensor decomposition, is generally NP-hard to compute. Also its convex relaxation is intractable. This makes the low CP rank tensor recovery challenging. The tractable Tucker rank \cite{kolda2009tensor} and its convex relaxation are more widely used. For a $k$-way tensor $\X$, the Tucker rank is a vector defined as $\text{rank}_{\text{tc}}(\X):=\left( \text{rank}(\Xm^{\{1\}}), \text{rank}(\Xm^{\{2\}}), \cdots, \text{rank}(\Xm^{\{k\}}) \right)$, where $\Xm^{\{i\}}$ is the mode-$i$ matricization of $\X$ \cite{kolda2009tensor}. Motivated by the fact that the nuclear norm is the convex envelope of the matrix rank within the unit ball of the spectral norm, the Sum of Nuclear Norms (SNN)~\cite{liu2013tensor}, defined as $\sum_i\norm{\Xm^{\{i\}}}_*$, is used as a convex surrogate of $\sum_i\text{rank}({\Xm^{\{i\}}})$. Then the work \cite{mu2013square} considers the Low-Rank Tensor Completion (LRTC) model based on SNN:
\begin{equation}\label{tensorc}
	\min_{\X} \ \sum_{i=1}^{k}\lambda_i\norm{\Xm^{\{i\}}}_*, \ \st \ \Pomega(\X) = \Pomega(\M),
\end{equation}
where $\lambda_i>0$, and $\Pomega(\X)$ denotes the projection of $\X$ on the observed set $\Omegat$. The effectiveness of this approach for image processing has been well studied in \cite{liu2013tensor,gandy2011tensor,tomioka2010estimation}. However, SNN is not the convex envelope of $\sum_i\text{rank}({\Xm^{\{i\}}})$ \cite{romera2013new}. Actually, the above model can be substantially suboptimal \cite{mu2013square}: reliably recovering a $k$-way tensor of length $n$ and Tucker rank $(r,r,\cdots,r)$ from Gaussian measurements requires $O(rn^{k-1})$ observations. In contrast, a certain (intractable) nonconvex formulation needs only $O(rK +nrK)$ observations. The work \cite{mu2013square} further proposes a better (but still suboptimal) convexification based on a more balanced matricization. The work \cite{huang2014provable} presents the recovery guarantee for the SNN based tensor RPCA model
\begin{equation}\label{eqsnn}
	\min_{\LL,\E} \ \sum_{i=1}^{k} \lambda_i\norm{{\bm{L}}^{\{i\}}}_*+\norm{\E}_1,	\st \ \X=\LL+\E.
\end{equation}
	
The limitations of existing works motivate us to consider an interesting problem: is it possible to define a new tensor nuclear norm such that it is a tight convex surrogate of certain tensor rank, and thus its resulting tensor RPCA enjoys a similar tight recovery guarantee to that of the matrix RPCA? This work will provide a positive answer to this question. Our solution is inspired by the recently proposed tensor-tensor product (t-product) \cite{kilmer2011factorization} which is a generalization of the matrix-matrix product. It enjoys several similar properties to the matrix-matrix product. For example, based on t-product, any tensors have the tensor Singular Value Decomposition (t-SVD) and this motivates a new tensor rank, \textit{i.e.}, tensor tubal rank \cite{kilmer2013third}. To recover a tensor of low tubal rank, we propose a new tensor nuclear norm which is rigorously induced by the t-product. First, the tensor spectral norm can be induced by the operator norm when treating the t-product as an operator. Then the tensor nuclear norm is defined as the dual norm of the tensor spectral norm. We further propose the tensor average rank (which is closely related to the tensor tubal rank), and prove that its convex envelope is the tensor nuclear norm within the unit ball of the tensor spectral norm. It is interesting that this framework, including the new tensor concepts and their relationships, is consistent with the one for the matrix cases. Equipped with these new tools, we then study the TRPCA problem which aims to recover the low tubal rank component $\LL_0$ and sparse component $\E_0$ from noisy observations $\X=\LL_0+\E_0\in\mathbb{R}^{n_1\times n_2\times n_3}$ (this work focuses on the 3-way tensor) by convex optimization
\begin{align}\label{trpca}
	\min_{\LL,\ \E} \ \norm{\LL}_*+\lambda\norm{\E}_1, \ \st \ \X=\LL+\E,
\end{align}
where $\norm{\LL}_*$ is our new tensor nuclear norm (see the definition in Section \ref{sec_TNN}). We prove that under certain incoherence conditions, the solution to (\ref{trpca}) perfectly recovers the low-rank and the sparse components, provided of course that the tubal rank of $\LL_0$ is not too large, and that $\E_0$ is reasonably sparse. A remarkable fact, like in RPCA, is that (\ref{trpca}) has no tunning parameter either. Our analysis shows that $\lambda=1/\sqrt{\max(n_1,n_2)n_3}$ guarantees the exact recovery no matter what $\LL_0$ and $\E_0$ are. As a special case, if $\X$ reduces to a matrix ($n_3=1$ in this case), all the new tensor concepts reduce to the matrix cases. Our TRPCA model (\ref{trpca}) reduces to RPCA in (\ref{rpca}), and also our recovery guarantee in Theorem \ref{thm1} reduces to Theorem 1.1 in \cite{RPCA}. Another advantage of (\ref{trpca}) is that it can be solved by polynomial-time algorithms.
	
The contributions of this work are summarized as follows:
\begin{enumerate}[1.]
	\item Motivated by the t-product \cite{kilmer2011factorization} which is a natural generalization of the matrix-matrix product, we rigorously deduce a new tensor nuclear norm and some other related tensor concepts, and they own the same relationship as the matrix cases. This is the foundation for the extensions of the models, optimization and theoretical analyzing techniques from matrix cases to tensor cases. 
	\item Equipped with the tensor nuclear norm, we theoretically show that under certain incoherence conditions, the solution to the convex TRPCA model (\ref{trpca}) perfectly recovers the underlying low-rank component $\LL_0$ and sparse component $\E_0$. RPCA \cite{RPCA} and its recovery guarantee fall into our special cases. 
	\item We give a new rigorous proof of t-SVD factorization and it leads to a more efficient way for computing t-SVD, so that TRPCA can be solved more efficiently than \cite{lu2016tensorrpca}. We further perform several simulations to corroborate our theoretical results. Numerical experiments on images and videos also show the superiority of TRPCA over RPCA and SNN.
\end{enumerate}

The rest of this paper is structured as follows. Section \ref{sec_notations} gives some notations and preliminaries. Section \ref{sec_TNN} presents the way for defining the tensor nuclear norm induced by the t-product. Section \ref{sec_tcTNN} provides the recovery guarantee of TRPCA and the optimization details. Section \ref{sec_exp} presents numerical experiments conducted on synthetic and real data. We conclude this work in Section \ref{sec_con}.

\section{Notations and Preliminaries}\label{sec_notations}


\subsection{Notations}

In this paper, we denote tensors by boldface Euler script letters, \textit{e.g.}, $\A$. Matrices are denoted by boldface capital letters, \textit{e.g.}, $\Am$; vectors are denoted by boldface lowercase letters, \textit{e.g.}, $\aa$, and scalars are denoted by lowercase letters, \textit{e.g.}, $a$. We denote $\bm I_n$ as the $n\times n$ sized identity matrix. The fields of real numbers and complex numbers are denoted as $\mathbb{R}$ and $\mathbb{C}$, respectively. For a 3-way tensor $\A\in\Cnnn$, we denote its $(i,j,k)$-th entry as $\A_{ijk}$ or $a_{ijk}$ and use the Matlab notation $\A(i,:,:)$, $\A(:,i,:)$ and $\A(:,:,i)$ to denote respectively the $i$-th horizontal, lateral and frontal slice. More often, the frontal slice $\A(:,:,i)$ is denoted compactly as $\Am^{(i)}$. The tube is denoted as $\A(i,j,:)$. The inner product between $\Am$ and $\Bm$ in $\mathbb{C}^{n_1\times n_2}$ is defined as $\inproduct{\Am}{\Bm}=\Tr(\Am^*\Bm)$, where $\Am^*$ denotes the conjugate transpose of $\Am$ and $\Tr(\cdot)$ denotes the matrix trace. The inner product between $\A$ and $\B$ in $\Cnnn$ is defined as $\langle\A,\B\rangle=\sum_{i=1}^{n_3}\inproduct{\Am^{(i)}}{\Bm^{(i)}}$. For any $\A\in\Cnnn$, the complex conjugate of $\A$ is denoted as $\conj(\A)$ which takes the complex conjugate of each entry of $\A$. We denote $\left\lfloor t \right\rfloor$ as the nearest integer less than or equal to $t$ and $\lceil t \rceil$ as the one greater than or equal to $t$.

Some norms of vector, matrix and tensor are used. We denote the $\ell_1$-norm as $\norm{\A}_1=\sum_{ijk}|a_{ijk}|$, the infinity norm as $\norm{\A}_\infty=\max_{ijk}|a_{ijk}|$ and the Frobenius norm as $\norm{\A}_F=\sqrt{\sum_{ijk}|a_{ijk}|^2}$, respectively. The above norms reduce to the vector or matrix norms if $\A$ is a vector or a matrix. For $\vv\in\mathbb{C}^n$, the $\ell_2$-norm is $\norm{\vv}_2 = \sqrt{\sum_{i}|v_{i}|^2}$. The spectral norm of a matrix $\Am$ is denoted as $\norm{\Am} = \max_{i}\sigma_i(\Am)$, where $\sigma_i(\Am)$'s are the singular values of $\Am$. The matrix nuclear norm is $\norm{\Am}_* = \sum_{i}\sigma_i(\Am)$. 

\subsection{Discrete Fourier Transformation}

The Discrete Fourier Transformation (DFT) plays a core role in tensor-tensor product introduced later. We give some related background knowledge and notations here. The DFT on $\vv\in\mathbb{R}^n$, denoted as $\vbar$, is given by
\begin{align}\label{eq_fourtrans}
	\vbar = \F_n\vv \in \mathbb{C}^n,
\end{align}
where $\F_n$ is the DFT matrix defined as
\begin{align*}
	\F_n = 
	\begin{bmatrix}
		1 & 1 & 1 & \cdots & 1 \\
		1 & \omega & \omega^2 & \cdots & \omega^{n-1} \\
		\vdots & \vdots & \vdots & \ddots & \vdots \\
		1 & \omega^{n-1} & \omega^{2(n-1)} & \cdots &\omega^{(n-1)(n-1)}
	\end{bmatrix}\in\mathbb{C}^{n\times n},
\end{align*}
where $\omega = \e^{-\frac{2\pi i}{n}}$ is a primitive $n$-th root of unity in which $i = \sqrt{-1}$. Note that $\F_n/\sqrt{n}$ is an orthogonal matrix, \textit{i.e.},
\begin{align}\label{eq_Fnorthogonal}
	\F_n^*\F_n = \F_n\F^*_n = n\Im_n.
\end{align}
Thus $\F_n^{-1} = \F^*_n/n$. The above property will be frequently used in this paper. Computing $\vbar$ by using (\ref{eq_fourtrans}) costs $O(n^2)$. A more widely used method is the Fast Fourier Transform (FFT) which costs $O(n\log n)$. By using the Matlab command $\mcode{fft}$, we have $\vbar = \mcode{fft}(\vv)$. Denote the circulant matrix of $\vv$ as
\begin{align*}
	\mcode{circ}(\vv) = 
	\begin{bmatrix}
		v_1 & v_n &\cdots &v_2 \\
		v_2 & v_1 & \cdots & v_3 \\
		\vdots & \vdots & \ddots & \vdots \\
		v_n & v_{n-1} & \cdots & v_1
	\end{bmatrix}\in\mathbb{R}^{n\times n}.
\end{align*} 
It is known that it can be diagonalized by the DFT matrix, \textit{i.e.},
\begin{align}\label{eq_diagonalv}
	\F_n \cdot \mcode{circ}(\vv) \cdot \F_n^{-1} = \Diag(\vbar),
\end{align}
where $\Diag(\vbar)$ denotes a diagonal matrix with its $i$-th diagonal entry as $\bar{v}_i$. The above equation implies that the columns of $\F_n$ are the eigenvectors of $(\mcode{circ}(\vv))^\top$ and $\bar{v}_i$'s are the corresponding eigenvalues. 
\begin{lemma}\label{lem_keyprofft}\cite{rojo2004some} 
	Given any real vector $\vv\in\mathbb{R}^n$, the associated $\vbar$ satisfies 
\begin{align}\label{fftvproper}
	\bar{v}_1 \in\mathbb{R} \text{ and } \conj({\bar{v}}_i) = \bar{v}_{n-i+2}, \ i=2,\cdots, \left\lfloor\frac{n+1}{2} \right\rfloor.
\end{align} 
Conversely, for any given complex $\vbar\in \Cn$ satisfying (\ref{fftvproper}), there exists a real block circulant matrix $\mcode{circ}(\vv)$ such that (\ref{eq_diagonalv}) holds.	
\end{lemma}
As will be seen later, the above properties are useful for efficient computation and important for proofs. Now we consider the DFT on tensors. For $\A\in\Rn$, we denote $\Abar\in\Cnnn$ as the result of DFT on $\A$ along the 3-rd dimension, \textit{i.e.}, performing the DFT on all the tubes of $\A$. By using the Matlab command $\mcode{fft}$, we have
\begin{equation*}
	\Abar=\mcode{fft}(\A,[\ ],3).
\end{equation*}
In a similar fashion, we can compute $\A$ from ${\Abar}$ using the inverse FFT, \textit{i.e.},
\begin{equation*}
	\A = \mcode{ifft}({\Abar},[\ ],3).
\end{equation*}
In particular, we denote $\Ambar\in\mathbb{C}^{n_1n_3\times n_2n_3}$ as a block diagonal matrix with its $i$-th block on the diagonal as the $i$-th frontal slice $\Ambar^{(i)}$ of ${\Abar}$, \textit{i.e.},
\begin{equation*}\label{eq_Abardef}
	\Ambar = \bdiag(\Abar) =
	\begin{bmatrix}
		\Ambar^{(1)} & & & \\
		& \Ambar^{(2)} & & \\
		& & \ddots & \\
		& & & \Ambar^{(n_3)}
	\end{bmatrix},
\end{equation*}
where $\bdiag$ is an operator which maps the tensor $\Abar$ to the block diagonal matrix $\Ambar$. Also, we define the block circulant matrix ${\bcirc}(\A)\in\mathbb{R}^{n_1n_3\times n_2n_3}$ of $\A$ as
\begin{align*} 
	{\bcirc}(\A) =
	\begin{bmatrix}
		\Am^{(1)} &\Am^{(n_3)} &\cdots &\Am^{(2)} \\
		\Am^{(2)} &\Am^{(1)} & \cdots &\Am^{(3)} \\
		\vdots & \vdots & \ddots & \vdots \\
		\Am^{(n_3)} & \Am^{(n_3-1)} & \cdots & \Am^{(1)}
	\end{bmatrix}.
\end{align*}
Just like the circulant matrix which can be diagonalized by DFT, the block circulant matrix can be block diagonalized, \textit{i.e.},
\begin{align}\label{dftpro}
	(\F_{n_3} \otimes \bm{I}_{n_1}) \cdot \mcode{bcirc}(\A) \cdot (\F_{n_3}^{-1} \otimes \bm{I}_{n_2}) = \Ambar,
\end{align}
where $\otimes$ denotes the Kronecker product and $(\F_{n_3}\otimes \bm{I}_{n_1})/\sqrt{n_3}$ is orthogonal. By using Lemma \ref{lem_keyprofft}, we have
\begin{equation}\label{keyprofffttensor}
	\begin{cases}
		\Ambar^{(1)} \in \mathbb{R}^{n_1\times n_2}, \\
		\conj({\Ambar}^{(i)}) = \Ambar^{(n_3-i+2)}, \ i=2,\cdots,\left\lfloor\frac{n_3+1}{2} \right\rfloor.
	\end{cases}
\end{equation}
Conversely, for any given $\Abar\in\mathbb{C}^{n_1\times n_2\times n_3}$ satisfying (\ref{keyprofffttensor}), there exists a real tensor $\A\in\Rn$ such that (\ref{dftpro}) holds. Also, by using (\ref{eq_Fnorthogonal}), we have the following properties which will be used frequently:
\begin{align}\label{eq_proFnormNuclear}
 	\norm{\A}_F=\frac{1}{\sqrt{n_3}}\norm{\Ambar}_F, 
\end{align}
\begin{align}\label{eq_proinproduct}
 	\inproduct{\A}{\B}=\frac{1}{n_3}\inproduct{\Ambar}{\Bmbar}.
\end{align}

\subsection{T-product and T-SVD}

For $\A\in\Rn$, we define
\begin{equation*}
	\mcode{unfold}(\A) =
	\begin{bmatrix}
		\Am^{(1)} \\ \Am^{(2)} \\ \vdots \\ \Am^{(n_3)} 
	\end{bmatrix}, \ \mcode{fold}(\mcode{unfold}(\A))=\A,
\end{equation*}
where the $\unfold$ operator maps $\A$ to a matrix of size $n_1n_3\times n_2$ and $\fold$ is its inverse operator. 
\begin{defn} \textbf{(T-product)} \cite{kilmer2011factorization}
	Let $\A\in\Rn$ and $\B\in\mathbb{R}^{n_2\times l\times n_3}$. Then the t-product $\A*\B$ is defined to be a tensor of size $n_1\times l\times n_3$, 
	\begin{equation}\label{tproducdef}
		\A*\B = \mcode{fold}(\mcode{bcirc}(\A)\cdot\mcode{unfold}(\B)).
	\end{equation} 
\end{defn}
The t-product can be understood from two perspectives. First, in the original domain, a 3-way tensor of size $n_1\times n_2\times n_3$ can be regarded as an $n_1\times n_2$ matrix with each entry being a tube that lies in the third dimension. Thus, the t-product is analogous to the matrix multiplication except that the circular convolution replaces the multiplication operation between the elements. Note that the t-product reduces to the standard matrix multiplication when $n_3=1$. This is a key observation which makes our tensor RPCA model shown later involve the matrix RPCA as a special case. Second, the t-product is equivalent to the matrix multiplication in the Fourier domain; that is, $\C=\A*\B$ is equivalent to $\Cmbar=\Ambar\Bmbar$ due to (\ref{dftpro}). Indeed, $\C=\A*\B$ implies
\begin{align}
	  & \unfold(\C) \notag \\ 
	= & \mcode{bcirc}(\A)\cdot\mcode{unfold}(\B) \notag \\
	= & (\F_{n_3}^{-1}\otimes \bm{I}_{n_1}) \cdot ( (\F_{n_3}\otimes \bm{I}_{n_1})\cdot \mcode{bcirc}(\A) \cdot (\F_{n_3}^{-1}\otimes \bm{I}_{n_2})) \notag \\
	  & \cdot ((\F_{n_3}\otimes \bm{I}_{n_2})\cdot \mcode{unfold}(\B))\label{eqn_tproducomputproer} \\
	= & (\F_{n_3}^{-1}\otimes \bm{I}_{n_1})\cdot\Ambar\cdot\unfold(\Bbar),\notag
\end{align}
where (\ref{eqn_tproducomputproer}) uses (\ref{dftpro}). Left multiplying both sides with 
$(\F_{n_3}\otimes \bm{I}_{n_1})$ leads to $\unfold(\Cbar)=\Ambar\cdot\unfold(\Bbar)$. This is equivalent to $\Cmbar=\Ambar\Bmbar$. This property suggests an efficient way based on FFT to compute t-product instead of using (\ref{tproducdef}). See Algorithm \ref{alg_ttprod}.
\begin{algorithm}[!t]
	\caption{Tensor-Tensor Product}
	\textbf{Input:} $\A\in\Rn$, $\B\in\mathbb{R}^{n_2\times l\times n_3}$.\\
	\textbf{Output:} $\C = \A * \B\in\mathbb{R}^{n_1\times l\times n_3}$.
	\begin{enumerate}[1.]
		\item Compute $\Abar=\fft(\A,[\ ],3)$ and $\Bbar = \fft(\B,[\ ],3)$.
		\item Compute each frontal slice of $\Cbar$ by
		\begin{equation*}
			\Cmbar^{(i)} =
			\begin{cases}
				\Ambar^{(i)}\Bmbar^{(i)}, \quad & i=1,\cdots, \lceil \frac{n_3+1}{2}\rceil,\\
				\conj(\Cmbar^{(n_3-i+2)}), \quad & i=\lceil \frac{n_3+1}{2}\rceil+1,\cdots, n_3.
			\end{cases}
		\end{equation*}
		\item Compute $\C=\ifft(\Cbar,[\ ],3)$.
	\end{enumerate}
	\label{alg_ttprod}	
\end{algorithm} 

The t-product enjoys many similar properties to the matrix-matrix product. For example, the t-product is associative, \textit{i.e.}, $\A*(\B*\C) = (\A*\B)*\C$. We also need some other concepts on tensors extended from the matrix cases. 
\begin{defn} \textbf{(Conjugate transpose)} 
	The conjugate transpose of a tensor $\A\in\Cnnn$ is the tensor $\A^*\in\mathbb{C}^{n_2\times n_1\times n_3}$ obtained by conjugate transposing each of the frontal slices and then reversing the order of transposed frontal slices 2 through $n_3$.
\end{defn}
	The tensor conjugate transpose extends the tensor transpose \cite{kilmer2011factorization} for complex tensors. As an example, let $\A\in\mathbb{C}^{n_1\times n_2\times 4}$ and its frontal slices be $\Am_1$, $\Am_2$, $\Am_3$ and $\Am_4$. Then
\begin{align*}
	\A^* = \fold\left(\begin{bmatrix} \Am_1^* \\ \Am_4^* \\ \Am_3^* \\ \Am_2^* \end{bmatrix}\right).
\end{align*}
 
\begin{defn} \textbf{(Identity tensor)} \cite{kilmer2011factorization}
	The identity tensor $\I\in\mathbb{R}^{n\times n\times n_3}$ is the tensor with its first frontal
	slice being the $n \times n$ identity matrix, and other frontal slices being all zeros.
\end{defn}
It is clear that $\A *\I = \A$ and $\I * \A = \A$ given the appropriate dimensions. The tensor $\Ibar=\fft(\I,[\ ],3)$ is a tensor with each frontal slice being the identity matrix.
\begin{defn} \textbf{(Orthogonal tensor)} \cite{kilmer2011factorization}
	A tensor $\Q\in\mathbb{R}^{n\times n\times n_3}$ is orthogonal if it satisfies $\Q^**\Q=\Q*\Q^*=\I$.
\end{defn}
\begin{defn}\textbf{(F-diagonal Tensor)} \cite{kilmer2011factorization}
	A tensor is called f-diagonal if each of its frontal slices is a diagonal matrix. 
\end{defn}
\begin{thm} \textbf{(T-SVD)} \label{thmtsvd}
	Let $\A\in\Rn$. Then it can be factorized as
	\begin{equation}\label{eq_tsvd}
		\A=\U*\Sbm*\V^*,
	\end{equation}
	where $\U\in \mathbb{R}^{n_1\times n_1\times n_3}$, $\V\in\mathbb{R}^{n_2\times n_2\times n_3}$ are orthogonal, and $\Sbm\in\mathbb{R}^{n_1\times n_2\times n_3}$ is an f-diagonal tensor. 
\end{thm}
\begin{proof} 
	The proof is by construction. Recall that (\ref{dftpro}) holds and $\Ambar^{(i)}$'s satisfy the property (\ref{keyprofffttensor}). Then we construct the SVD of each $\Ambar^{(i)}$ in the following way. For $i=1,\cdots, \lceil \frac{n_3+1}{2}\rceil$, let $\Ambar^{(i)} = \Umbar^{(i)}\Smbar^{(i)}(\Vmbar^{(i)})^*$ be the full SVD of $\Ambar^{(i)}$. Here the singular values in $\Smbar^{(i)}$ are real. For $i=\lceil \frac{n_3+1}{2}\rceil+1,\cdots, n_3$, let $\Umbar^{(i)}=\conj(\Umbar^{(n_3-i+2)})$, $\Smbar^{(i)}=\Smbar^{(n_3-i+2)}$ and $\Vmbar^{(i)}=\conj(\Vmbar^{(n_3-i+2)})$. Then, it is easy to verify that $\Ambar^{(i)} = \Umbar^{(i)}\Smbar^{(i)}(\Vmbar^{(i)})^*$ gives the full SVD of $\Ambar^{(i)}$ for $i=\lceil \frac{n_3+1}{2}\rceil+1,\cdots, n_3$. Then,
	\begin{equation}\label{profTSVDeqn1}
		\Ambar = \Umbar \Smbar \Vmbar^*.
	\end{equation}
	By the construction of $\Umbar$, $\Smbar$ and $\Vmbar$, and Lemma \ref{lem_keyprofft}, we have that $(\F_{n_3}^{-1}\otimes \bm{I}_{n_1}) \cdot \Umbar \cdot (\F_{n_3}\otimes \bm{I}_{n_1})$, $(\F_{n_3}^{-1}\otimes \bm{I}_{n_1}) \cdot \Smbar \cdot (\F_{n_3}\otimes \bm{I}_{n_2})$ and $(\F_{n_3}^{-1}\otimes \bm{I}_{n_2}) \cdot \Vmbar \cdot (\F_{n_3}\otimes \bm{I}_{n_2})$ are real block circulant matrices. Then we can obtain an expression for $\bcirc(\A)$ by applying the appropriate matrix $(\F_{n_3}^{-1}\otimes \bm{I}_{n_1})$ to the left and the appropriate matrix $(\F_{n_3}\otimes \bm{I}_{n_2})$ to the right of each of the matrices in (\ref{profTSVDeqn1}), and folding up the result. This gives a decomposition of the form $\U*\Sbm*\V^*$, where $\U$, $\Sbm$ and $\V$ are real. 	
\end{proof}

\begin{figure}[!t]
	\centering
	\includegraphics[width=0.35\textwidth]{fig/fig_tsvd.pdf}
	\caption{\small An illustration of the t-SVD of an $n_1\times n_2\times n_3$ tensor \cite{hao2013facial}.}
	\label{fig_tsvd}
	\vspace{-0.4cm}
\end{figure}

Theorem \ref{thmtsvd} shows that any 3 way tensor can be factorized into 3 components, including 2 orthogonal tensors and an f-diagonal tensor. See Figure \ref{fig_tsvd} for an intuitive illustration of the t-SVD factorization. T-SVD reduces to the matrix SVD when $n_3=1$. We would like to emphasize that the result of Theorem \ref{thmtsvd} was given first in \cite{kilmer2011factorization} and later some other related works \cite{hao2013facial,martin2013order}. But their proof and the way for computing $\U$ and $\V$ are not rigorous. The issue is that their method cannot guarantee that $\U$ and $\V$ are real tensors. They construct each frontal slice $\Umbar^{(i)}$ (or $\Vmbar^{(i)}$) of $\Ubar$ (or $\Vbar$ resp.) from the SVD of $\Ambar^{(i)}$ independently for all $i=1,\cdots,n_3$. However, the matrix SVD is not unique. Thus, $\Umbar^{(i)}$'s and $\Vmbar^{(i)}$'s may not satisfy property 
(\ref{keyprofffttensor}) even though $\Ambar^{(i)}$'s do. In this case, the obtained $\U$ (or $\V$) from the inverse DFT of $\Ubar$ (or $\Vbar$ resp.) may not be real. Our proof above instead uses property (\ref{keyprofffttensor}) to construct $\U$ and $\V$ and thus avoids this issue. Our proof further leads to a more efficient way for computing t-SVD shown in Algorithm \ref{alg_tsvd}. Note that existing works require computing $n_3$ matrix SVDs while we reduce this number to $\lceil \frac{n_3+1}{2} \rceil$. This is a significant improvement when $n_3$ is large as the t-SVD computing takes the main cost in solving TRPCA.
\begin{algorithm}[!h]
	\caption{T-SVD}
	\textbf{Input:} $\A\in\Rn$.\\
	\textbf{Output:} T-SVD components $\U$, $\Sbm$ and $\V$ of $\A$.
	\begin{enumerate}[1.]
		\item Compute $\Abar=\fft(\A,[\ ],3)$.
		\item Compute each frontal slice of $\Ubar$, $\Sbar$ and $\Vbar$ from $\Abar$ by
		
		\textbf{for} $i=1,\cdots, \lceil 		 \frac{n_3+1}{2}\rceil$ \textbf{do}
		
		\hspace*{0.4cm} $[\Umbar^{(i)},\Smbar^{(i)},\Vmbar^{(i)}] = \text{SVD}(\Ambar^{(i)})$;
		
		\textbf{end for}
		
		\textbf{for} $i=\lceil \frac{n_3+1}{2}\rceil+1,\cdots, n_3$ \textbf{do}
		
		\hspace*{0.4cm}$\Umbar^{(i)} = \conj(\Umbar^{(n_3-i+2)})$;
		 
		\hspace*{0.4cm}$\Smbar^{(i)} = \Smbar^{(n_3-i+2)}$;
		
		\hspace*{0.4cm}$\Vmbar^{(i)} = \conj(\Vmbar^{(n_3-i+2)})$;
		
		\textbf{end for}
		
		\item Compute $\U=\ifft(\Ubar,[\ ],3)$, $\Sbm=\ifft(\Sbar,[\ ],3)$, and $\V=\ifft(\Vbar,[\ ],3)$.
	\end{enumerate}
	\label{alg_tsvd}	
\end{algorithm} 

It is known that the singular values of a matrix have the decreasing order property. Let $\A=\U*\Sbm*\V^*$ be the t-SVD of $\A\in\Rn$. The entries on the diagonal of the first frontal slice $\Sbm(:,:,1)$ of $\Sbm$ have the same decreasing property, \textit{i.e.}, 
\begin{align}\label{eqndecreasingvalue}
	\Sbm(1,1,1)\geq \Sbm(2,2,1)\geq \cdots \geq \Sbm(n',n',1) \geq 0,
\end{align}
where $n' = \min(n_1,n_2)$. The above property holds since the inverse DFT gives 
\begin{equation}\label{prosandsbar}
	\Sbm(i,i,1)=\frac{1}{n_3}\sum_{j=1}^{n_3}\Sbar(i,i,j),
\end{equation}
and the entries on the diagonal of $\Sbar(:,:,j)$ are the singular values of $\Abar(:,:,j)$. As will be seen in Section \ref{sec_TNN}, the tensor nuclear norm depends only on the first frontal slice $\Sbm(:,:,1)$. Thus, we call the entries on the diagonal of $\Sbm(:,:,1)$ as the singular values of $\A$.
\begin{defn} \textbf{(Tensor tubal rank)} \cite{kilmer2013third,zhang2014novel}
	For $\A\in\Rn$, the tensor {tubal rank}, denoted as $\rankt(\A)$, is defined as the number of nonzero singular tubes of $\Sbm$, where $\Sbm$ is from the t-SVD of $\A=\U*\Sbm*\V^*$. We can write
	\begin{align*}
		\rankt(\A) = & \#\{i,\Sbm(i,i,:)\neq\bm{0}\}.
	\end{align*} 
\end{defn}
By using property (\ref{prosandsbar}), the tensor tubal rank is determined by the first frontal slice $\Sbm(:,:,1)$ of $\Sbm$, \textit{i.e.},
\begin{align*}
	\rankt(\A) = \#\{i,\Sbm(i,i,1)\neq {0}\}.	\label{eqndefturank}	
\end{align*} 
Hence, the tensor tubal rank is equivalent to the number of nonzero singular values of $\A$. This property is the same as the matrix case. Define $\A_k=\sum_{i=1}^{k}\U(:,i,:)*\Sbm(i,i,:)*\V(:,i,:)^*$ for some $k<\min(n_1,n_2)$. Then $\A_k = \arg\min_{\rankt(\tilde{\A})\leq k} \norm{\A-\tilde{\A}}_F$, so $\A_k$ is the best approximation of $\A$ with the tubal rank at most $k$. It is known that the real color images can be well approximated by low-rank matrices on the three channels independently. If we treat a color image as a three way tensor with each channel corresponding to a frontal slice, then it can be well approximated by a tensor of low tubal rank. A similar observation was found in \cite{hao2013facial} with the application to facial recognition. Figure \ref{fig_lenna} gives an example to show that a color image can be well approximated by a low tubal rank tensor since most of the singular values of the corresponding tensor are relatively small.

In Section \ref{sec_TNN}, we will define a new tensor nuclear norm which is the convex surrogate of the tensor average rank defined as follows. This rank is closely related to the tensor tubal rank.

\begin{defn} \textbf{(Tensor average rank)} 
	For $\A\in\Rn$, the tensor average rank, denoted as $\ranka(\A)$, is defined as
	\begin{align}
		\ranka(\A) = \frac{1}{n_3} \rankm(\bcirc(\A)).
	\end{align}
\end{defn}
The above definition has a factor $\frac{1}{n_3}$. Note that this factor is crucial in this work as it guarantees that the convex envelope of the tensor average rank within a certain set is the tensor nuclear norm defined in Section \ref{sec_TNN}. The underlying reason for this factor is the t-product definition. Each element of $\A$ is repeated $n_3$ times in the block circulant matrix $\bcirc(\A)$ used in the t-product. Intuitively, this factor alleviates such an entries expansion issue.

\begin{figure}[!t]
	\centering
	\begin{subfigure}[b]{0.15\textwidth}
		\centering
		\includegraphics[width=\textwidth]{fig/lenaoriginal.pdf}
		\caption{ }
	\end{subfigure} 
	\begin{subfigure}[b]{0.15\textwidth}
		\centering
		\includegraphics[width=\textwidth]{fig/lena_r=50.pdf}
		\caption{ }
	\end{subfigure}
	\begin{subfigure}[b]{0.165\textwidth}
		\centering
		\includegraphics[width=\textwidth]{fig/fig_lenna_sin_values.pdf}
		\caption{ }
	\end{subfigure}
	\caption{\small{Color images can be approximated by low tubal rank tensors. (a) A color image can be modeled as a tensor $\M \in \mathbb{R}^{512\times 512\times 3}$; (b) approximation by a tensor with tubal rank $r=50$; (c) plot of the singular values of $\M$.}}
	\label{fig_lenna}
\end{figure}

There are some connections between different tensor ranks and these properties imply that the low tubal rank or low average rank assumptions are reasonable for their applications in real visual data. 
First, $\ranka(\A) \leq \rankt(\A)$. Indeed, 
\begin{equation*}
	\ranka(\A) = \frac{1}{n_3} \rankm(\Ambar) \leq \max_{i=1,\cdots,n_3} \rankm(\Ambar^{(i)}) = \rankt(\A),
\end{equation*}
where the first equality uses (\ref{dftpro}). This implies that a low tubal rank tensor always has low average rank. Second, let $\text{rank}_{\text{tc}}(\A)=\left( \text{rank}(\Am^{\{1\}}), \text{rank}(\Am^{\{2\}}), \text{rank}(\Am^{\{3\}}) \right)$, where $\Am^{\{i\}}$ is the mode-$i$ matricization of $\A$, be the Tucker rank of $\A$. Then $\ranka(\A)\leq \rankm(\Am^{\{1\}})$. This implies that a tensor with low Tucker rank has low average rank. The low Tucker rank assumption used in some  applications, \textit{e.g.}, image completion \cite{liu2013tensor}, is applicable to the low average rank assumption. Third, if the CP rank of $\A$ is $r$, then its tubal rank is at most $r$ \cite{zhang2015exact}. Let $\A=\sum_{i=1}^{r} \aa_i^{(1)} \circ \aa_i^{(2)} \circ \aa_i^{(3)}$, where $\circ$ denotes the outer product, be the CP decomposition of $\A$. Then $\Abar=\sum_{i=1}^{r} \aa_i^{(1)} \circ \aa_i^{(2)} \circ \bar{\aa}_i^{(3)}$, where $\bar{\aa}_i^{(3)} = \fft(\aa_i^{(3)})$. So $\Abar$ has the CP rank at most $r$, and each frontal slice of $\Abar$ is the sum of $r$ rank-1 matrices. Thus, the tubal rank of $\A$ is at most $r$. In summary, we show that the low average rank assumption is weaker than the low Tucker rank and low CP rank assumptions. 

\section{Tensor Nuclear Norm (TNN)}
\label{sec_TNN}

In this section, we propose a new tensor nuclear norm which is a convex surrogate of tensor average rank. Based on t-SVD, one may have many different ways to define the tensor nuclear norm intuitively. We give a new and rigorous way to deduce the tensor nuclear norm from the t-product, such that the concepts and their properties are consistent with the matrix cases. This is important since it guarantees that the theoretical analysis of the tensor nuclear norm based tensor RPCA model in Section \ref{sec_tcTNN} can be done in a similar way to RPCA. Figure \ref{fig_tensoroperators} summarizes the way for the new definitions and their relationships. It begins with the known operator norm \cite{atkinson2009theoretical} and t-product. First, the tensor spectral norm is induced by the tensor operator norm by treating the t-product as an operator. Then the tensor nuclear norm is defined as the dual norm of the tensor spectral norm. Finally, we show that the tensor nuclear norm is the convex envelope of the tensor average rank within the unit ball of the tensor spectral norm.

\begin{figure}[!t]
	\centering
	\includegraphics[width=0.4\textwidth]{fig/fig_tensoroperators.pdf}
	\caption{\small An illustration of the way to define the tensor nuclear norm and the relationship with other tensor concepts. First, the tensor operator norm is a special case of the known operator norm performed on the tensors. The tensor spectral norm is induced by the tensor operator norm by treating the tensor-tensor product as an operator. Then the tensor nuclear norm is defined as the dual norm of the tensor spectral norm. We also define the tensor average rank and show that its convex envelope is the tensor nuclear norm within the unit ball of the tensor spectral norm. As detailed in Section \ref{sec_TNN}, the tensor spectral norm, tensor nuclear norm and tensor average rank are also defined on the matricization of the tensor.}
	\label{fig_tensoroperators}
\end{figure}

Let us first recall the concept of operator norm \cite{atkinson2009theoretical}. Let $(V,\norm{\cdot}_V)$ and $(W,\norm{\cdot}_W)$ be normed linear spaces and $L: V\rightarrow W$ be the bounded linear operator between them, respectively. The operator norm of $L$ is defined as
\begin{align}\label{eq_generaloperatornnorm}
	\norm{L} = \sup_{\norm{\vv}_V\leq1} \norm{L(\vv)}_W.
\end{align}
Let $V=\mathbb{C}^{n_2}$, $W=\mathbb{C}^{n_1}$ and $L(\vv) = \Am\vv$, $\vv\in V$, where $\Am\in\mathbb{C}^{n_1\times n_2}$. Based on different choices of $\norm{\cdot}_V$ and $\norm{\cdot}_W$, many matrix norms can be induced by the operator norm in (\ref{eq_generaloperatornnorm}). For example, if $\norm{\cdot}_V$ and $\norm{\cdot}_W$ are $\norm{\cdot}_F$, then the operator norm (\ref{eq_generaloperatornnorm}) reduces to the matrix spectral norm.

Now, consider the normed linear spaces $(V,\norm{\cdot}_F)$ and $(W,\norm{\cdot}_F)$, where $V=\mathbb{R}^{n_2\times 1\times n_3}$, $W=\mathbb{R}^{n_1\times 1\times n_3}$, and $\LL: V\rightarrow W$ is a bounded linear operator. In this case, (\ref{eq_generaloperatornnorm}) reduces to the tensor operator norm
\begin{align}\label{tensoroperatornorm}
	\norm{\LL} = \sup_{\norm{\V}_F\leq1} \norm{\LL(\V)}_F.
\end{align}
As a special case, if $\LL(\V) = \A*\V$, where $\A\in\Rn$ and $\V\in V$, then the tensor operator norm (\ref{tensoroperatornorm}) gives the tensor spectral norm, denoted as $\norm{\A}$,
\begin{align}
	\norm{\A} 
	:= & \sup_{\norm{\V}_F\leq1} \norm{\A*\V}_F \notag\\
	 = & \sup_{\norm{\V}_F\leq1} \norm{\bcirc(\A)\cdot\unfold(\V)}_F \label{eq_deriveoper1}\\
	 = & \norm{\bcirc(\A)} \label{eq_deriveoper2},
\end{align}
where (\ref{eq_deriveoper1}) uses (\ref{tproducdef}), and (\ref{eq_deriveoper2}) uses the definition of matrix spectral norm.
\begin{defn}
	\textbf{(Tensor spectral norm)} The tensor spectral norm of $\A\in\Rn$ is defined as $\norm{\A} := \norm{\bcirc(\A)}$.
\end{defn}
\begin{table*}[!t]
	\centering
	\caption{Parallelism of sparse vector, low-rank matrix and low-rank tensor.}\label{tab_sparselowmlowt}
	\begin{tabular}{c||c|c|c}
		\hline
					   & Sparse vector     & Low-rank matrix  & Low-rank tensor (this work)\\ \hline
		Degeneracy of  & 1-D signal  $\x\in\mathbb{R}^n$      & 2-D correlated signals $\Xm\in\mathbb{R}^{n_1\times n_2}$ & 3-D correlated signals $\X\in\mathbb{R}^{n_1\times n_2\times n_3}$\\ \hline
		Parsimony concept & cardinality    & rank   		  & tensor average rank{\scriptsize\footnotemark}\\ \hline
		Measure  	   & $\ell_0$-norm $\norm{\x}_0$   & $\text{rank}(\Xm)$   & $\ranka(\X)$ \\ \hline
		Convex surrogate  & $\ell_1$-norm $\norm{\x}_1$ & nuclear norm $\norm{\Xm}_*$ & tensor nuclear norm $\norm{\X}_*$ \\ \hline
		Dual norm      & $\ell_\infty$-norm $\norm{\x}_\infty$ & spectral norm $\norm{\Xm}$ & tensor spectral norm $\norm{\X}$ \\ \hline
	\end{tabular}\\
	\footnotesize\footnotemark[2]{Strictly speaking, the tensor tubal rank, which bounds the tensor average rank, is also the parsimony concept of the low-rank tensor.} 
	\vspace{-0.5cm}
\end{table*}
By (\ref{eq_Fnorthogonal}) and (\ref{dftpro}), we have 
\begin{align}\label{eq_protsc}
	\norm{\A} = \norm{\bcirc(\A)} = \norm{\Ambar}.
\end{align}
This property is frequently used in this work. It is known that the matrix nuclear norm is the dual norm of the matrix spectral norm. Thus, we define the tensor nuclear norm, denoted as $\norm{\A}_*$, as the dual norm of the tensor spectral norm. For any $\B\in\Rn$ and $\Bmtilde\in\mathbb{C}^{n_1n_3\times n_2n_3}$, we have
\begin{align}
	\norm{\A}_* 
	:= & \sup_{\norm{\B}\leq1} \langle{\A},{\B}\rangle \label{eq_drivenuclearnorm0}\\
	 =&\sup_{\norm{\Bmbar}\leq1}\frac{1}{n_3}\langle{\Ambar},{\Bmbar}\rangle\label{eq_drivenuclearnorm1}\\
	 \leq &\frac{1}{n_3} \sup_{\norm{\Bmtilde}\leq1} |\langle{\Ambar},{\Bmtilde}\rangle|\label{eq_drivenuclearnorm2} \\
	 = &\frac{1}{n_3}\norm{\Ambar}_*,\label{eq_drivenuclearnorm3}\\
	 = & \frac{1}{n_3}\norm{\bcirc(\A)}_*,\label{eq_drivenuclearnorm4}
\end{align}
where (\ref{eq_drivenuclearnorm1}) is from (\ref{eq_proinproduct}), (\ref{eq_drivenuclearnorm2}) is due to the fact that $\Bmbar$ is a block diagonal matrix in $\mathbb{C}^{n_1n_3\times n_2n_3}$ while $\Bmtilde$ is an arbitrary matrix in $\mathbb{C}^{n_1n_3\times n_2n_3}$, (\ref{eq_drivenuclearnorm3}) uses the fact that the matrix nuclear norm is the dual norm of the matrix spectral norm, and (\ref{eq_drivenuclearnorm4}) uses (\ref{dftpro}) and~(\ref{eq_Fnorthogonal}). Now we show that there exists $\B\in\Rn$ such that the equality (\ref{eq_drivenuclearnorm2}) holds and thus $\norm{\A}_* = \frac{1}{n_3}\norm{\bcirc(\A)}_*$. Let $\A=\U*\Sbm*\V^*$ be the t-SVD of $\A$ and $\B=\U*\V^*$. We have
\begin{align}
	\langle{\A},{\B}\rangle 
	= & \langle{\U*\Sbm*\V^*},{\U*\V^*}\rangle \label{eqntnn1}\\
	= & \frac{1}{n_3} \inproduct{\overline{\U*\Sbm*\V^*}}{\overline{\U*\V^*}} \notag\\
	= & \frac{1}{n_3} \inproduct{\Umbar\Smbar\Vmbar^*}{\Umbar\Vmbar^*} = \frac{1}{n_3} \Tr(\Smbar)\notag\\
	= & \frac{1}{n_3} \norm{\Ambar}_* = \frac{1}{n_3} \norm{\bcirc(\A)}_*. \label{eqntnn2}
\end{align}
Combining (\ref{eq_drivenuclearnorm0})-(\ref{eq_drivenuclearnorm4}) and (\ref{eqntnn1})-(\ref{eqntnn2}) leads to $\norm{\A}_* = \frac{1}{n_3} \norm{\bcirc(\A)}_*$. On the other hand, by (\ref{eqntnn1})-(\ref{eqntnn2}), we have 
\begin{align}
	\norm{\A}_* 
	= & \langle{\U*\Sbm*\V^*},{\U*\V^*}\rangle \notag\\
 	= & \langle{\U^**\U*\Sbm},{\V^**\V}\rangle \notag\\
 	= & \langle{\Sbm},\I\rangle = \sum_{i=1}^{r} \Sbm(i,i,1), \label{defsipro}
\end{align}
where $r=\rankt(\A)$ is the tubal rank. Thus, we have the following definition of tensor nuclear norm.
\begin{defn}\label{defntnnours}
	\textbf{(Tensor nuclear norm)} Let $\A=\U*\Sbm*\V^*$ be the t-SVD of $\A\in\Rn$. The tensor nuclear norm of $\A$ is defined as
	\begin{equation*}
		\norm{\A}_*:= \inproduct{\Sbm}{\I} = \sum_{i=1}^{r} \Sbm(i,i,1),
	\end{equation*}
	where $r=\rankt(\A)$.
\end{defn}
From (\ref{defsipro}), it can be seen that only the information in the first frontal slice of $\Sbm$ is used when defining the tensor nuclear norm. Note that this is the first work which directly uses the singular values $\Sbm(:,:,1)$ of a tensor to define the tensor nuclear norm. Such a definition makes it consistent with the matrix nuclear norm. The above TNN definition is also different from existing works \cite{lu2016tensorrpca,zhang2014novel,semerci2014tensor}.
 
It is known that the matrix nuclear norm $\norm{\Am}_*$ is the convex envelope of the matrix rank $\rankm(\Am)$ within the set $\{\Am |\norm{\Am}\leq1\}$~\cite{fazel2002matrix}. Now we show that the tensor average rank and tensor nuclear norm have the same relationship. 
\begin{thm}\label{thm_convexenvelope}
	On the set $\{\A\in\Rn |\norm{\A}\leq1\}$, the convex envelope of the tensor average rank $\ranka(\A)$ is the tensor nuclear norm $\norm{\A}_*$.
\end{thm}
We would like to emphasize that the proposed tensor spectral norm, tensor nuclear norm and tensor ranks are not arbitrarily defined. They are rigorously induced by the t-product and t-SVD. These concepts and their relationships are consistent with the matrix cases. This is important for the proofs, analysis and computation in optimization. Table \ref{tab_sparselowmlowt} summarizes the parallel concepts in sparse vector, low-rank matrix and low-rank tensor. With these elements in place, the existing proofs of low-rank matrix recovery provide a template for the more general case of low-rank tensor recovery.
 
Also, from the above discussions, we have the property
\begin{align}\label{eqpropertytnn1}
	\norm{\A}_* = \frac{1}{n_3} \norm{\bcirc(\A)}_* = \frac{1}{n_3} \norm{\Ambar}_*. 
\end{align}
It is interesting to understand the tensor nuclear norm from the perspectives of $\bcirc(\A)$ and $\Ambar$. The block circulant matrix can be regarded as a new way of matricization of $\A$ in the original domain. The frontal slices of $\A$ are arranged in a circulant way, which is expected to preserve more spacial relationships across frontal slices, compared with previous matricizations along a single dimension. Also, the block diagonal matrix $\Ambar$ can be regarded as a matricization of $\A$ in the Fourier domain. Its blocks on the diagonal are the frontal slices of $\Abar$, which contains the information across frontal slices of $\A$ due to the DFT on $\A$ along the third dimension. So $\bcirc(\A)$ and $\Ambar$ play a similar role to matricizations of $\A$ in different domains. Both of them capture the spacial information within and across frontal slices of $\A$. This intuitively supports our tensor nuclear norm definition.

Let $\Am=\Um\Sm\Vm^*$ be the skinny SVD of $\Am$. It is known that any subgradient of the nuclear norm at $\Am$ is of the form $\Um\Vm^*+\Wm$, where $\Um^*\Wm=\0$, $\Wm\Vm=\0$ and $\norm{\Wm}\leq 1$ \cite{watson1992characterization}. Similarly, for $\A\in\Rn$ with tubal rank $r$, we also have the skinny t-SVD, \textit{i.e.}, $\A=\U*\Sbm*\V^*$, where $\U\in\mathbb{R}^{n_1\times r\times n_3}$, $\Sbm\in\mathbb{R}^{r\times r\times n_3}$, and $\V\in\mathbb{R}^{n_2\times r\times n_3}$, in which $\U^**\U=\I$ and $\V^**\V=\I$. The skinny t-SVD will be used throughout this paper. With skinny t-SVD, we introduce the subgradient of the tensor nuclear norm, which plays an important role in the proofs. 

\begin{thm}\textbf{(Subgradient of tensor nuclear norm)}\label{thmsubgradient}
	Let $\A\in\Rn$ with $\rankt(\A)=r$ and its skinny t-SVD be $\A=\U*\Sbm*\V^*$. The subdifferential (the set of subgradients) of $\norm{\A}_*$ is $\partial \norm{\A}_*= \{\U*\V^*+\W |\U^**\W=\0, \W*\V=\0,\norm{\W}\leq 1 \}$.
\end{thm}
 
\section{Exact Recovery Guarantee of TRPCA}\label{sec_tcTNN}

With TNN defined above, we now consider the exact recovery guarantee of TRPCA in (\ref{trpca}). The problem we study here is to recover a low tubal rank tensor $\LL_0$ from highly corrupted measurements $\X=\LL_0+\Sbm_0$. In this section, we show that under certain assumptions, the low tubal rank part $\LL_0$ and sparse part $\Sbm_0$ can be exactly recovered by solving convex program (\ref{trpca}). We will also give the optimization detail for solving (\ref{trpca}).

\subsection{Tensor Incoherence Conditions}

Recovering the low-rank and sparse components from their sum suffers from an identifiability issue. For example, the tensor $\X$, with $x_{ijk}=1$ when $i=j=k=1$ and zeros everywhere else, is both low-rank and sparse. One is not able to identify the low-rank component and the sparse component in this case. To avoid such pathological situations, we need to assume that the low-rank component $\LL_0$ is not sparse. To this end, we assume $\LL_0$ to satisfy some incoherence conditions. We denote $\mathring{\mathfrak{e}}_i$ as the tensor column basis, which is a tensor of size $n_1\times 1\times n_3$ with its $(i,1,1)$-th entry equaling  1 and the rest equaling 0 \cite{zhang2015exact}. We also define the tensor tube basis $\dot{\mathfrak{e}}_k$, which is a tensor of size $1\times 1\times n_3$ with its $(1,1,k)$-th entry equaling  1 and the rest equaling 0.

\begin{defn} \textbf{(Tensor Incoherence Conditions)} 
	For $\LL_0\in\Rn$, assume that $\rankt(\LL_0)=r$ and it has the skinny t-SVD $\LL_0=\U*\Sbm*\V^*$, where $\U\in\mathbb{R}^{n_1\times r\times n_3}$ and
	$\V\in\mathbb{R}^{n_2\times r\times n_3}$ satisfy $\U^**\U=\I$ and $\V^**\V=\I$, and 
	$\Sbm\in\mathbb{R}^{r\times r\times n_3}$ is an f-diagonal tensor. Then $\LL_0$ is said to satisfy the tensor incoherence conditions with parameter $\mu$ if
	\begin{align}
		\max_{i=1,\cdots,n_1} \norm{\U^**\mathring{\mathfrak{e}}_i}_F\leq\sqrt{\frac{\mu r}{n_1n_3}}, \label{tic1}\\
		\max_{j=1,\cdots,n_2} \norm{\V^**\mathring{\mathfrak{e}}_j}_F\leq\sqrt{\frac{\mu r}{n_2n_3}},\label{tic2}
	\end{align}
	\begin{equation}
		\norm{\U*\V^*}_\infty\leq \sqrt{\frac{\mu r}{n_1n_2n_3^2}}.\label{tic3}
	\end{equation}	
\end{defn}
The exact recovery guarantee of RPCA \cite{RPCA} also requires some incoherence conditions. Due to property (\ref{eq_proFnormNuclear}), conditions (\ref{tic1})-(\ref{tic2}) have equivalent matrix forms in the Fourier domain, and they are intuitively similar to the matrix incoherence conditions (1.2) in \cite{RPCA}. But the joint incoherence condition (\ref{tic3}) is more different from the matrix case (1.3) in \cite{RPCA}, since it does not have an equivalent matrix form in the Fourier domain. As observed in \cite{chen2013incoherence}, the joint incoherence condition is not necessary for low-rank matrix completion. However, for RPCA, it is unavoidable for polynomial-time algorithms. In our proofs, the joint incoherence (\ref{tic3}) condition is necessary. Another identifiability issue arises if the sparse tensor $\Sbm_0$ has low tubal rank. This can be avoided by assuming that the support of $\Sbm_0$ is uniformly distributed.

\subsection{Main Results}
Now we show that the convex program (\ref{trpca}) is able to perfectly recover the low-rank and sparse components. We define $n_{(1)}=\max(n_1,n_2)$ and $n_{(2)}=\min ({n_1,n_2})$.
\begin{thm}\label{thm1}
	Suppose that $\LL_0\in \mathbb{R}^{n\times n\times n_3}$ obeys (\ref{tic1})-(\ref{tic3}). Fix any $\nss$ tensor $\M$ of signs. Suppose that the support set $\Omegat$ of $\Sbm_0$ is uniformly distributed among all sets of cardinality $m$, and that $\sgn{[\Sbm_0]_{ijk}}=[\M]_{ijk}$ for all $(i,j,k)\in\Omegat$. Then, there exist universal constants $c_1, c_2>0$ such that with probability at least $1-c_1(nn_3)^{-c_2}$ (over the choice of support of $\Sbm_0$), $(\LL_0,\Sbm_0)$ is the unique minimizer to (\ref{trpca}) with $\lambda = 1/\sqrt{nn_3}$, provided that
	\begin{equation}
		\rankt(\LL_0)\leq \frac{\rho_r nn_3}{\mu(\log(nn_3))^{2}} \text{ and } m\leq \rho_sn^2n_3,
	\end{equation}
	where $\rho_r$ and $\rho_s$ are positive constants. If $\LL_0\in\Rn$ has rectangular frontal slices, TRPCA with $\lambda = 1/\sqrt{\none n_3}$ succeeds with probability at least $1-c_1(\none n_3)^{-c_2}$, provided that $\rankt(\LL_0)\leq \frac{ \rho_r \ntwo n_3}{\mu(\log(\none n_3))^{2}} \text{ and } m\leq \rho_sn_1n_2 n_3$.
\end{thm}
The above result shows that for incoherent $\LL_0$, the perfect recovery is guaranteed with high probability for $\rankt(\LL_0)$ on the order of $nn_3/(\mu(\log n n_3)^2)$ and a number of nonzero entries in $\Sbm_0$ on the order of $n^2n_3$. For $\Sbm_0$, we make only one assumption on the random location distribution, but no assumption about the magnitudes or signs of the nonzero entries. Also TRPCA is parameter free. The mathematical analysis implies that the parameter $\lambda = 1/\sqrt{nn_3}$ leads to the correct recovery. Moreover, since the t-product of 3-way tensors reduces to the standard matrix-matrix product when $n_3=1$, the tensor nuclear norm reduces to the matrix nuclear norm. Thus, RPCA is a special case of TRPCA and the guarantee of RPCA in Theorem 1.1 in \cite{RPCA} is a special case of our Theorem \ref{thm1}. Both our model and theoretical guarantee are consistent with RPCA. Compared with SNN \cite{huang2014provable}, our tensor extension of RPCA is much more simple and elegant.

The detailed proof of Theorem \ref{thm1} can be found in the supplementary material. It is interesting to understand our proof from the perspective of the following equivalent formulation
\begin{equation}\label{trpcaeqnmixed}
	\min_{\LL,\ \E} \ \frac{1}{n_3}\left(\norm{\Lmbar}_*+\lambda\norm{\bcirc(\E)}_1\right), \ \st \ \X=\LL+\E,
\end{equation}
where (\ref{eqpropertytnn1}) is used. Program (\ref{trpcaeqnmixed}) is a mixed model since the low-rank regularization is performed on the Fourier domain while the sparse regularization is performed on the original domain. Our proof of Theorem \ref{thm1} is also conducted based on the interaction between both domains. By interpreting the tensor nuclear norm of $\LL$ as the matrix nuclear norm of $\Lmbar$ (with a factor $\frac{1}{n_3}$) in the Fourier domain, we are then able to use some existing properties of the matrix nuclear norm in the proofs. The analysis for the sparse term is kept on the original domain since the $\ell_1$-norm has no equivalent form in the Fourier domain. Though both two terms of the objective function of (\ref{trpcaeqnmixed}) are given on two matrices ($\Lmbar$ and $\bcirc(\E)$), the analysis for model (\ref{trpcaeqnmixed}) is very different from that of matrix RPCA. The matrices $\Lmbar$ and $\bcirc(\E)$ can be regarded as two matricizations of the tensor objects $\LL$ and $\E$, respectively. Their structures are more complicated than those in matrix RPCA, and thus make the proofs different from 
\cite{RPCA}. For example, our proofs require proving several bounds of norms on random tensors. Theses results and proofs, which have to use the properties of block circulant matrices and the Fourier transformation, are completely new. Some proofs are challenging due to the dependent structure of $\bcirc(\E)$ for $\E$ with an independent elements assumption. Also, TRPCA is of a different nature from the tensor completion problem \cite{zhang2015exact}. The proof of the exact recovery of TRPCA is more challenging since the $\ell_1$-norm (and its dual norm $\ell_\infty$-norm used in (\ref{tic3})) has no equivalent formulation in the Fourier domain.

It is worth mentioning that this work focuses on the analysis for 3-way tensors. But it is not difficult to generalize our model in (\ref{trpca}) and results in Theorem \ref{thm1} to the case of order-$p$ ($p\geq3$) tensors, by using the t-SVD for order-$p$ tensors in \cite{martin2013order}. 

When considering the application of TRPCA, the way for constructing a 3-way tensor from data is important. The reason is that the t-product is orientation dependent, and so is the tensor nuclear norm. Thus, the value of TNN may be different if the tensor is rotated. For example, a 3-channel color image can be formated as 3 different sizes of tensors. Therefore, when using TRPCA which is based on TNN, one has to format the data into tensors in a proper way by leveraging some priori knowledge, \textit{e.g.}, the low tubal rank property of the constructed tensor.
	
\begin{algorithm}[!t]
	\caption{Tensor Singular Value Thresholding (t-SVT)}
	\textbf{Input:} $\Y\in\Rn$, $\tau>0$.\\
	\textbf{Output:} $\mathcal{D}_\tau(\Y)$ as defined in (\ref{eqn_tsvt}).
	\begin{enumerate}[1.]
		\item Compute $\Ybar=\fft(\Y,[\ ],3)$.
		\item Perform matrix SVT on each frontal slice of $\Ybar$ by
		
		\textbf{for} $i=1,\cdots, \lceil \frac{n_3+1}{2}\rceil$ \textbf{do}
		
		\hspace*{0.4cm} $[\Um,\Sm,\Vm] = \text{SVD}(\Ymbar^{(i)})$;
		
		\hspace*{0.4cm}
		$\Wmbar^{(i)} = \Um \cdot (\Sm-\tau)_+ \cdot \Vm^*$;
		
		\textbf{end for}
		
		\textbf{for} $i=\lceil \frac{n_3+1}{2}\rceil+1,\cdots, n_3$ \textbf{do}
		
		\hspace*{0.4cm}$\Wmbar^{(i)} = \conj(\Wmbar^{(n_3-i+2)})$;
		
		\textbf{end for}
		
		\item Compute $\mathcal{D}_\tau(\Y)=\ifft(\Wbar,[\ ],3)$.
	\end{enumerate}
	\label{alg_tsvt}	
\end{algorithm} 

\subsection{Tensor Singular Value Thresholding}

Problem (\ref{trpca}) can be solved by the standard Alternating Direction Method of Multiplier (ADMM) \cite{liu2013linearized}. A key step is to compute the proximal operator of TNN
\begin{equation}\label{potnn}
	\min_{\X\in\Rn} \ \tau \norm{\X}_*+\frac{1}{2}\norm{\X - \Y}_F^2.
\end{equation}
We show that it also has a closed-form solution as the proximal operator of the matrix nuclear norm. 
Let $\Y= \U * \Sbm *\V^*$ be the tensor SVD of $\Y\in\Rn$. For each $\tau>0$, we define the tensor Singular Value Thresholding (t-SVT) operator as follows
\begin{equation}\label{eqn_tsvt}
	\mathcal{D}_{\tau}(\Y) = \U * \Sbm_\tau *\V^*,
\end{equation}
where 
\begin{equation}\label{staudef}
	\Sbm_\tau = \ifft((\Sbar -\tau)_+,[\ ],3).
\end{equation}
Note that the entries of $\Sbar$ are real. Above $t_+$ denotes the positive part of $t$, \textit{i.e.}, $t_+ = \max(t,0)$. That is, this operator simply applies a soft-thresholding rule to the singular values $\Sbar$ (not $\Sbm$) of the frontal slices of $\Ybar$, effectively shrinking these towards zero. The t-SVT operator is the proximity operator associated with TNN. 
\begin{thm}\label{thmtsvt}
	For any $\tau>0$ and $\Y \in\Rn$, the tensor singular value thresholding operator (\ref{eqn_tsvt}) obeys 
	\begin{equation}\label{thmeqnsvt}
		\mathcal{D}_{\tau}(\Y) = \arg\min_{\X\in\Rn} \ \tau \norm{\X}_*+\frac{1}{2}\norm{\X - \Y}_F^2.
	\end{equation}
\end{thm}
\begin{proof}
	The required solution to (\ref{thmeqnsvt}) is a real tensor and thus we first show that $\mathcal{D}_{\tau}(\Y)$ in (\ref{eqn_tsvt}) is real. Let $\Y= \U * \Sbm *\V^*$ be the tensor SVD of $\Y$. We know that the frontal slices of $\Sbar$ satisfy the property (\ref{keyprofffttensor}) and so do the frontal slices of $(\Sbar-\tau)_+$. By Lemma \ref{lem_keyprofft}, $\Sbm_\tau$ in (\ref{staudef}) is real. Thus, $\mathcal{D}_{\tau}(\Y)$ in (\ref{eqn_tsvt}) is real. Secondly, by using properties (\ref{eqpropertytnn1}) and (\ref{eq_proFnormNuclear}), problem (\ref{thmeqnsvt}) is equivalent to 
	\begin{align}
		\arg\min_{\X}  & \ \frac{1}{n_3}(\tau\norm{\Xmbar}_*+\frac{1}{2}\norm{\Xmbar-\Ymbar}_F^2) 	\notag \\
		=\arg\min_{\X} & \ \frac{1}{n_3}\sumi(\tau\norm{\Xmbar^{(i)}}_*+\frac{1}{2}\norm{\Xmbar^{(i)}-\Ymbar^{(i)}}_F^2). \label{eqnthmtsvteqnform}
	\end{align}
	By Theorem 2.1 in \cite{cai2010singular}, we know that the $i$-th frontal slice of $\overline{\mathcal{D}_\tau(\Y)}$ solves the $i$-th subproblem of (\ref{eqnthmtsvteqnform}). Hence, $\mathcal{D}_\tau(\Y)$ solves problem (\ref{thmeqnsvt}).
\end{proof}

Theorem \ref{thmtsvt} shows that the t-SVT operator $\mathcal{D}_\tau(\Y)$ gives a closed-form of the proximal operator of TNN, and it is a natural extension of the matrix SVT \cite{cai2010singular}. Note that $\mathcal{D}_\tau(\Y)$ is real when $\Y$ is real. By using property (\ref{keyprofffttensor}), we show how to compute $\mathcal{D}_\tau(\Y)$ efficiently in Algorithm \ref{alg_tsvt}. 

With t-SVT, we now give the details of ADMM to solve (\ref{trpca}). The augmented Lagrangian function of (\ref{trpca}) is
\begin{align*}
	L(\LL,\E,\Y,\mu) = & \norm{\LL}_* + \lambda\norm{\E}_1+ \inproduct{\Y}{\LL+\E-\X} \\
					   & + \frac{\mu}{2}\norm{\LL+\E-\X}_F^2.
\end{align*}
Then $\LL$ and $\E$ can be updated by minimizing the augmented Lagrangian function $L$ alternately. Both subproblems have closed-form solutions. See Algorithm \ref{alg1} for the whole procedure. The main per-iteration cost lies in the update of $\LL_{k+1}$, which requires computing FFT and $\lceil \frac{n_3+1}{2}\rceil$ SVDs of $n_1\times n_2$ matrices. The per-iteration complexity is $O\left(n_1n_2n_3\log n_3+\none\ntwo^2n_3\right)$.

\begin{algorithm}[t!]
	\caption{Solve (\ref{trpca}) by ADMM}
	\textbf{Input:} tensor data $\X$, parameter $\lambda$.\\
	\textbf{Initialize:} $\LL_0=\Sbm_0=\Y_0={0}$, $\rho=1.1$, $\mu_0=1\e{-3}$, $\mu_{\max}=1\e{10}$, $\epsilon=1\e{-8}$. \\ 
	\textbf{while} not converged \textbf{do}
	\begin{enumerate}[1.]
		\item Update $\LL_{k+1}$ by
		\begin{equation*}
			\LL_{k+1} = \argmin_{\LL} \ \norm{\LL}_*+\frac{\mu_k}{2}\normlarge{\LL+\E_{k}-\X+\frac{\Y_{k}}{\mu_k}}_F^2;
		\end{equation*}
		\item Update $\E_{k+1}$ by
		\begin{equation*}
			\E_{k+1} = \argmin_{\E} \ \lambda\norm{\E}_1+\frac{\mu_k}{2}\normlarge{\LL_{k+1}+\E-\X+\frac{\Y_{k}}{\mu_k}}_F^2;
		\end{equation*}
		\item $\Y_{k+1}=\Y_{k}+\mu_k(\LL_{k+1}+\E_{k+1}-\X)$;
		\item Update $\mu_{k+1}$ by $\mu_{k+1}=\min(\rho\mu_k,\mu_{\max})$;
		\item Check the convergence conditions
		\begin{align*}
			& \norm{\LL_{k+1}-\LL_{k}}_\infty\leq\epsilon, \ \norm{\E_{k+1}-\E_{k}}_\infty\leq\epsilon, \\
			& \norm{\LL_{k+1}+\E_{k+1}-\X}_\infty\leq\epsilon.
		\end{align*}
	\end{enumerate}
	\textbf{end while}
	\label{alg1}	
\end{algorithm}

\section{Experiments}\label{sec_exp}

In this section, we conduct numerical experiments to verify our main results in Theorem \ref{thm1}. We first investigate the ability of the convex TRPCA model (\ref{trpca}) to recover tensors with varying tubal rank and different levels of sparse noises. We then apply it for image recovery and background modeling. As suggested by Theorem \ref{thm1}, we set $\lambda=1/\sqrt{\none n_3}$ in all the experiments. But note that it is possible to further improve the performance by tuning $\lambda$ more carefully. The suggested value in theory provides a good guide in practice. All the simulations are conducted on a PC with an Intel Xeon E3-1270 3.60GHz CPU and 64GB memory.
 

\begin{table}[]
	\scriptsize 
	\centering
	\caption{\small Correct recovery for random problems of varying sizes.}\vspace{-0.2cm}
	$r=\rankt(\LL_0)=0.05n$, $m=\norm{\E_0}_0=0.05n^3$
	\begin{tabular}{c|c|c|c|c|c|c}
		\hline
		$n$ & $r$ & $m$ & $\rankt(\Lhat)$ & $\|\Shat\|_0$ & $\frac{\norm{\Lhat-\LL_0}_F}{\norm{\LL_0}_F}$ & $\frac{\norm{\Ehat-\E_0}_F}{\norm{\E_0}_F}$ \\ \hline\hline
		100 & 5 & $5\e{4}$ & 5 & 50,029 & $2.6\e{-7}$ & $5.4\e{-10}$\\ \hline
		200 & 10 & $4\e{5}$ & 10 & 400,234 & $5.9\e{-7}$ & $6.7\e{-10}$\\ \hline
	\end{tabular}
	\vspace{0.1cm}
	
	$r=\rankt(\LL_0)=0.05n$, $m=\norm{\E_0}_0=0.1n^3$
	\begin{tabular}{c|c|c|c|c|c|c}
		\hline
		$n$ & $r$ & $m$ & $\rankt(\Lhat)$ & $\|\Shat\|_0$ & $\frac{\norm{\Lhat-\LL_0}_F}{\norm{\LL_0}_F}$ & $\frac{\norm{\Ehat-\E_0}_F}{\norm{\E_0}_F}$    \\ \hline\hline
		100 & 5 & $1\e{5}$ & 5 & 100,117 & $4.1\e{-7}$ & $8.2\e{-10}$\\ \hline
		200 & 10 & $8\e{5}$ & 10 & 800,901 & $4.4\e{-7}$ & $4.5\e{-10}$\\ \hline
	\end{tabular}
	\vspace{0.1cm}
	
	$r=\rankt(\LL_0)=0.1n$, $m=\norm{\E_0}_0=0.1n^3$
	\begin{tabular}{c|c|c|c|c|c|c}
		\hline
		$n$ & $r$ & $m$ & $\rankt(\Lhat)$ & $\|\Shat\|_0$ & $\frac{\norm{\Lhat-\LL_0}_F}{\norm{\LL_0}_F}$ & $\frac{\norm{\Ehat-\E_0}_F}{\norm{\E_0}_F}$        \\ \hline\hline
		100  & 10 & $1\e{5}$ & 10 & 101,952 & $4.8\e{-7}$ & $1.8\e{-9}$\\ \hline
		200  & 20 & $8\e{5}$ & 20 & 815,804  & $4.9\e{-7}$ & $9.3\e{-10}$\\ \hline
	\end{tabular}
	\vspace{0.1cm}
	
	$r=\rankt(\LL_0)=0.1n$, $m=\norm{\E_0}_0=0.2n^3$
	\begin{tabular}{c|c|c|c|c|c|c}
		\hline
		$n$ & $r$ & $m$  & $\rankt(\Lhat)$ & $\|\Ehat\|_0$  & $\frac{\norm{\Lhat-\LL_0}_F}{\norm{\LL_0}_F}$ & $\frac{\norm{\Ehat-\E_0}_F}{\norm{\E_0}_F}$  \\ \hline\hline
		100  & 10 & $2\e{5}$ & 10 & 200,056 & $7.7\e{-7}$ & $4.1\e{-9}$ \\ \hline
		200  & 20 & $16\e{5}$ & 20 & 1,601,008 & $1.2\e{-6}$ & $3.1\e{-9}$ \\ \hline
	\end{tabular}\label{tab_recov}
	\vspace{-0.3cm}
\end{table}

\subsection{Exact Recovery from Varying Fractions of Error}

We first verify the correct recovery guarantee of Theorem \ref{thm1} on randomly generated problems. We simply consider the tensors of size $n\times n \times n$, with varying dimension $n=$100 and 200. We generate a tensor with tubal rank $r$ as a product $\LL_0=\PP*\Q^*$, where $\PP$ and $\Q$ are ${n\times r\times n}$ tensors with entries independently sampled from $\mathcal{N}(0,1/n)$ distribution. The support set $\Omegat$ (with size $m$) of $\E_0$ is chosen uniformly at random. For all $(i,j,k)\in\Omegat$, let $[\E_0]_{ijk}=\M_{ijk}$, where $\M$ is a tensor with independent Bernoulli $\pm 1$ entries.

Table \ref{tab_recov} reports the recovery results based on varying choices of the tubal rank $r$ of $\LL_0$ and the sparsity $m$ of $\E_0$. It can be seen that our convex program (\ref{trpca}) gives the correct tubal rank estimation of $\LL_0$ in all cases and also the relative errors ${\|{\hat{\LL}-\LL_0}\|_F}/{\norm{\LL_0}_F}$ are very small, less than $10^{-5} $. The sparsity estimation of $\E_0$ is not as exact as the rank estimation, but note that the relative errors ${\|{\Ehat-\E_0}\|_F}/{\norm{\E_0}_F}$ are all very small, less than $10^{-8}$ (actually much smaller than the relative errors of the recovered low-rank component). These results well verify the correct recovery phenomenon as claimed in Theorem \ref{thm1}. Figure \ref{fig_TRPCA_toy} also intuitively illustrates the recovered low rank and sparse tensors obtain by TRPCA. 

\subsection{Phase Transition in Tubal Rank and Sparsity}

The results in Theorem \ref{thm1} show the perfect recovery for incoherent tensor with $\rankt(\LL_0)$ on the order of $nn_3/(\mu(\log n n_3)^2)$ and the sparsity of $\E_0$ on the order of $n^2n_3$. Now we examine the recovery phenomenon with varying tubal rank of $\LL_0$ from varying sparsity of $\E_0$. We consider two sizes of $\LL_0\in\mathbb{R}^{n\times n\times n_3}$: (1) $n=100$, $n_3=50$; (2) $n=200$, $n_3=50$. We generate $\LL_0=\PP*\Q^*$, where $\PP$ and $\Q$ are ${n\times r\times n_3}$ tensors with entries independently sampled from a $\mathcal{N}(0,1/n)$ distribution. For $\E_0$, we consider a Bernoulli model for its support and random signs for its values:
\begin{equation*}
	{[\E_0]}_{ijk} =
	\begin{cases}
		1,  & \text{w.p.} \ \rho_s/2,\\
		0,  & \text{w.p.} \ 1-\rho_s,\\
		-1, & \text{w.p.} \ \rho_s/2.\\
	\end{cases}
\end{equation*}
We set $r/n=[0.01:0.01:0.5]$ and $\rho_s=[0.01:0.01:0.5]$. For each $(r,\rho_s)$-pair, we simulate 10 test instances and declare a trial to be successful if the recovered $\hat{\LL}$ satisfies ${\|{\hat{\LL}-\LL_0}\|_F}/{\norm{\LL_0}_F}\leq 10^{-3}$. Figure \ref{fig_sparsityvsrank} plots the fraction of correct recovery for each pair $(r,\rho_s)$ (black = $0\%$ and white = 100$\%$). It can be seen that there is a large region in which the recovery is correct. The recovery phenomenon is similar in both cases with different sizes of $\LL_0$. Intuitively, the experiment shows that the recovery is correct when the tubal rank of $\LL_0$ is relatively low and the errors $\E_0$ is relatively sparse. Such an observation is consistent with Theorem \ref{thm1}. Similar observations can be found in the matrix RPCA (see Figure 1 (a) in \cite{RPCA}).

\begin{figure}[!t]
	\centering
	\includegraphics[width=0.4\textwidth]{fig/fig_TRPCA_toy.pdf}
	\caption{\small Illustration of TRPCA on synthetic data.}
	\label{fig_TRPCA_toy}
	\vspace{-0.3cm}
\end{figure}
\begin{figure}[!t]
	\centering
	\begin{subfigure}[b]{0.22\textwidth}
		\centering
		\includegraphics[width=\textwidth]{fig/fig_sparsityvsrank1.pdf}
		\caption{$n_1=n_2=100$, $n_3=50$}
	\end{subfigure}
	\begin{subfigure}[b]{0.22\textwidth}
		\centering
		\includegraphics[width=\textwidth]{fig/fig_sparsityvsrank2.pdf}
		\caption{$n_1=n_2=200$, $n_3=50$}
	\end{subfigure}
	\caption{\small{Correct recovery for varying tubal ranks of $\LL_0$ and sparsities of $\E_0$. Fraction of correct recoveries across 10 trials, as a function of $\rankt(\LL_0)$ (x-axis) and sparsity of $\E_0$ (y-axis). The experiments are performed on two different sizes of $\LL_0\in\Rn$.}}
	\label{fig_sparsityvsrank}
	\vspace{-0.4cm}
\end{figure}

\begin{figure*}[!t]
	\centering
	\begin{subfigure}[b]{1\textwidth}
		\centering
		\includegraphics[width=\textwidth]{fig/fig_img_res_psnr.pdf}
	\end{subfigure}
	\begin{subfigure}[b]{1\textwidth}
		\centering
		\includegraphics[width=\textwidth]{fig/fig_img_res_time.pdf}
	\end{subfigure}
	\caption{\small Comparison of the PSNR values (top) and running time (bottom) obtained by RPCA, SNN and TRPCA on 100 images.}
	\label{fig_img_res_psnr_time}
	\vspace{-0.4cm}
\end{figure*}

\begin{figure*}[!t]
	\centering	
	\begin{subfigure}[b]{0.195\textwidth}
		\centering
		\includegraphics[width=\textwidth]{fig/fig_img1_original.pdf}
	\end{subfigure}
	\begin{subfigure}[b]{0.195\textwidth}
		\centering
		\includegraphics[width=\textwidth]{fig/fig_img1_noisy.pdf}
	\end{subfigure}
	\begin{subfigure}[b]{0.195\textwidth}
		\centering
		\includegraphics[width=\textwidth]{fig/fig_img1_RPCA.pdf}
	\end{subfigure}
	\begin{subfigure}[b]{0.195\textwidth}
		\centering
		\includegraphics[width=\textwidth]{fig/fig_img1_SNN.pdf}
	\end{subfigure}
	\begin{subfigure}[b]{0.195\textwidth}
		\centering
		\includegraphics[width=\textwidth]{fig/fig_img1_TRPCA.pdf}
	\end{subfigure}	
	\begin{subfigure}[b]{0.195\textwidth}
		\centering
		\includegraphics[width=\textwidth]{fig/fig_img2_original.pdf}
	\end{subfigure}
	\begin{subfigure}[b]{0.195\textwidth}
		\centering
		\includegraphics[width=\textwidth]{fig/fig_img2_noisy.pdf}
	\end{subfigure}
	\begin{subfigure}[b]{0.195\textwidth}
		\centering
		\includegraphics[width=\textwidth]{fig/fig_img2_RPCA.pdf}
	\end{subfigure}
	\begin{subfigure}[b]{0.195\textwidth}
		\centering
		\includegraphics[width=\textwidth]{fig/fig_img2_SNN.pdf}
	\end{subfigure}
	\begin{subfigure}[b]{0.195\textwidth}
		\centering
		\includegraphics[width=\textwidth]{fig/fig_img2_TRPCA.pdf}
	\end{subfigure}
	\begin{subfigure}[b]{0.195\textwidth}
		\centering
		\includegraphics[width=\textwidth]{fig/fig_img3_original.pdf}
	\end{subfigure}
	\begin{subfigure}[b]{0.195\textwidth}
		\centering
		\includegraphics[width=\textwidth]{fig/fig_img3_noisy.pdf}
	\end{subfigure}
	\begin{subfigure}[b]{0.195\textwidth}
		\centering
		\includegraphics[width=\textwidth]{fig/fig_img3_RPCA.pdf}
	\end{subfigure}
	\begin{subfigure}[b]{0.195\textwidth}
		\centering
		\includegraphics[width=\textwidth]{fig/fig_img3_SNN.pdf}
	\end{subfigure}
	\begin{subfigure}[b]{0.195\textwidth}
		\centering
		\includegraphics[width=\textwidth]{fig/fig_img3_TRPCA.pdf}
	\end{subfigure}
	\begin{subfigure}[b]{0.195\textwidth}
		\centering
		\includegraphics[width=\textwidth]{fig/fig_img4_original.pdf}
	\end{subfigure}
	\begin{subfigure}[b]{0.195\textwidth}
		\centering
		\includegraphics[width=\textwidth]{fig/fig_img4_noisy.pdf}
	\end{subfigure}
	\begin{subfigure}[b]{0.195\textwidth}
		\centering
		\includegraphics[width=\textwidth]{fig/fig_img4_RPCA.pdf}
	\end{subfigure}
	\begin{subfigure}[b]{0.195\textwidth}
		\centering
		\includegraphics[width=\textwidth]{fig/fig_img4_SNN.pdf}
	\end{subfigure}
	\begin{subfigure}[b]{0.195\textwidth}
		\centering
		\includegraphics[width=\textwidth]{fig/fig_img4_TRPCA.pdf}
	\end{subfigure}
	\begin{subfigure}[b]{0.195\textwidth}
		\centering
		\includegraphics[width=\textwidth]{fig/fig_img5_original.pdf}
	\end{subfigure}
	\begin{subfigure}[b]{0.195\textwidth}
		\centering
		\includegraphics[width=\textwidth]{fig/fig_img5_noisy.pdf}
	\end{subfigure}
	\begin{subfigure}[b]{0.195\textwidth}
		\centering
		\includegraphics[width=\textwidth]{fig/fig_img5_RPCA.pdf}
	\end{subfigure}
	\begin{subfigure}[b]{0.195\textwidth}
		\centering
		\includegraphics[width=\textwidth]{fig/fig_img5_SNN.pdf}
	\end{subfigure}
	\begin{subfigure}[b]{0.195\textwidth}
		\centering
		\includegraphics[width=\textwidth]{fig/fig_img5_TRPCA.pdf}
	\end{subfigure}
	\begin{subfigure}[b]{0.195\textwidth}
		\centering
		\includegraphics[width=\textwidth]{fig/fig_img6_original.pdf}
		\caption{\small Orignal image}
	\end{subfigure}
	\begin{subfigure}[b]{0.195\textwidth}
		\centering
		\includegraphics[width=\textwidth]{fig/fig_img6_noisy.pdf}
		\caption{\small Observed image}
	\end{subfigure}
	\begin{subfigure}[b]{0.195\textwidth}
		\centering
		\includegraphics[width=\textwidth]{fig/fig_img6_RPCA.pdf}
		\caption{\small RPCA}
	\end{subfigure}
	\begin{subfigure}[b]{0.195\textwidth}
		\centering
		\includegraphics[width=\textwidth]{fig/fig_img6_SNN.pdf}
		\caption{\small SNN}
	\end{subfigure}
	\begin{subfigure}[b]{0.195\textwidth}
		\centering
		\includegraphics[width=\textwidth]{fig/fig_img6_TRPCA.pdf}
		\caption{\small TRPCA}
	\end{subfigure}
	\\
	\vspace{0.8em}\small 
	\centering
\begin{minipage}[c]{0.48\textwidth}%
	\centering
	\begin{tabular}{c|c|c|c|c|c|c}
		\hline
		Index	& 1		& 2	    & 3	    & 4	    & 5	    & 6    	\\ \hline
		RPCA	& 29.10 & 24.53 & 25.12 & 24.31 & 27.50 & 26.77 \\ \hline
		SNN 	& 30.91 & 26.45 & 27.66 & 26.45 & 29.26 & 28.19 \\ \hline
		TRPCA	& \textbf{32.33} & \textbf{28.30} & \textbf{28.59} & \textbf{28.62} & \textbf{31.06} & \textbf{30.16} \\ \hline
	\end{tabular}
	\caption*{\small (f) Comparison of the PSNR values on the above 6 images.}
\end{minipage}	
\begin{minipage}[c]{0.48\textwidth}%
	\centering
	\begin{tabular}{c|c|c|c|c|c|c}
		\hline
		Index 	& 1	    & 2    & 3    & 4    & 5    & 6    \\ \hline
		RPCA 	& 14.98 &13.79 &14.35 &12.45 &12.72 &15.73 \\ \hline
		SNN 	& 26.93 &25.20 &25.33 &23.47 &23.38 &28.16 \\ \hline
		TRPCA 	& \textbf{12.96} &\textbf{12.24} &\textbf{12.76} &\textbf{10.70} &\textbf{10.64} &\textbf{14.31} \\ \hline
	\end{tabular}
	\caption*{\small (g) Comparison of the running time (s) on the above 6 images.}
\end{minipage}
	\caption{\small{Recovery performance comparison on 6 example images. (a) Original image; (b) observed image; (c)-(e) recovered images by RPCA, SNN and TRPCA, respectively; (f) and (g) show the comparison of PSNR values and running time (second) on the above 6 images. }}\label{fig_imageinpr}
	\vspace{-0.3cm}
\end{figure*}

\subsection{Application to Image Recovery}

We apply TRPCA to image recovery from the corrupted images with random noises.The motivation is that the color images can be approximated by low rank matrices or tensors \cite{liu2013tensor,gandy2011tensor}. We will show that the recovery performance of TRPCA is still satisfactory with the suggested parameter in theory on real data. 

We use 100 color images from the Berkeley Segmentation Dataset \cite{martin2001database} for the test. The sizes of images are $321\times 481$ or $481\times 321$. For each image, we randomly set $10\%$ of pixels to random values in [0, 255], and the positions of the corrupted pixels are unknown. All the 3 channels of the images are corrupted at the same positions (the corruptions are on the whole tubes). This problem is more challenging than the corruptions on 3 channels at different positions. See Figure \ref{fig_imageinpr} (b) for some sample images with noises. We compare our TRPCA model with RPCA \cite{RPCA} and SNN \cite{huang2014provable} which also own the theoretical recovery guarantee. For RPCA, we apply it on each channel separably and combine the results to obtain the recovered image. The parameter $\lambda$ is set to $\lambda = 1/\max{(n_1,n_2)}$ as suggested in theory. For SNN in (\ref{eqsnn}), we find that it does not perform well when $\lambda_i$'s are set to the values suggested in theory \cite{huang2014provable}. We empirically set $\bm{\lambda}=[15, 15, 1.5]$ in (\ref{eqsnn}) to make SNN perform well in most cases. For our TRPCA, we format a $n_1\times n_2$ sized image as a tensor of size $n_1\times n_2\times 3$. We find that such a way of tensor construction usually performs better than some other ways. This may be due to the noises which present on the tubes. We set $\lambda = 1/\sqrt{3\max{(n_1,n_2)}}$ in TRPCA. We use the Peak Signal-to-Noise Ratio (PSNR), defined as
\begin{equation*}\label{psnreq}
	\text{PSNR} = 10\log_{10}\left( \frac{\norm{\M}_\infty^2}{\frac{1}{n_1n_2n_3} {\|\hat{\mathbf{\X}}-\M\|_F^2}} \right), 
\end{equation*}
to evaluate the recovery performance. 

Figure \ref{fig_img_res_psnr_time} gives the comparison of the PSNR values and running time on all 100 images. Some examples with the recovered images are shown in Figure \ref{fig_imageinpr}. From these results, we have the following observations. First, both SNN and TRPCA perform much better than the matrix based RPCA. The reason is that RPCA performs on each channel independently, and thus is not able to use the information across channels. The tensor methods instead take advantage of the multi-dimensional structure of data. Second, TRPCA outperforms SNN in most cases. This not only demonstrates the superiority of our TRPCA, but also validates our recovery guarantee in Theorem \ref{thm1} on image data. Note that SNN needs some additional effort to tune the weighted parameters $\lambda_i$'s empirically. Different from SNN which is a loose convex surrogate of the sum of Tucker rank, our TNN is a tight convex relaxation of the tensor average rank, and the recovery performance of the obtained optimal solutions has the tight recovery guarantee as RPCA. Third, we use the standard ADMM to solve RPCA, SNN and TRPCA. Figure \ref{fig_img_res_psnr_time} (bottom) shows that TRPCA is as efficient as RPCA, while SNN requires the highest cost in this experiment.

\subsection{Application to Background Modeling}

In this section, we consider the background modeling problem which aims to separate the foreground objects from the background. The frames of the background are highly corrected and thus can be modeled as a low rank tensor. The moving foreground objects occupy only a fraction of image pixels and thus can be treated as sparse errors. We solve this problem by using RPCA, SNN and TRPCA. We consider two example videos, \textit{Hall} and \textit{WaterSurface}\footnote{http://perception.i2r.a-star.edu.sg/bk model/bk index.html}, which have color frames. The sequence \textit{Hall} has 300 frames with frame size $144\times 176$. To use RPCA, we reshape it to a $76032\times 300$ matrix. To use SNN and TRPCA, we reshape it to a $25344\times 3\times 300$ tensor\footnote{We observe that this way of tensor construction performs well, despite one has some other ways.}. The sequence \textit{WaterSurface} has 300 frames with frame size $128\times 160$. To use RPCA, we reshape it to a $61440\times 300$ matrix. To use SNN and TRPCA, we reshape it to a $20480\times 3\times 300$ tensor. The parameter of SNN in \ref{eqsnn} is set to $\bm{\lambda}=[10,\ 0.1,\ 1]\times 20$ in this experiment. 

Figure \ref{fig_bg_res} shows the performance and running time comparison of RPCA, SNN and TRPCA on \textit{Hall} and \textit{WaterSurface} sequences. It can be seen that the low rank components identify the main illuminations as background, while the sparse parts correspond to the motion in the scene. Generally, our TRPCA again outperforms the competing methods. Also, TRPCA is as efficient as RPCA and SNN requires much higher computational cost. The efficiency of TRPCA is benefited from our faster way for computing tensor SVT in Algorithm \ref{alg_tsvt} which is the key step for solving TRPCA. 

\captionsetup{position=top}
\begin{figure}[!t]
	\centering
	\begin{subfigure}[b]{0.118\textwidth}
		\centering
		\includegraphics[width=\textwidth]{fig/fig_bg_hall_X.pdf}
	\end{subfigure}
	\begin{subfigure}[b]{0.118\textwidth}
		\centering
		\includegraphics[width=\textwidth]{fig/fig_bg_hall_L_RPCA.pdf}
	\end{subfigure}
	\begin{subfigure}[b]{0.118 \textwidth}
		\centering
		\includegraphics[width=\textwidth]{fig/fig_bg_hall_L_SNN.pdf}
	\end{subfigure}
	\begin{subfigure}[b]{0.118 \textwidth}
		\centering
		\includegraphics[width=\textwidth]{fig/fig_bg_hall_L_TRPCA.pdf}
	\end{subfigure}

	\begin{subfigure}[b]{0.118\textwidth}
	\end{subfigure}
	\hfill
	\begin{subfigure}[b]{0.118\textwidth}
		\centering
		\includegraphics[width=\textwidth]{fig/fig_bg_hall_S_RPCA.pdf}
	\end{subfigure}
	\begin{subfigure}[b]{0.118 \textwidth}
		\centering
		\includegraphics[width=\textwidth]{fig/fig_bg_hall_S_SNN.pdf}
	\end{subfigure}
	\begin{subfigure}[b]{0.118 \textwidth}
		\centering
		\includegraphics[width=\textwidth]{fig/fig_bg_hall_S_TRPCA.pdf}
	\end{subfigure}

	\begin{subfigure}[b]{0.118\textwidth}
		\centering
		\includegraphics[width=\textwidth]{fig/fig_bg_WaterSurface_X.pdf}
	\end{subfigure}
	\begin{subfigure}[b]{0.118\textwidth}
		\centering
		\includegraphics[width=\textwidth]{fig/fig_bg_WaterSurface_L_RPCA.pdf}
	\end{subfigure}
	\begin{subfigure}[b]{0.118 \textwidth}
		\centering
		\includegraphics[width=\textwidth]{fig/fig_bg_WaterSurface_L_SNN.pdf}
	\end{subfigure}
	\begin{subfigure}[b]{0.118 \textwidth}
		\centering
		\includegraphics[width=\textwidth]{fig/fig_bg_WaterSurface_L_TRPCA.pdf}
	\end{subfigure}
	
	\begin{subfigure}[b]{0.118\textwidth}
		
		\caption{Original}
	\end{subfigure}
	\hfill
	\begin{subfigure}[b]{0.118\textwidth}
		\centering
		\includegraphics[width=\textwidth]{fig/fig_bg_WaterSurface_S_RPCA.pdf}
		\caption{RPCA}
	\end{subfigure}
	\begin{subfigure}[b]{0.118 \textwidth}
		\centering
		\includegraphics[width=\textwidth]{fig/fig_bg_WaterSurface_S_SNN.pdf}
		\caption{SNN}
	\end{subfigure}
	\begin{subfigure}[b]{0.118 \textwidth}
		\centering
		\includegraphics[width=\textwidth]{fig/fig_bg_WaterSurface_S_TRPCA.pdf}
		\caption{TRPCA}
	\end{subfigure}
	\vspace{0.8em}\small 
	\centering
	\begin{minipage}[c]{0.48\textwidth}
			\centering
			\begin{tabular}{c|c|c|c}
				\hline
				 & RPCA & SNN & TRPCA \\ \hline
				\textit{Hall}   & \textbf{301.8} & 1553.2 & 323.0 \\ \hline
				\textit{WaterSurface} & 250.1 & 887.3 & \textbf{224.2} \\ \hline
			\end{tabular}
		\caption*{\small (e) Running time (seconds) comparison.}
	\end{minipage}
	\vspace{-0.3cm}
	\caption{\small{Background modeling from videos. Two example frames from \textit{Hall} and \textit{WaterSurface} sequences are shown. (a) Original frames; (b)-(d) low rank and sparse components obtained by RPCA, SNN and TRPCA, respectively; (e) running time comparison.}}
	\label{fig_bg_res}
	\vspace{-0.6cm}
\end{figure}

\section{Conclusions and Future Work} \label{sec_con}

Based on the recently developed tensor-tensor product, which is a natural extension of the matrix-matrix product, we rigorously defined the tensor spectral norm, tensor nuclear norm and tensor average rank, such that their properties and relationships are consistent with the matrix cases. We then studied the Tensor Robust Principal Component (TRPCA) problem which aims to recover a low tubal rank tensor and a sparse tensor from their sum. We proved that under certain suitable assumptions, we can recover both the low-rank and the sparse components exactly by simply solving a convex program whose objective is a weighted combination of the tensor nuclear norm and the $\ell_1$-norm. Benefitting from the ``good" property of tensor nuclear norm, both our model and theoretical guarantee are natural extensions of RPCA. We also developed a more efficient method to compute the tensor singular value thresholding problem which is the key for solving TRPCA.  Numerical experiments verify our theory and the results on images and videos demonstrate the effectiveness of our model. 

This work verifies the ability of low-rank tensors recovery by tensor nuclear norm minimization. This also suggests the potential to use these tools of tensor analysis for other applications, \textit{e.g.}, image/video processing, web data analysis, and bioinformatics. Also, the computational cost is high when the data is of high dimension. Thus, developing more efficient solvers, \textit{e.g.}, \cite{mu2016scalable,yuan2016exact}, is important. It is also interesting to study the tensor extensions of other convex/nonconvex low-rank models \cite{zhang2015exactrecovera,balcan2016noise,liu2017new,sun2015nonconvex}.

{
	\bibliographystyle{ieee}
	\bibliography{ref}
}

%
%

\appendices

At the following, we give the detailed proofs of Theorem 3.1, Theorem 3.2, and the main result in Theorem 4.1. Section \ref{supp_sec_notations} first gives some notations and properties which will be used in the proofs. Section \ref{supp_profthm3132} gives the proofs of Theorem 3.1 and 3.2 in our paper. Section \ref{supp_sec_dual} provides a way for the construction of the solution to the TRPCA model, and Section \ref{supp_sec_proofofdual} proves that the constructed solution is optimal to the TRPCA problem. Section \ref{supp_sec_prooflemmas} gives the proofs of some lemmas which are used in Section \ref{supp_sec_proofofdual}.

\section{Preliminaries}
\label{supp_sec_notations}


Beyond the notations introduced in the paper, we need some other notations used in the proofs.
At the following, we define $\eijk=\ei*\ek*\ej^*$. Then we have $\X_{ijk} = \inproduct{\X}{\eijk}$. We define the projection $$\Pomega(\Z)=\sum_{ijk}\delta_{ijk}z_{ijk}\eijk,$$
where $\delta_{ijk}=1_{(i,j,k)\in\Omegat}$, where $1_{(\cdot)}$ is the indicator function. Also $\Omegat^c$ denotes the 
complement of $\Omegat$ and $\Pomegao$ is the projection onto $\Omegat^c$. Denote $\Tm$ by the set 
\begin{align}
\Tm = \{ \U*\Y^* + \W*\V^*, \ \Y, \W\in\mathbb{R}^{n\times r\times n_3} \},
\end{align}
and by $\Tm^\bot$  its orthogonal complement. Then the projections onto $\Tm$ and $\Tm^\bot$ are respectively
\begin{align*}
\PT(\Z) = \U*\U^**\Z + \Z *\V*\V^* -  \U*\U^**\Z *\V*\V^*, 
\end{align*}
\begin{align*}
\PTo(\Z) =& \Z-\PT(\Z)\\
=&(\I_{n_1} - \U*\U^*)*\Z*(\I_{n_2}-\V*\V^*),
\end{align*}
where $\I_n$ denotes the $n\times n\times n_3$ identity tensor. Note that $\PT$ is self-adjoint. So we have 
\begin{align*}
&\norm{\PT(\eijk)}_F^2 \\
= & \inproduct{\PT(\eijk)}{\eijk} \\
= & \inproduct{\U*\U^**\eijk + \eijk *\V*\V^*}{\eijk} \\
& -\inproduct{ \U*\U^**\eijk *\V*\V^*}{\eijk}
\end{align*}
Note that 
\begin{align*}
&\inproduct{\U*\U^**\eijk}{\eijk}\\
=&\inproduct{\U*\U^**\ei*\ek*\ej^*}{\ei*\ek*\ej^*}\\
=&\inproduct{\U^**\ei}{\U^**\ei*(\ek*\ej^**\ej*\ek^*)}\\
=&\inproduct{\U^**\ei}{\U^**\ei}\\
=&\norm{\U^**\ei}_F^2,
\end{align*}
where we use the fact that $\ek*\ej^**\ej*\ek^*=\I_1$, which is the $1\times 1\times n_3$ identity tensor. Therefore, it is easy to see that 
\begin{align}
&\norm{\PT(\eijk)}_F^2 \notag \\
= & \norm{\U^**\ei}_F^2 + \norm{\V^**\ej}_F^2 - \norm{\U^**\ei*\ek*\ej^**\V}_F^2, \notag\\
\leq & \norm{\U^**\ei}_F^2 + \norm{\V^**\ej}_F^2 \notag \\
\leq & \frac{\mu r(n_1+n_2)}{n_1n_2n_3}  \label{supp_proabouPTn1n2}\\
= & \frac{2\mu r}{nn_3}, \ \text{when } n_1=n_2=n.\label{supp_proabouPT}
\end{align}
where (\ref{supp_proabouPTn1n2}) uses the following tensor incoherence conditions
\begin{align}
\max_{i=1,\cdots,n_1} \norm{\U^**\mathring{\mathfrak{e}}_i}_F\leq\sqrt{\frac{\mu r}{n_1n_3}}, \label{supp_tic1}\\
\max_{j=1,\cdots,n_2} \norm{\V^**\mathring{\mathfrak{e}}_j}_F\leq\sqrt{\frac{\mu r}{n_2n_3}},\label{supp_tic2}
\end{align}
and
\begin{equation} \label{supp_tic3}
\norm{\U*\V^*}_\infty\leq \sqrt{\frac{\mu r}{n_1n_2n_3^2}},
\end{equation}	
which are assumed to be satisfied in Theorem 4.1 in our manuscript.


\section{Proofs of Theorem 3.1 and Theorem 3.2}\label{supp_profthm3132}

\subsection{Proof of Theorem 3.1}

\begin{proof}
	To complete the proof, we need the conjugate function concept. The conjugate $\phi^*$ of a function $\phi: C\rightarrow \mathbb{R}$, where $C \subset \mathbb{R}^n$, is defined as
	\begin{equation*}
	\phi^*(\y) = \sup \{\inproduct{\y}{\x} - \phi(\x) | \x \in C \}.
	\end{equation*}
	Note that the conjugate of the conjugate, $\phi^{**}$, is the convex envelope of the function $\phi$. See Theorem 1.3.5 in \cite{hiriart1991convex,fazel2002matrix}.  The proofs has two steps which compute $\phi^*$ and $\phi^{**}$, respectively.
	
	\textbf{Step 1.} \textit{Computing $\phi^*$.} For any $\A\in\Rn$, the conjugate function of the tensor average rank
	\begin{equation*}
	\phi(\A) = \rankt(\A)=\frac{1}{n_3}\rank(\bcirc(\A))=\frac{1}{n_3}\rank(\Ambar),
	\end{equation*}
	on the set $S = \{\A\in\Rn |\norm{\A}\leq 1\}$ is
	\begin{align*}
	\phi^*(\B) = & \sup_{\norm{\A}\leq 1} \left(\inproduct{\B}{\A} - \rankt(\A)\right) \\
	= & \sup_{\norm{\A}\leq 1} \frac{1}{n_3}(\inproduct{\Bmbar}{\Ambar} - \rank(\Ambar)).
	\end{align*}
	Here $\Ambar, \Bmbar\in\mathbb{C}^{n_1n_3\times n_2n_3}$. Let $q = \min\{n_1n_3, n_2n_3\}$. By von Neumann's trace theorem, 
	\begin{align}\label{supp_vonneutrace}
	\inproduct{\Bmbar}{\Ambar} \leq \sum_{i = 1}^{q} \sigma_i(\Bmbar) \sigma_i(\Ambar),
	\end{align} 
	where $\sigma_i(\Ambar)$ denotes the $i$-th largest singular value of $\Ambar$. Let $\Ambar = \Umbar_1\Smbar_1\Vm^*_1$ and $\Bmbar = \Umbar_2\Smbar_2\Vmbar^*_2$ be the SVD of $\Ambar$ and $\Bmbar$, respectively. Note that the equality (\ref{supp_vonneutrace}) holds when
	\begin{align}\label{supp_conduvab}
	\Umbar_1 = \Umbar_2  \ \text{and} \ \Vmbar_1 = \Vmbar_2.
	\end{align}
	So we can pick $\Umbar_1$ and $\Vmbar_1$ such that (\ref{supp_conduvab}) holds to maximize $\inproduct{\Bmbar}{\Ambar}$. Note that the corresponding $\U$ and $\V$ of $\Umbar_1$ and $\Vmbar_1$  respectively are real tensors and so is $\A$ in this case. Thus, we have
	\begin{align*}
	\phi^*(\B) = & \sup_{\norm{\A}\leq 1} \frac{1}{n_3}\left(\sum_{i = 1}^{q} \sigma_i(\Bmbar) \sigma_i(\Ambar) - \rank(\Ambar)\right).
	\end{align*}
	
	If $\A = 0$, then $\Ambar = 0$, and thus we have $\phi^*(\B) = 0$ for all $\B$. If $\rank(\Ambar) = r$, $1\leq r\leq q$, then $\phi^*(\B) = \frac{1}{n_3}\left(\sum_{i = 1}^{r} \sigma_i(\Bmbar) - r\right)$. Hence $\phi^*(\B)$ can be expressed as
	\begin{align*}
	& n_3 \cdot \phi^*(\B) \\
	= & {\max \left\{0, \sigma_1(\Bmbar) -1, \cdots, \sum_{i = 1}^{r}\sigma_i(\Bmbar) -r, \cdots,  \sum_{i = 1}^{q} \sigma_i(\Bmbar) - q\right\}}.
	\end{align*}
	The largest term in this set is the one that sums all positive $(\sigma_i(\Bmbar) -1)$ terms. Thus, we have
	\begin{align*}
	&\phi^*(\B) \\
	= & \begin{cases}
	0, \qquad\qquad \qquad \quad \quad \quad \quad       \norm{\Bmbar}\leq 1, \\
	\frac{1}{n_3}\left(\sum_{i = 1}^{r}\sigma_i(\Bmbar) -r\right), \quad \sigma_r(\Bmbar) > 1 \text{ and } \sigma_{r+1}(\Bmbar) \leq 1 
	\end{cases} \\
	= &  \frac{1}{n_3} \sum_{i = 1}^{q}(\sigma_i(\Bmbar) -1)_+. 	\end{align*}
	Note that above $\norm{\Bmbar}\leq 1$ is equivalent to $\norm{\B}\leq 1$.
	
	\textbf{Step 2.} \textit{Computing $\phi^{**}$.} Now we compute the conjugate of $\phi^*$, defined as
	\begin{align*}
	\phi^{**}(\C) = & \sup_{\B} ( \inproduct{\C}{\B} - \phi^*(\B) ) \\
	= & \sup_{\B} \left(\frac{1}{n_3} \inproduct{\Cmbar}{\Bmbar}-\phi^*(\B)\right),
	\end{align*}
	for all $\C \in S$. As before, we can choose $\B$ such that 
	\begin{align*}
	\phi^{**}(\C) = & \sup_{\B} \left(\frac{1}{n_3}\sum_{i = 1}^{q} \sigma_i(\Cmbar) \sigma_i(\Bmbar) - \phi^*(\B)\right).
	\end{align*}
	At the following, we consider two cases, $\norm{\C} > 1$ and $\norm{\C} \leq 1$.
	
	If $\norm{\C} > 1$, then $\sigma_1(\Cmbar) = \norm{\Cmbar} = \norm{\C} > 1$.  We can choose $\sigma_1(\Bmbar)$ large enough so that $\phi^{**}(\C)\rightarrow\infty$. To see this, note that  in
	\begin{align*}
	\phi^{**}(\C) = & \sup_{\B}\frac{1}{n_3} \left(\sum_{i = 1}^{q} \sigma_i(\Cmbar) \sigma_i(\Bmbar) - \left( \sum_{i=1}^{r} \sigma_i(\Bmbar)  - r \right)\right),
	\end{align*}
	the coefficient of $\sigma_1(\Bmbar)$ is $\frac{1}{n_3}(\sigma_1(\Cmbar)-1)$ which is positive.
	
	If $\norm{\C} \leq 1$, then $\sigma_1(\Cmbar) = \norm{\Cmbar} = \norm{\C} \leq 1$. If $\norm{\B} = \norm{\Bmbar} \leq 1$, then $\phi^*(\B)=0$ and the supremum is achieved for $\sigma_i(\Bmbar)=1$, $i=1,\cdots,q$, yielding 
	\begin{align*}
	\phi^{**}(\C) = \frac{1}{n_3} \sum_{i=1}^{q} \sigma_i(\Cmbar ) = \frac{1}{n_3} \norm{\Cmbar}_* = \norm{\C}_*.
	\end{align*}
	If $\norm{\C}>1$, we show that the argument of sup is is always smaller than $\norm{\C}_*$. By adding and subtracting the term $\frac{1}{n_3}\sum_i^q\sigma_i(\Cmbar)$ and rearranging the terms, we have
	\begin{align*}
	&\frac{1}{n_3} \left(\sum_{i = 1}^{q} \sigma_i(\Cmbar) \sigma_i(\Bmbar) -  \sum_{i=1}^{r}  \left(\sigma_i(\Bmbar)  - 1 \right)\right) \\
	=& \frac{1}{n_3} \left(\sum_{i = 1}^{q} \sigma_i(\Cmbar) \sigma_i(\Bmbar) -  \sum_{i=1}^{r}  \left(\sigma_i(\Bmbar)  - 1 \right)\right) \\
	& - \frac{1}{n_3}\sum_{i=1}^q\sigma_i(\Cmbar) + \frac{1}{n_3}\sum_{i=1}^q\sigma_i(\Cmbar) \\
	=&\frac{1}{n_3} \sum_{i=1}^r (\sigma_i(\Bmbar)-1)(\sigma_i(\Cmbar)-1)\\
	&+\frac{1}{n_3}\sum_{i=r+1}^q (\sigma_i(\Bmbar)-1)\sigma_i(\Cmbar) +\frac{1}{n_3} \sum_{i=1}^{q} \sigma_i(\Cmbar) \\
	< & \frac{1}{n_3}\sum_{i=1}^{q} \sigma_i(\Cmbar)\\
	= & \norm{\C}_*.
	\end{align*}
	
	In a summary, we have shown that
	\begin{align*}
	\phi^{**}(\C) = \norm{\C}_*, 
	\end{align*}
	over the set $S=\{\C | \norm{\C} \leq 1\}$. Thus, $\norm{\C}_*$ is the convex envelope of the tensor average rank $\rankt(\C)$ over $S$.
\end{proof}

\subsection{Proof of Theorem 3.2}

\begin{proof} Let $\G\in \partial \norm{\A}_*$. It is equivalent to the following statements   \cite{watson1992characterization}
	\begin{align}
	\norm{\A}_* &= \inproduct{\G}{\A}, \label{supp_subp1}\\
	\norm{\G} &\leq 1. \label{supp_subp2}
	\end{align}
	So, to   complete the proof, we only need to show that $\G = \U *\V^*+\W$, where $\U^**\W=\0$, $\W*\V=\0$ and $\norm{\W}\leq1$, satisfies (\ref{supp_subp1}) and (\ref{supp_subp2}). 
	First, we have
	\begin{align*}
	\inproduct{\G}{\A} = & \inproduct{\U*\V^*+\W}{\U*\Sbm*\V^*} \\
	= & \inproduct{\I}{\Sbm}+0 =  \frac{1}{n_3}\inproduct{\Imbar}{\Smbar} \\
	= & \frac{1}{n_3} \norm{\Ambar}_* =  \norm{\A}_*.
	\end{align*}
	Also, (\ref{supp_subp2}) is obvious when considering the property of $\W$. 
\end{proof}

\section{Dual Certification}
\label{supp_sec_dual}

In this section, we first introduce  conditions for  $(\LL_0,\Sbm_0)$  to be the unique solution to TRPCA in subsection \ref{supp_subdualcertif}.  Then we construct a dual certificate in subsection \ref{supp_subsecducergs} which satisfies the conditions in subsection \ref{supp_subdualcertif}, and thus our main result in Theorem 4.1 in our paper are proved.

\subsection{Dual Certificates}\label{supp_subdualcertif}

\begin{lemma}\label{supp_lem_dual2}
	Assume that $\norm{\Pomega\PT}\leq \frac{1}{2}$ and $\lambda<\frac{1}{\sqrt{n_3}}$. Then $(\LL_0,\Sbm_0)$ is the unique solution to the TRPCA problem if there is a pair $(\W,\F)$ obeying
	\begin{equation*}
	(\U*\V^*+\W)=\lambda(\sgn{\Sbm_0}+\F+\Pomega\D),
	\end{equation*}
	with $\PT\W=\0$, $\norm{\W}\leq\frac{1}{2}$, $\Pomega\F=\0$ and $\norm{\F}_{\infty}\leq\frac{1}{2}$, and $\norm{\Pomega\D}_F\leq\frac{1}{4}$.
\end{lemma}

\begin{proof}
	For any $\HH\neq\0$, $(\LL_0+\HH,\Sbm_0-\HH)$ is also a feasible solution. We show that its objective is larger than that at $(\LL_0,\Sbm_0)$, hence proving that $(\LL_0,\Sbm_0)$ is the unique solution. To do this, let $\U*\V^*+\W_0$ be an arbitrary subgradient of the tensor nuclear norm at $\LL_0$, and $\sgn{\Sbm_0}+\F_0$ be an arbitrary subgradient of the $\ell_1$-norm at $\Sbm_0$. Then we have
	\begin{align*}
	& \norm{\LL_0+\HH}_*+\lambda\norm{\Sbm_0-\HH}_1 \\
	\geq& \norm{\LL_0}_*+\lambda\norm{\Sbm_0}_1+\langle\U*\V^*+\W_0,\HH\rangle \\
	&-\lambda\inproduct{\sgn{\Sbm_0}+\F_0}{\HH}.
	\end{align*}
	Now pick $\W_0$ such that $\inproduct{\W_0}{\HH}=\norm{\PTo\HH}_*$ and $\inproduct{\F_0}{\HH}=-\norm{\Pomegao\HH}$. We have
	\begin{align*}
	& \norm{\LL_0+\HH}_*+\lambda\norm{\Sbm_0-\HH}_1 \\
	\geq& \norm{\LL_0}_*+\lambda\norm{\Sbm_0}_1+\norm{\PTo\HH}_*+\lambda\norm{\Pomegao\HH}_1\\
	&+\inproduct{\U*\V^*-\lambda\sgn{\Sbm_0}}{\HH}.
	\end{align*}
	By assumption
	\begin{align*}
	&\abs{\inproduct{\U*\V^*-\lambda\sgn{\Sbm_0}}{\HH}}\\
	\leq& \abs{\inproduct{\W}{\HH}}+\lambda\abs{\inproduct{\F}{\HH}}+\lambda\abs{\inproduct{\Pomega\D}{\HH}}\\
	\leq&\beta(\norm{\PTo\HH}_*+\lambda\norm{\Pomegao\HH}_1)+\frac{\lambda}{4}\norm{\Pomega\HH}_F,
	\end{align*}
	where $\beta=\max(\norm{\W},\norm{\F}_\infty)<\frac{1}{2}$. Thus	
	\begin{align*}
	& \norm{\LL_0+\HH}_*+\lambda\norm{\Sbm_0-\HH}_1 \\
	\geq& \norm{\LL_0}_*+\lambda\norm{\Sbm_0}_1+\frac{1}{2}(\norm{\PTo\HH}_*+\lambda\norm{\Pomegao\HH}_1) \\
	& -\frac{\lambda}{4}\norm{\Pomega\HH}_F.
	\end{align*}
	On the other hand,
	\begin{align*}
	\norm{\Pomega\HH}_F\leq & \norm{\Pomega\PT\HH}_F+\norm{\Pomega\PTo\HH}_F \\
	\leq & \frac{1}{2}\norm{\HH}_F+\norm{\PTo\HH}_F \\
	\leq & \frac{1}{2}\norm{\Pomega\HH}_F+\frac{1}{2}\norm{\Pomegao\HH}_F+\norm{\PTo\HH}_F.
	\end{align*}
	Thus
	\begin{align*}
	\norm{\Pomega\HH}_F\leq & \norm{\Pomegao\HH}_F+2\norm{\PTo\HH}_F \\
	\leq & \norm{\Pomegao\HH}_1+{2}{\sqrt{n_3}}\norm{\PTo\HH}_*.
	\end{align*}
	In conclusion, 
	\begin{align*}
	& \norm{\LL_0+\HH}_*+\lambda\norm{\Sbm_0-\HH}_1 \\
	\geq& \norm{\LL_0}_*+\lambda\norm{\Sbm_0}_1+\frac{1}{2}\left(1-{\lambda}{\sqrt{n_3}}\right)\norm{\PTo\HH}_*\\
	&+\frac{\lambda}{4}\norm{\Pomegao\HH}_1,
	\end{align*}
	where the last two terms are strictly positive when $\HH\neq \0$. Thus, the proof is completed.
\end{proof}

Lemma \ref{supp_lem_dual2} implies that it is suffices to produce a dual certificate $\W$ obeying
\begin{equation}\label{supp_conddualcertf}
\begin{cases}
\W \in \Tm^\bot, \\
\norm{\W} < \frac{1}{2},\\
\norm{\Pomega(\U*\V^*+\W-\lambda\sgn{\Sbm_0})}_F\leq \frac{\lambda}{4},\\
\norm{\Pomegao(\U*\V^*+\W)}_\infty < \frac{\lambda}{2}.
\end{cases}
\end{equation}

\subsection{Dual Certification via the Golfing Scheme}\label{supp_subsecducergs}
In this subsection, we show how to construct a dual certificate obeying (\ref{supp_conddualcertf}). Before we introduce our construction, our model assumes that $\Omegat\sim \Ber(\rho)$, or equivalently that $\Omegat^c\sim\Ber(1-\rho)$. Now the distribution of $\Omegat^c$ is the same as that of
$\Omegat^c = \Omegat_1 \cup \Omegat_2 \cup \cdots \cup \Omegat_{j_0}$, where each $\Omegat_j$  follows the Bernoulli model with parameter $q$, which satisfies
\begin{equation*}
\mathbb{P}((i,j,k)\in\Omegat) = \mathbb{P}(\text{Bin}(j_0,q)=0) = (1-q)^{j_0},
\end{equation*}
so that the two models are the same if 
$\rho = (1-q)^{j_0}$.
Note that  because of overlaps between the $\Omegat_j$'s, $q\geq (1-\rho)/j_0$.

Now, we construct a dual certificate
\begin{equation}
\W = \WL+\WS,
\end{equation}
where each component is as follows:
\begin{enumerate}
	\item  Construction of $\WL$ via the Golfing Scheme. Let $j_0= 2\log(nn_3)$ and $\Omegat_j$, $j=1,\cdots,j_0$, be defined as previously described so that $\Omegat^c = \cup_{1\leq j \leq j_0}\Omegat_j $. Then define
	\begin{equation}\label{supp_eq_wl}
	\WL = \PTo\Y_{j_0},
	\end{equation}
	where 
	\begin{equation*}
	\Y_j = \Y_{j-1} + q^{-1}\PP_{\Omegat_j}\PT(\U*\V^*-\Y_{j-1}), \ \Y_0=\0.
	\end{equation*}
	\item Construction of $\WS$ via the Method of Least Squares. Assume that $\norm{\Pomega\PT}<1/2$. 
	Then, $\norm{\Pomega\PT\Pomega}<1/4$, and thus, the operator $\Pomega-\Pomega\PT\Pomega$  mapping $\Omegat$ onto itself is invertible; we denote its inverse by $(\Pomega-\Pomega\PT\Pomega)^{-1}$. We then set
	\begin{equation}\label{supp_eq_ws}
	\WS = \lambda \PTo (\Pomega-\Pomega\PT\Pomega)^{-1} \sgn{\Sbm_0}.
	\end{equation}
	This is equivalent to 
	\begin{equation*}
	\WS = \lambda \PTo \sum_{k\geq 0}   (\Pomega\PT\Pomega)^{k} \sgn{\Sbm_0}.
	\end{equation*}
\end{enumerate}
Since both $\WL$ and $\WS$ belong to $\Tm^{\bot}$ and $\Pomega\WS  = \lambda  \Pomega (\I-\PT) (\Pomega-\Pomega\PT\Pomega)^{-1} \sgn{\Sbm_0}=\lambda\sgn{\Sbm_0}$, we will establish that $\WL+\WS$ is a valid dual certificate if it obeys 
\begin{equation}
\begin{cases}
\norm{\WL+\WS} < \frac{1}{2}, \\
\norm{\Pomega(\U*\V^*+\WL)}_F\leq \frac{\lambda}{4}, \\
\norm{ \Pomegao(\U*\V^*+\WL+\WS}_\infty < \frac{\lambda}{2}.
\end{cases}
\end{equation}
This can be achieved by using the following two key lemmas:
\begin{lemma}\label{supp_eq_keylem1}
	Assume that $\Omegat\sim\Ber(\rho)$ with parameter $\rho\leq \rho_s$ for some $\rho_s>0$. Set $j_0=2\lceil\log (nn_3)\rceil$ (use $\log(\none n_3)$ for the tensors of rectangular frontal slice). Then,  the tensor $\WL$ obeys
	\begin{enumerate}[(a)]
		\item $\norm{\WL} < \frac{1}{4}$,
		\item $\norm{\Pomega(\U*\V^*+\WL)}_F<\frac{\lambda}{4}$,
		\item $\norm{\Pomegao(\U*\V^*+\WL) }_\infty < \frac{\lambda}{4}$.
	\end{enumerate}
\end{lemma}
\begin{lemma}\label{supp_eq_keylem2}
	Assume that $\Sbm_0$ is supported on a set $\Omegat$ sampled as in Lemma \ref{supp_eq_keylem1}, and that the signs of $\Sbm_0$ are independent and identically distributed symmetric (and independent of $\Omegat$). Then, the tensor $\WS$ (\ref{supp_eq_ws}) obeys
	\begin{enumerate}[(a)]
		\item $\norm{\WS} < \frac{1}{4}$,
		\item $\norm{\Pomegao\WS }_\infty < \frac{\lambda}{4}$.
	\end{enumerate}
\end{lemma}

So the left task is to prove Lemma \ref{supp_eq_keylem1} and Lemma \ref{supp_eq_keylem2}, which are given in Section \ref{supp_sec_proofofdual}.

\section{Proofs of Dual Certification}
\label{supp_sec_proofofdual}

This section gives the proofs of Lemma \ref{supp_eq_keylem1} and Lemma \ref{supp_eq_keylem2}. To do this, we first introduce some lemmas with their proofs given in Section \ref{supp_sec_prooflemmas}. 

\begin{lemma}\label{supp_lemspectrm}
	For the Bernoulli sign variable $\M\in\mathbb{R}^{n\times n\times n_3}$ defined as
	\begin{equation}\label{supp_defM0}
	\M_{ijk} = \begin{cases}
	1, & \text{w.p.} \ \rho/2, \\
	0, & \text{w.p.} \ 1-\rho, \\
	-1, & \text{w.p.} \ \rho/2,
	\end{cases}
	\end{equation}
	where $\rho>0$, there exists a function $\varphi(\rho)$  satisfying  $\lim\limits_{\rho\rightarrow0^+}\varphi(\rho)=0$, such that the following statement holds with with large probability,
	\begin{equation*}\label{supp_boundspecm}
	\norm{\M} \leq \varphi(\rho)\sqrt{ nn_3}.
	\end{equation*}
\end{lemma}

\begin{lemma}\label{supp_lem_kem1}
	Suppose $\Omegat\sim\Ber(\rho)$. Then with high probability,
	\begin{equation*}
	\norm{\PT-\rho^{-1}\PT\Pomega\PT}\leq\epsilon,
	\end{equation*}
	provided that $\rho\geq C_0\epsilon^{-2}(\mu r\log (nn_3))/(nn_3)$ for some numerical constant $C_0>0$. For the tensor of  rectangular frontal slice, we need $\rho\geq C_0\epsilon^{-2}(\mu r\log (\none n_3))/(\ntwo n_3)$.
\end{lemma}
\begin{cor}\label{supp_corollar31}
	Assume that $\Omegat\sim\Ber(\rho)$, then $\norm{\Pomega\PT}^2\leq \rho+\epsilon$, provided that $1-\rho\geq C\epsilon^{-2}(\mu r\log(nn_3))/(nn_3)$, where $C$ is as in Lemma \ref{supp_lem_kem1}. For the tensor with frontal slice, the modification is as in Lemma  \ref{supp_lem_kem1}.
\end{cor}
Note that this corollary shows that $\norm{\Pomega\PT}\leq 1/2$, provided $|\Omegat|$ is not too large. 
\begin{lemma}\label{supp_lem_keyinf}
	Suppose that $\Z\in\Tm$ is a fixed tensor, and $\Omegat\sim\Ber(\rho)$. Then, with high probability,
	\begin{equation}
	\norm{\Z-\rho^{-1}\PT\Pomega\Z}_{\infty} \leq \epsilon \norm{\Z}_\infty,
	\end{equation}
	provided that $\rho\geq C_0 \epsilon^{-2}(\mu r \log(nn_3))/(nn_3)$ (for the tensor of  rectangular frontal slice, $\rho\geq C_0\epsilon^{-2}(\mu r\log (\none n_3))/(\ntwo n_3)$) for some numerical constant $C_0>0$.
\end{lemma}

\begin{lemma}\label{supp_lempre3}
	Suppose $\Z$ is fixed, and $\Omegat\sim\Ber(\rho)$. Then, with high probability,
	\begin{equation}
	\norm{(\I-\rho^{-1}\Pomega)\Z} \leq \sqrt{\frac{C_0nn_3\log(nn_3)}{\rho}} \norm{\Z}_\infty,
	\end{equation}
	for some numerical constant $C_0>0$ provided that $\rho\geq C_0\log(nn_3)/(nn_3)$ (or $\rho\geq C_0\log(\none n_3)/(\ntwo n_3)$ for the tensors with rectangular frontal slice).
\end{lemma} 

\subsection{Proof of Lemma \ref{supp_eq_keylem1}}
\begin{proof}
	We first introduce some notations. Set $\Z_j =   \U*\V^* - \PT\Y_j$ obeying
	\begin{equation*}
	\Z_j = (\PT-q^{-1}\PT\PP_{\Omegat_j}\PT)\Z_{j-1}.
	\end{equation*}
	So $\Z_j\in\T$ for all $j\geq 0$. Also, note that when
	\begin{equation}\label{supp_eq_q}
	q\geq C_0\epsilon^{-2}\frac{\mu r\log(nn_3)}{nn_3},
	\end{equation}
	or for the tensors with rectangular frontal slices $q\geq C_0\epsilon^{-2}\frac{\mu r\log(\none n_3)}{\ntwo n_3}$,
	we have
	\begin{equation}\label{supp_eq_Zrel1}
	\norm{\Z_j}_\infty \leq \epsilon \norm{\Z_{j-1}}_\infty\leq \epsilon^j \norm{\U*\V^*}_\infty,
	\end{equation}
	by Lemma \ref{supp_lem_keyinf} and 
	\begin{equation}\label{supp_eq_Zrel2}
	\norm{\Z_j}_F \leq \epsilon \norm{\Z_{j-1}}_F \leq \epsilon^j \norm{\U*\V^*}_F \leq   \epsilon^j \sqrt{r}.
	\end{equation}
	We assume $\epsilon\leq e^{-1}$.
	
	\noindent\textbf{1. Proof of (a).} Note that $\Y_{j_0} = \sum_j q^{-1}\PP_{\Omegat_j}\Z_{j-1}$. We have
	\begin{align*}
	\norm{\WL} = &\norm{\PTo\Y_{j_0}} \leq \sum_j \norm{q^{-1} \PTo\PP_{\Omegat_j} \Z_{j-1} } \\
	\leq & \sum_j \norm{\PTo(q^{-1}\PP_{\Omegat_j}\Z_{j-1}-\Z_{j-1})} \\
	\leq & \sum_j  \norm{q^{-1}\PP_{\Omegat_j}\Z_{j-1}-\Z_{j-1}} \\
	\leq & C_0'\sqrt{\frac{nn_3\log(nn_3)}{q}} \sum_j \norm{\Z_{j-1}}_\infty \\
	\leq & C_0'\sqrt{\frac{nn_3\log(nn_3)}{q}} \sum_j \epsilon^{j-1}\norm{\U*\V^*}_\infty \\
	\leq & C_0'(1-\epsilon)^{-1} \sqrt{\frac{nn_3\log(nn_3)}{q}}\norm{\U*\V^*}_\infty.
	\end{align*}
	The fourth step is from Lemma \ref{supp_lempre3} and the fifth is from (\ref{supp_eq_Zrel1}). Now by using (\ref{supp_eq_q}) and (\ref{supp_tic3}), we have 
	$$\norm{\WL}\leq C'\epsilon,$$
	for some numerical constant $C'$.

	\noindent\textbf{2. Proof of (b).} Since $\Pomega\Y_{j_0} = \0$, $\Pomega(\U*\V^*+\PTo\Y_{j_0}) =  \Pomega(\U*\V^*-\PT\Y_{j_0}) = \Pomega(\Z_{j_0}) $, and it follows from (\ref{supp_eq_Zrel2}) that $$\norm{\Z_{j_0}}_F \leq \epsilon^{j_0} \norm{\U*\V^*}_F \leq \epsilon^{j_0} \sqrt{r}.$$
	Since $\epsilon\leq e^{-1}$ and $j_0 \geq 2\log(nn_3)$, $\epsilon^{j_0}\leq (nn_3)^{-2}$ and this proves the claim.
	
	\noindent\textbf{3. Proof of (c).} We have $\U*\V^*+\WL = \Z_{j_0}+\Y_{j_0}$ and know that $\Y_{j_0}$ is supported on $\Omegat^c$. Therefore, since $\norm{\Z_{j_0}}_F\leq \lambda/8$. We only need to show that $\norm{\Y_{j_0}}_\infty\leq\lambda/8$. Indeed,
	\begin{align*}
	\norm{\Y_{j_0}}_\infty \leq & q^{-1}\sum_j \norm{\PP_{\Omegat_j} \Z_{j-1}}_\infty \\
	\leq & q^{-1}\sum_j \norm{ \Z_{j-1}}_\infty \\
	\leq & q^{-1}\sum_j\epsilon^{j-1} \norm{  \U*\V^*}_\infty.
	\end{align*}
	Since $\norm{\U*\V^*}_\infty \leq \sqrt{\frac{\mu r}{n^2n^2_3}}$, this gives
	\begin{equation*}
	\norm{\Y_{j_0}}_\infty \leq C'\frac{\epsilon^2}{\sqrt{\mu r(\log(nn_3))^2}},
	\end{equation*}
	for some numerical constant $C'$ whenever $q$ obeys (\ref{supp_eq_q}). Since $\lambda = 1/\sqrt{nn_3}$, $\norm{\Y_{j_0}}_\infty \leq \lambda/8$ if
	\begin{align*}
	\epsilon\leq C \left(\frac{\mu r(\log(nn_3))^2}{nn_3}\right)^{1/4}.
	\end{align*}
	The claim is proved by using (\ref{supp_eq_q}), (\ref{supp_tic3}) and sufficiently small $\epsilon$ (provided that $\rho_r$ is sufficiently small. Note that everything is consistent since $C_0\epsilon^{-2}\frac{\mu r\log(nn_3)}{nn_3}<1$.
\end{proof}

\subsection{Proof of Lemma \ref{supp_eq_keylem2}}
\begin{proof}
	We denote $\M=\sgn{\Sbm_0}$ distributed as
	\begin{equation*}\label{supp_defM}
	\M_{ijk} = \begin{cases}
	1, & \text{w.p.} \ \rho/2, \\
	0, & \text{w.p.} \ 1-\rho, \\
	-1, & \text{w.p.} \ \rho/2.
	\end{cases}
	\end{equation*}
	Note that for any $\sigma>0$, $\{\norm{\Pomega\PT}\leq\sigma\}$ holds with high probability provided that $\rho$ is sufficiently small, see Corollary \ref{supp_corollar31}.
	
	\noindent\textbf{1. Proof of (a).} By construction,
	\begin{align*}
	\WS = &\lambda\PTo\M+\lambda\PTo\sum_{k\geq 1} (\Pomega\PT\Pomega)^k\M \\
	:=&  \PTo\WS_0 + \PTo\WS_1.
	\end{align*}
	Note that  $\norm{\PTo\WS_0} \leq \norm{\WS_0} = \lambda\norm{\M}$ and $\norm{\PTo\WS_1} \leq \norm{\WS_1} = \lambda\norm{\R(\M)}$, where $\R = \sum_{k\geq1} (\Pomega\PT\Pomega)^k$. Now, we will respectively show that $\lambda\norm{\M}$ and $\lambda\norm{\R(\M)}$ are small enough when $\rho$ is sufficiently small for $\lambda=1/\sqrt{nn_3}$. Therefor, $\norm{\WS}\leq1/4$. 
	
	\textbf{1) Bound $\lambda\norm{\M}$.} 
	
	By using Lemma \ref{supp_lemspectrm} directly, we have that $\lambda\norm{\M}\leq \varphi(\rho)$ is sufficiently small given $\lambda=1/\sqrt{nn_3}$ and $\rho$ is sufficiently small.

	\textbf{2) Bound $\norm{\R(\M)}$.} 
	
	For simplicity, let $\Z=\R(\M)$. We have
	\begin{equation}\label{supp_boundrm1}
	\norm{\Z}=\norm{\Zmbar}=\sup_{\x\in\mathbb{S}^{nn_3-1}} \norm{\Zmbar\x}_2.
	\end{equation}
	The optimal $\x$ to (\ref{supp_boundrm1}) is an eigenvector of $\Zmbar^*\Zmbar$. Since $\Zmbar$ is a block diagonal matrix, the optimal $\x$ has a  block sparse structure, i.e., 
	$\x \in B=\big\{  \x  \in\mathbb{R}^{nn_3} | \x = [\x_1^\top,\cdots , \x_i^\top \cdots, \x^\top_{n_3} ], \text{with }  \x_i\in\mathbb{R}^n, \text{ and there exists } j \text{ such that } \x_j\neq\0 \text{ and } \x_i=\0, i\neq j  \}$. Note that $\norm{\x}_2 = \norm{\x_j}_2 = 1$. Let $N$ be the $1/2$-net for $\mathbb{S}^{n-1}$ of size at most $5^n$ (see Lemma 5.2 in \cite{vershynin2010introduction}). Then the $1/2$-net, denoted as $N'$, for $B$ has the size at most $n_3\cdot 5^n$.  We have 
	\begin{align*}
	\norm{\R(\M)} =  &\norm{\bdiag( \overline{ \R(\M) }  )} \\
	= & \sup_{\x,\y\in B} \inproduct{\x}{\bdiag( \overline{ \R(\M) }  )\y} \\ 
	= & \sup_{\x,\y\in B}  \inproduct{\x\y^*}{\bdiag(\overline{\R({\M)}})}  \\
	=& \sup_{\x,\y\in B}  \inproduct{\bdiag^*(\x\y^*)}{\overline{\R({\M)}}} ,
	\end{align*}
	where $\bdiag^*$, the joint operator of $\bdiag$, maps the block diagonal matrix $\x\y^*$ to a tensor of size $n\times n\times n_3$. Let $\Z' = \bdiag^*(\x\y^*)$ and $\Z = \mcode{ifft}(\Z',[\ ],3)$. We have
	\begin{align*}
	\norm{\R(\M)} =& \sup_{\x,\y\in B}   \inproduct{\Z'}{\overline{\R({\M)}}} \\
	=&\sup_{\x,\y\in B} n_3   \inproduct{\Z}{{\R({\M)}}}  \\
	=&\sup_{\x,\y\in B} n_3   \inproduct{\R(\Z)}{{{\M}}}  \\
	\leq & \sup_{\x,\y\in N'} 4 n_3   \inproduct{\R(\Z)}{{{\M}}}.
	\end{align*}
	For a fixed pair $(\x,\y)$ of unit-normed vectors, define the random variable
	\begin{equation*}
	X(\x,\y) = \inproduct{4n_3\R(\Z)}{{{\M}}}.
	\end{equation*}
	Conditional on $\Omegat=\text{supp}(\M)$, the signs of $\M$ are independent and identically distributed symmetric and Hoeffding's inequality gives
	\begin{align*}
	\mathbb{P}(|X(\x,\y)|>t| \Omegat) \leq 2\exp\left(\frac{-2t^2}{\norm{4n_3\R(\Z)}_F^2}\right).
	\end{align*}
	Note that $\norm{4n_3\R(\Z))}_F\leq 4n_3\norm{\R}\norm{\Z}_F = 4\sqrt{n_3}\norm{\R}\norm{\Z'}_F =4\sqrt{n_3}\norm{\R}$. Therefore, we have 
	\begin{align*}
	\mathbb{P}\left( \sup_{\x,\y\in N'} |X(\x,\y)| >t | \Omegat \right) \leq 2|N'|^2\exp\left(-\frac{t^2}{8n_3\norm{\R}^2}\right).
	\end{align*}
	Hence,
	\begin{align*}
	\mathbb{P}\left( \norm{\R(\M)} >t | \Omegat \right) \leq 2|N'|^2\exp\left(-\frac{t^2}{8n_3\norm{\R}^2}\right).
	\end{align*}
	On the event $\{\norm{\Pomega\PT} \leq\sigma \}$,
	\begin{align*}
	\norm{\R} \leq \sum_{k\geq1} \sigma^{2k} = \frac{\sigma^2}{1-\sigma^2},
	\end{align*}
	and, therefore, unconditionally,
	\begin{align*}
	& \mathbb{P}\left( \norm{\R(\M)} >t   \right) \\
	\leq & 2|N'|^2\exp\left(-\frac{\gamma^2t^2}{8n_3}\right) + \mathbb{P}\left( \norm{\Pomega\PT} \geq \sigma   \right), \ \gamma =\frac{1-\sigma^2}{2\sigma^2} \\
	=& 2n_3^2 \cdot 5^{2n} \exp\left(-\frac{\gamma^2t^2}{8n_3}\right) + \mathbb{P}\left( \norm{\Pomega\PT} \geq \sigma   \right).
	\end{align*}
	Let $t=c\sqrt{nn_3}$, where $c$ can be a small absolute constant. Then the above inequality implies that $\norm{\R(\M)}\leq t$ with high probability.

	\noindent\textbf{2. Proof of (b)}
	Observe that 
	\begin{align*}
	\Pomegao\WS = -\lambda\Pomegao\PT(\Pomega-\Pomega\PT\Pomega)^{-1}\M.
	\end{align*}
	Now for $(i,j,k)\in\Omegat^c$, $\WS_{ijk} = \inproduct{\WS}{\eijk}$, and we have $\WS_{ijk} =\lambda \inproduct{ \Q(i,j,k)}{\M}$, where $\Q(i,j,k)$ is the tensor $-(\Pomega-\Pomega\PT\Pomega)^{-1}\Pomega\PT(\eijk)$. Conditional on $\Omegat = \text{supp}(\M)$, the signs of $\M$ are independent and identically distributed symmetric, and the Hoeffding's inequality gives
	\begin{equation*}
	\mathbb{P}(|\WS_{ijk}| > t\lambda|\Omegat) \leq 2\exp\left( -\frac{2t^2}{\norm{\Q(i,j,k)}_F^2}\right),
	\end{equation*}
	and
	\begin{align*}
	&\mathbb{P}(\sup_{i,j,k}|\WS_{ijk}| > t\lambda/n_3|\Omegat) \\
	\leq & 2n^2n_3\exp\left( -\frac{2t^2}{\sup_{i,j,k}\norm{\Q(i,j,k)}_F^2}\right).
	\end{align*}
	By using (\ref{supp_proabouPT}), we have
	\begin{align*}
	\norm{\Pomega\PT(\eijk)}_F \leq & \norm{\Pomega\PT}\norm{\PT(\eijk)}_F \\
	\leq & \sigma\sqrt{\frac{2\mu r}{nn_3}},
	\end{align*}
	on the event $\{\norm{\Pomega\PT}\leq\sigma\}$. On the same event, we have $\norm{(\Pomega-\Pomega\PT\Pomega)^{-1}}\leq (1-\sigma^2)^{-1}$ and thus
	$\norm{\Q(i,j,k)}_F^2 \leq \frac{2\sigma^2}{(1-\sigma^2)^2}\frac{\mu r}{nn_3}$.
	Then, unconditionally,
	\begin{align*}
	& \mathbb{P}\left(\sup_{i,j,k}|\WS_{ijk}| >t\lambda\right)  \\
	\leq & 2n^2n_3\exp\left( -\frac{nn_3\gamma^2t^2}{\mu r} \right) + \mathbb{P}(\norm{\Pomega\PT}\geq\sigma),
	\end{align*}
	where $\gamma=\frac{(1-\sigma^2)^2}{2\sigma^2}$. This proves the claim when $\mu r<\rho'_rnn_3\log(nn_3)^{-1}$ and $\rho_r'$ is sufficiently small.
\end{proof}

\section{Proofs of Some  Lemmas}\label{supp_sec_prooflemmas}

\begin{lemma} \cite{tropp2012user}   \label{supp_lembenmatrix}
	Consider a finite sequence $\{\Zm_k\}$ of independent, random $n_1\times n_2$ matrices that satisfy the assumption $\mathbb{E} \Zm_k=\0$ and $\norm{\Zm_k}\leq R$ almost surely. Let $\sigma^2 = \max\{\norm{\sum_k\mathbb{E}[\Zm_k\Zm_k^*]} , \max\{\norm{\sum_k\mathbb{E}[\Zm_k^*\Zm_k]} \}$. Then, for any $t\geq0$, we have
	\begin{align*}
	\mathbb{P}\left[ \normlarge{\sum_k\Zm_k} \geq t \right] \leq & (n_1+n_2) \exp\left( -\frac{t^2}{2\sigma^2+\frac{2}{3}Rt} \right) \\
	\leq &  (n_1+n_2) \exp\left( -\frac{3t^2}{8\sigma^2} \right), \ \text{for } t\leq\frac{\sigma^2}{R}.
	\end{align*}
	Or, for any $c>0$, we have
	\begin{align*}\label{supp_eq_tropbound2}
	\normlarge{\sum_k\Zm_k} \geq 2\sqrt{c\sigma^2\log(n_1+n_2)} + cB\log(n_1+n_2),
	\end{align*}
	with probability at least $1-(n_1+n_2)^{1-c}$.
\end{lemma}

\subsection{Proof of Lemma \ref{supp_lemspectrm}}

\begin{proof}
	The proof has three steps.
	
	\textit{Step 1: Approximation.}
	We first introduce some notations. Let $\f_i^*$ be the $i$-th row of $\F_{n_3}$,  and $\Mh=\begin{bmatrix}
	\Mh_1 \\ \Mh_2 \\ \vdots \\ \Mh_{n} 
	\end{bmatrix} \in\mathbb{R}^{nn_3\times n}$ be a matrix unfolded by $\M$, where $\Mh_i\in\mathbb{R}^{n_3\times n}$ is the $i$-th horizontal slice of $\M$, i.e., $[\Mh_i]_{kj} = \M_{ikj}$. Consider that $\Mbar = \mcode{fft}(\M,[\ ],3)$, we have
	\begin{equation*}\label{supp_pfl351}
	\Mmbar_i = \begin{bmatrix}
	\f_i^*\Mh_1 \\  \f_i^*\Mh_2 \\ \vdots \\  \f_i^*\Mh_{n} 
	\end{bmatrix},
	\end{equation*}\label{supp_pfl352}
	where $\Mmbar_i\in\mathbb{R}^{n\times n}$ is the $i$-th frontal slice of $\M$. Note that 
	\begin{equation}\label{supp_pfl3544440}
	\norm{\M} = \norm{\Mmbar} = \max_{i=1,\cdots,n_3} \ \norm{\Mmbar_i}.
	\end{equation}
	Let $N$ be the $1/2$-net for $\mathbb{S}^{n-1}$ of size at most $5^n$ (see Lemma 5.2 in \cite{vershynin2010introduction}). Then Lemma 5.3 in  \cite{vershynin2010introduction} gives 
	\begin{equation}\label{supp_prf13555}
	\norm{\Mmbar_i} \leq 2 \ \max_{\x\in N} \ \norm{\Mmbar_i\x}_2.
	\end{equation}
	So we consider to bound $\norm{\Mmbar_i\x}_2$. 
	
	\textit{Step 2: Concentration.} 
	We can express $\norm{\Mmbar_i\x}_2^2$ as a sum of independent random variables
	\begin{equation}\label{supp_pfl353}
	\norm{\Mmbar_i\x}_2^2 = \sum_{j=1}^{n} (\f_i^*\Mh_j\x)^2 := \sum_{j=1}^{n} z_j^2,
	\end{equation}
	where $z_j = \langle {\Mh_j},{\f_i\x^*}\rangle$, $j=1,\cdots,n$, are independent sub-gaussian random variables with $\mathbb{E}z_j^2=\rho\norm{\f_i\x^*}_F^2=\rho n_3$. Using (\ref{supp_defM0}), we have
	\begin{equation*} 
	|[\Mh_j]_{kl}| = \begin{cases}
	1, & \text{w.p.} \ \rho, \\
	0, & \text{w.p.} \ 1-\rho.
	\end{cases}
	\end{equation*}
	Thus, the sub-gaussian norm of $[\Mh_j]_{kl}$, denoted as $\norm{\cdot}_{\psi_2}$, is 
	\begin{align*}
	\norm{[\Mh_j]_{kl}}_{\psi_2} = & \sup_{p\geq 1} p^{-\frac{1}{2}}(\mathbb{E}[|[\Mh_j]_{kl}|^p])^{\frac{1}{p}} \\
	= & \sup_{p\geq 1} p^{-\frac{1}{2}}\rho^{\frac{1}{p}}.
	\end{align*}
	Define the function $\phi(x) =x^{-\frac{1}{2}}\rho^{\frac{1}{x}}$ on $[1,+\infty)$. The only stationary point occurs at $x^*=\log \rho^{-2}$. Thus, 
	\begin{align}
	\phi(x)\leq & \max(\phi(1),\phi(x^*)) \notag\\
	= &\max\left(\rho,(\log\rho^{-2})^{-\frac{1}{2}}\rho^{\frac{1}{\log \rho^{-2}}}\right) \notag\\
	:= & \psi(\rho). \label{supp_abcefeqdd}
	\end{align}
	Therefore,  $\norm{[\Mh_j]_{kl}}_{\psi_2}\leq   \psi(\rho)$. Consider that $z_j$ is a sum of independent centered sub-gaussian random variables $[\Mh_j]_{kl}$'s, by using Lemma 5.9 in  \cite{vershynin2010introduction}, we have $\norm{z_j}^2_{\psi_2}\leq c_1(\psi(\rho))^2 n_3$, where $c_1$ is an absolute constant. Therefore, by Remark 5.18 and Lemma 5.14 in \cite{vershynin2010introduction}, $z_j^2-\rho n_3$ are independent centered sub-exponential random variables with $\norm{z_j^2-\rho n_3}_{\psi_1} \leq 2 \norm{z_j}^2_{\psi_1} \leq 4 \norm{z_j}^2_{\psi_2}\leq 4c_1 (\psi(\rho))^2 n_3$.
	
	Now, we use an exponential deviation inequality, Corollary 5.17 in \cite{vershynin2010introduction}, to control the sum  (\ref{supp_pfl353}). We have
	\begin{align*}
	& \mathbb{P}\left(| \norm{\Mmbar_i\x}_2^2 -\rho nn_3 | \geq tn \right) \\
	= & \mathbb{P}\left(\left| \sum_{j=1}^{n} (z_j^2-\rho n_3) \right| \geq tn \right) \\
	\leq & 2\exp\left( -c_2n\min \left( \left(\frac{t}{4c_1 (\psi(\rho))^2 n_3}\right)^2, \frac{t}{4c_1(\psi(\rho))^2 n_3}\right) \right),
	\end{align*}
	where $c_2>0$. Let $t = c_3 (\psi(\rho))^2 n_3$ for some absolute constant $c_3$, we have   
	\begin{align*}
	& \mathbb{P}\left(| \norm{\Mmbar_i\x}_2^2 -\rho nn_3 | \geq c_3(\psi(\rho))^2 nn_3 \right) \\
	\leq & 2\exp\left( -c_2n\min\left( \left(\frac{c_3}{4c_1}\right)^2,\frac{c_3}{4c_1}\right) \right).
	\end{align*}

	\textit{Step 3： Union bound.}
	Taking the union bound over all $\x$ in the net $N$ of cardinality $|N|\leq 5^n$, we obtain
	\begin{align*}
	&\mathbb{P}\left(\left| \max_{\x \in N} \norm{\Mmbar_i\x}_2^2 -\rho nn_3 \right| \geq c_3(\psi(\rho))^2 nn_3 \right) \\
	\leq & 2\cdot 5^n \cdot\exp\left( -  c_2n\min\left(\left(\frac{c_3}{4c_1}\right)^2,\frac{c_3}{4c_1}\right) \right).
	\end{align*}
	Furthermore, taking the union bound over all $i=1,\cdots,n_3$, we have 
	\begin{align*}
	&\mathbb{P}\left(\max_i \ \left| \max_{\x \in N} \norm{\Mmbar_i\x}_2^2 -\rho nn_3 \right| \geq c_3(\psi(\rho))^2 nn_3 \right) \notag\\
	\leq &2\cdot 5^n\cdot n_3 \cdot \exp\left( -  c_2n\min\left(\left(\frac{c_3}{4c_1}\right)^2,\frac{c_3}{4c_1}\right) \right). \label{supp_prfnlemmada000}
	\end{align*}
	This implies that, with high probability (when the constant $c_3$ is  large enough),
	\begin{equation}\label{supp_profboundm2}
	\max_i \  \max_{\x \in N} \norm{\Mmbar_i\x}_2^2 \leq (\rho+c_3(\psi(\rho))^2)  nn_3.
	\end{equation}
	Let $\varphi(\rho) = 2\sqrt{\rho+c_3(\psi(\rho))^2}$ and it satisfies  $\lim\limits_{\rho\rightarrow0^+}\varphi(\rho)=0$ by using (\ref{supp_abcefeqdd}). The proof is completed by further combining  (\ref{supp_pfl3544440}), (\ref{supp_prf13555}) and (\ref{supp_profboundm2}).
\end{proof}

\subsection{Proof of Lemma \ref{supp_lem_kem1}}
\begin{proof}
	For any tensor $\Z$, we can write
	\begin{align*}
	&(\rho^{-1}\PT\Pomega\PT-\PT)\Z \\
	= &\sum_{ijk}\left(\rho^{-1}{\delta_{ijk}}-1\right)\inproduct{\eijk }{\PT\Z}\PT(\eijk) \\
	:=&\sum_{ijk} \HH_{ijk}(\Z)
	\end{align*}
	where $\HH_{ijk}: \mathbb{R}^{\nss}\rightarrow\mathbb{R}^{\nss}$ is a self-adjoint random operator with $\mathbb{E}[\HH_{ijk}]=\0$. Define the matrix operator $\Hmbar_{ijk}: \mathbb{B}\rightarrow \mathbb{B}$, where $\mathbb{B}=\{\Bmbar: \B\in \mathbb{R}^{\nss} \}$ denotes the set consists of block diagonal matrices with the blocks as the frontal slices of $\Bbar$,  as
	\begin{align*}
	\Hmbar_{ijk}(\Zmbar) = & \left(\rho^{-1}{\delta_{ijk}}-1\right)\inproduct{\eijk }{\PT(\Z)}\bdiag(\overline{\PT(\eijk)}).
	\end{align*}
	By the above definitions, we have $\norm{\HH_{ijk}} = \norm{\Hmbar_{ijk}}$ and $\norm{\sum_{ijk}\HH_{ijk}} = \norm{\sum_{ijk}\Hmbar_{ijk}}$. Also $\Hmbar_{ijk}$ is  self-adjoint and $\mathbb{E}[\Hmbar_{ijk}]=0$. 
	To prove the result by the non-commutative Bernstein inequality, we need to bound $\norm{\Hmbar_{ijk}}$ and $\normlarge{\sum_{ijk}\mathbb{E}[\Hmbar^2_{ijk}]}$. First, we have
	\begin{align*}
	\norm{\Hmbar_{ijk}} = & \sup_{\norm{\Zmbar}_F=1} \norm{\Hmbar_{ijk}(\Zmbar)}_F \\
	\leq &  \sup_{\norm{\Zmbar}_F=1} \rho^{-1} \norm{\PT(\eijk)}_F \norm{\bdiag(\overline{\PT(\eijk)})}_F \norm{\Z}_F \\
	= & \sup_{\norm{\Zmbar}_F=1} \rho^{-1} \norm{\PT(\eijk)}_F^2 \norm{\Zmbar}_F \\
	\leq & \frac{2\mu r}{nn_3\rho},
	\end{align*}
	where the last inequality uses (\ref{supp_proabouPT}). On the other hand, by direct computation, we have $\Hmbar_{ijk}^2(\Zmbar) = (\rho^{-1}\delta_{ijk}-1)^2\inproduct{\eijk}{\PT(\Z)}\inproduct{\eijk}{\PT(\eijk)}\bdiag(\overline{\PT(\eijk)})$. Note that $\mathbb{E}[(\rho^{-1}\delta_{ijk}-1)^2]\leq \rho^{-1}$. We have
	\begin{align*}
	& \normlarge{\sum_{ijk}\mathbb{E}[\Hmbar^2_{ijk}(\Zmbar)]}_F \\
	\leq & \rho^{-1}\normlarge{\sum_{ijk} \inproduct{\eijk}{\PT(\Z)}\inproduct{\eijk}{\PT(\eijk)}\bdiag(\overline{\PT(\eijk)})}_F \\
	\leq & \rho^{-1} \sqrt{n_3} \norm{\PT(\eijk)}_F^2\normlarge{\sum_{ijk} \inproduct{\eijk}{\PT(\Z)} }_F \\
	= & \rho^{-1}\sqrt{n_3} \norm{\PT(\eijk)}_F^2\norm{\PT(\Z)}_F\\
	\leq & \rho^{-1}\sqrt{n_3} \norm{\PT(\eijk)}_F^2\norm{\Z}_F\\
	= & \rho^{-1} \norm{\PT(\eijk)}_F^2\norm{\Zmbar}_F\\
	\leq & \frac{2\mu r}{nn_3\rho}\norm{\Zmbar}_F.	
	\end{align*}
	This implies $\normlarge{\sum_{ijk}\mathbb{E}[\Hmbar^2_{ijk}]}\leq \frac{2\mu r}{nn_3\rho}$.  Let $\epsilon\leq 1$. By Lemma \ref{supp_lembenmatrix}, we have
	\begin{align*}
	& \mathbb{P} \left[\norm{\rho^{-1}\PT\Pomega\PT-\PT} > \epsilon \right] \\
	= & \mathbb{P}\left[ \normlarge{  \sum_{ijk} {\HH}_{ijk} } > \epsilon\right] \\
	= & \mathbb{P}\left[ \normlarge{  \sum_{ijk} {\Hmbar}_{ijk} } > \epsilon \right] \\
	\leq & 2nn_3 \exp\left( -\frac{3}{8} \cdot \frac{\epsilon^2}{2\mu r/(nn_3\rho) } \right) \\
	\leq & 2(nn_3)^{1-\frac{3}{16}C_0},
	\end{align*}
	where the last inequality uses  $\rho\geq C_0\epsilon^{-2}\mu r\log(nn_3)/(nn_3)$.
	Thus, $\norm{\rho^{-1}\PT\Pomega\PT-\PT} \leq \epsilon$ holds with high probability for some numerical constant $C_0$.
\end{proof}

\subsection{Proof of Corollary \ref{supp_corollar31}}
\begin{proof}
	From Lemma \ref{supp_lem_kem1}, we have 
	\begin{equation*}
	\norm{\PT-(1-\rho)^{-1}\PT\Pomegao\PT}\leq \epsilon,
	\end{equation*}
	provided that $1-\rho\geq C_0 \epsilon^{-2}(\mu r\log(nn_3))/n$. Note that $\I = \Pomega+\Pomegao$, we have 
	\begin{equation*}
	\norm{\PT-(1-\rho)^{-1}\PT\Pomegao\PT} = (1-\rho)^{-1} (\PT\Pomega\PT - \rho \PT).
	\end{equation*}
	Then, by the triangular inequality
	\begin{equation*}
	\norm{\PT\Pomega\PT} \leq \epsilon (1-\rho) + \rho \norm{\PT} = \rho + \epsilon(1-\rho).
	\end{equation*}
	The proof is completed by using $\norm{\Pomega\PT}^2 = \norm{\PT\Pomega\PT}$.
\end{proof}

\subsection{Proof of Lemma \ref{supp_lem_keyinf}}
\begin{proof} For any tensor $\Z\in\Tm$, we write
	\begin{align*}
	\rho^{-1}\PT\Pomega(\Z) = \sum_{ijk}\rho^{-1}\delta_{ijk} z_{ijk}\PT(\eijk).
	\end{align*}
	The $(a,b,c)$-th entry of $\rho^{-1}\PT\Pomega(\Z)-\Z$ can be written as a sum of independent random variables, i.e.,
	\begin{align*}
	& \inproduct{\rho^{-1}\PT\Pomega(\Z)-\Z}{\eabc} \\
	= & \sum_{ijk} (\rho^{-1}\delta_{ijk}-1)z_{ijk} \inproduct{\PT(\eijk)}{\eabc} \\
	:=& \sum_{ijk} t_{ijk},
	\end{align*}
	where $t_{ijk}$'s are independent and $\mathbb{E}(t_{ijk})=0$. Now we bound $|t_{ijk}|$ and $|\sum_{ijk}\mathbb{E}[t_{ijk}^2]|$. First
	\begin{align*}
	&|t_{ijk}|\\	
	\leq & \rho^{-1} \norm{\Z}_\infty \norm{\PT(\eijk)}_F\norm{\PT(\eabc)}_F \\
	\leq & \frac{2\mu r}{nn_3\rho}\norm{\Z}_\infty.
	\end{align*}
	Second, we have
	\begin{align*}
	& \left|\sum_{ijk}\mathbb{E}[t_{ijk}^2]\right |\\
	\leq & \rho^{-1}\norm{\Z}_\infty^2\sum_{ijk}\inproduct{\PT(\eijk)}{\eabc}^2 \\
	=  & \rho^{-1}\norm{\Z}_\infty^2\sum_{ijk}\inproduct{\eijk}{\PT(\eabc)}^2 \\
	=  & \rho^{-1}\norm{\Z}_\infty^2\norm{\PT(\eabc)}_F^2 \\
	\leq & \frac{2\mu r}{nn_3\rho}\norm{\Z}_\infty^2.
	\end{align*}
	Let $\epsilon\leq 1$. By Lemma \ref{supp_lembenmatrix}, we have
	\begin{align*}
	& \mathbb{P} \left[ |[\rho^{-1}\PT\Pomega(\Z)-\Z]_{abc}| > \epsilon\norm{\Z}_\infty\right] \\
	= & \mathbb{P}\left[ \left| \sum_{ijk} {t}_{ijk}\right|> \epsilon\norm{\Z}_\infty\right] \\
	\leq & 2 \exp\left( -\frac{3}{8} \cdot \frac{\epsilon^2\norm{\Z}^2_\infty}{2\mu r\norm{\Z}^2_\infty/(nn_3\rho) } \right) \\
	\leq & 2(nn_3)^{-\frac{3}{16}C_0},
	\end{align*}
	where the last inequality uses  $\rho\geq C_0\epsilon^{-2}\mu r\log(nn_3)/(nn_3)$.
	Thus, $\norm{\rho^{-1}\PT\Pomega(\Z)-\Z}_\infty\leq \epsilon\norm{\Z}_\infty$ holds with high probability for some numerical constant $C_0$. 	
\end{proof}

\subsection{Proof of Lemma \ref{supp_lempre3}}

\begin{proof}
	Denote the tensor $\HH_{ijk} = \left(1-\rho^{-1}\delta_{ijk}\right)z_{ijk}\eijk$. Then we have
	\begin{equation*}
	(\I-\rho^{-1}\Pomega)\Z = \sum_{ijk}\HH_{ijk}.
	\end{equation*}
	Note that $\delta_{ijk}$'s are independent random scalars. Thus, $\HH_{ijk}$'s are independent random tensors and $\Hmbar_{ijk}$'s are independent random matrices. 
	Observe that $\mathbb{E}[{\Hmbar}_{ijk}] = \0$ and $\norm{{\Hmbar}_{ijk}} \leq {\rho}^{-1} \norm{\Z}_\infty$. We have
	\begin{align*}
	&\normlarge{\sum_{ijk} \mathbb{E} [ {\Hmbar}^*_{ijk} {\Hmbar}_{ijk} ]  } \\
	= & \normlarge{\sum_{ijk} \mathbb{E} [ {\HH}^*_{ijk} *{\HH}_{ijk} ]  } \\
	= & \normlarge{\sum_{ijk}   \mathbb{E}[ (1-\rho^{-1}{\delta_{ijk}})^2 ] z_{ijk}^2 ({\ej}*\ej^*) } \\
	= & \normlarge{\frac{1-\rho}{\rho} \sum_{ijk} z_{ijk}^2 ({\ej}*\ej^*) } \\
	\leq & { \frac{nn_3}{\rho} }\norm{\Z}_\infty^2.
	\end{align*}
	A similar calculation yields $\normlarge{\sum_{ijk} \mathbb{E} [{\Hmbar}_{ijk}^* {\Hmbar}_{ijk} ]  }\leq { \rho^{-1}nn_3 }\norm{\Z}_\infty^2$. Let $t = \sqrt{C_0{nn_3\log(nn_3)}/{\rho}}\norm{\Z}_\infty$. When $\rho\geq C_0\log(nn_3)/(nn_3)$, we apply Lemma \ref{supp_lembenmatrix} and obtain
	\begin{align*}
	& \mathbb{P}\left[ \norm{(\I-\rho^{-1}\Pomega)\Z } > t \right]  \\
	= & \mathbb{P}\left[ \normlarge{  \sum_{ijk} {\HH}_{ijk} } > t \right] \\
	= & \mathbb{P}\left[ \normlarge{  \sum_{ijk} {\Hmbar}_{ijk} } > t \right] \\
	\leq & 2nn_3 \exp\left( -\frac{3}{8} \cdot \frac{C_0nn_3\log(nn_3)\norm{\Z}_\infty^2/\rho}{nn_3\norm{\Z}_\infty^2/\rho } \right) \\
	\leq & 2(nn_3)^{1-\frac{3}{8}C_0}.
	\end{align*}
	Thus, $ \norm{(\I-\rho^{-1}\Pomega)\Z } > t$ holds with high probability for some numerical constant $C_0$.
\end{proof}